\documentclass[10pt,logo,copyright]{nvidiatechreport}
\usepackage[authoryear,sort&compress,round]{natbib}

\usepackage[utf8]{inputenc} % allow utf-8 input
\usepackage[T1]{fontenc}    % use 8-bit T1 fonts

\usepackage{parskip}        % no paragraph indents
\usepackage{url}            % simple URL typesetting
\usepackage{booktabs}       % professional-quality tables
\usepackage{amsfonts}       % blackboard math symbols
\usepackage{nicefrac}       % compact symbols for 1/2, etc.
\usepackage{microtype}      % microtypography
\usepackage{xcolor}         % colors
\usepackage[dvipsnames]{xcolor} % more color names
\usepackage{graphicx}
\usepackage{animate}        % for 360 video in teaser
\usepackage{subcaption}
\usepackage{tabularx}
\usepackage{makecell}
\usepackage{adjustbox}
\usepackage{setspace}
\newcolumntype{M}[1]{>{\centering\arraybackslash}m{#1}}
\usepackage{float}
\usepackage{tikz}
\usetikzlibrary{positioning,shapes,arrows}
\usepackage{amsmath,amsfonts,bm, bbm,leftindex}
\usepackage{multirow}
\usepackage{comment}
\usepackage{gensymb}
\usepackage{lipsum}
\usetikzlibrary{arrows.meta, positioning, fit}
\usepackage[para]{threeparttable}
\usepackage{tikz}
\usetikzlibrary{tikzmark}

\usepackage[symbol]{footmisc}

\renewcommand{\today}{2025-12-8}

\usepackage[group-separator = {,},  % use comma
            group-minimum-digits = 4,  % start grouping at 4 digits
            group-digits = integer     % only group the integer part
           ]{siunitx}

\definecolor{darkred}{rgb}{0.7, 0.0, 0.0}

\usepackage{pifont}
\usepackage[nameinlink]{cleveref}
\crefname{equation}{Eq.}{Eqs.}
\crefname{figure}{Fig.}{Figs.}
\crefname{section}{Sec.}{Sec.}
\crefname{appendix}{App.}{App.}
\crefname{table}{Tab.}{Tabs.}
\crefname{algorithm}{Algo}{Algo}
\crefname{thm}{Thm}{Thm}
\Crefname{thm}{Thm}{Thm}
\crefname{prop}{Prop}{Prop}
\usepackage{xcolor}
\usepackage{tcolorbox}
\usepackage{wrapfig}
\tcbuselibrary{breakable} % Important for multi-line content
\tcbset{
  colback=gray!5!white,
  colframe=gray!60!black,
  boxrule=0.6pt,
  arc=3pt
}

\newtcolorbox{promptbox}[1][]{%
  title=#1,
  coltitle=black,
  fonttitle=\bfseries,
  colbacktitle=gray!15,
  breakable,
}

\usepackage{listings}
\usepackage{courier}
\usepackage[dvipsnames]{xcolor}

\usepackage{microtype}
\usepackage{tcolorbox}
\usepackage{adjustbox}

\newcommand{\crefnames}[3]{%
  \@for\next:=#1\do{%
    \expandafter\crefname\expandafter{\next}{#2}{#3}%
  }%
}

\title{Nemotron-Cascade: Scaling Cascaded Reinforcement Learning for General-Purpose Reasoning Models}

\definecolor{c0}{cmyk}{1,0.3968,0,0.2588} 
\definecolor{c1}{cmyk}{0,0.6175,0.8848,0.1490} 
\definecolor{c2}{cmyk}{0.1127,0.6690,0,0.4431} 
\definecolor{c3}{cmyk}{0.3081,0,0.7209,0.3255} 
\definecolor{nvg}{HTML}{33CC00}
\newtcbox{\hlprimary}{on line,colback=c0!10,colframe=white,size=fbox,arc=3pt, box align=base,before upper=\strut, top=-2pt, bottom=-4pt, left=-1pt, right=-1pt, boxrule=0pt}
\newtcbox{\hlprimarytab}{on line, box align=base, colback=c0!10,colframe=white,size=fbox,arc=3pt, before upper=\strut, top=-2pt, bottom=-4pt, left=-2pt, right=-2pt, boxrule=0pt}
\newtcbox{\hlsecondary}{on line,colback=c1!10,colframe=white,size=fbox,arc=3pt, box align=base,before upper=\strut, top=-2pt, bottom=-4pt, left=-1pt, right=-1pt, boxrule=0pt}
\newtcbox{\hlsecondarytab}{on line, box align=base, colback=c1!10,colframe=white,size=fbox,arc=3pt, before upper=\strut, top=-2pt, bottom=-4pt, left=-2pt, right=-2pt, boxrule=0pt}
\newtcolorbox{hlmultiline}{on line,colback=decentgrey!75,colframe=white,size=fbox,arc=3pt, box align=base, top=0pt, bottom=2pt, boxrule=0pt, before=\adjustbox{valign=c}\bgroup, after=\egroup, before upper=\strut}
\newtcbox{\hlmaintab}{on line, box align=base, colback=nvg!25,colframe=white,size=fbox,arc=3pt, before upper=\strut, top=-2pt, bottom=-4pt, left=-2pt, right=-2pt, boxrule=0pt}

\newcolumntype{Y}{>{\centering\arraybackslash}X}
\newcolumntype{Z}{>{\raggedleft\arraybackslash}X}

\newcommand{\dashifted}{{\scriptsize$\downarrow$}}
\newcommand{\da}[1]{{\scriptsize\hlprimarytab{\dashifted{#1}}}}

\newcommand{\uashifted}{{\scriptsize$\uparrow$}}
\newcommand{\ua}[1]{{\scriptsize\hlsecondarytab{\uashifted{#1}}}}

\newcommand{\xashifted}{{\scriptsize$\uparrow$}}
\newcommand{\xa}[1]{{\scriptsize\hlmaintab{\xashifted{\textbf{#1}}}}}

\definecolor{c4}{cmyk}{0.6765,0.2017,0,0.0667} 
\definecolor{c5}{cmyk}{0,0.8765,0.7099,0.3647} 

\definecolor{darkgrey}{RGB}{149,149,149}
\definecolor{decentgrey}{RGB}{242,242,242}

\author{\vspace{-.3cm}
Boxin Wang\footnote[1]{Equal technical contribution, with author names ordered alphabetically by first name.},~Chankyu Lee$^*$,~Nayeon Lee$^*$,~Sheng-Chieh Lin$^*$,~Wenliang Dai$^*$,~Yang Chen$^*$,~Yangyi Chen$^*$, 
Zhuolin Yang$^*$,  
\small\textbf{~Zihan Liu$^*$, Mohammad~Shoeybi,~ Bryan~Catanzaro,~
Wei Ping$^*$\footnote[2]{Leads the effort. Correspondence to: <wping@nvidia.com>.}
}
\vspace{-.5cm}
}

\begin{abstract}
{\normalfont
\vspace{-.1cm}
Building general-purpose reasoning models with reinforcement learning (RL) entails substantial cross-domain heterogeneity, including large variation in inference-time response lengths and verification latency.
Such variability complicates the RL infrastructure, slows training, and makes training curriculum (e.g., response length extension) and hyperparameter selection challenging. 
In this work, we propose cascaded domain-wise reinforcement learning (Cascade RL) to develop Nemotron-Cascade, capable of operating in both \emph{instruct} and deep \emph{thinking} modes, without any performance gap relative to a thinking-only counterpart.
Departing from conventional approaches that blend heterogeneous prompts from different domains, Cascade RL orchestrates sequential, domain-wise RL, reducing engineering complexity and delivering state-of-the-art performance across a wide range of benchmarks. 
Notably, RLHF for alignment, when used as a pre-step, boosts the model's reasoning ability far beyond mere preference optimization, and subsequent domain-wise RLVR stages rarely degrade the benchmark performance attained in earlier domains and may even improve it (see an illustration in Figure~\ref{fig:lcb_performance_through_cascade_rl}).
Our 14B model, after RL, outperforms its SFT teacher, DeepSeek-R1-0528, on LiveCodeBench v5/v6/Pro and achieves silver-medal performance in the 2025 International Olympiad in Informatics (IOI).
We transparently share our training and data recipes.
}
\end{abstract}

\begin{document}

\maketitle

\abscontent

\begin{figure}[h]
\vspace{-.2cm}
\centerline{\includegraphics[width=.95\linewidth]
{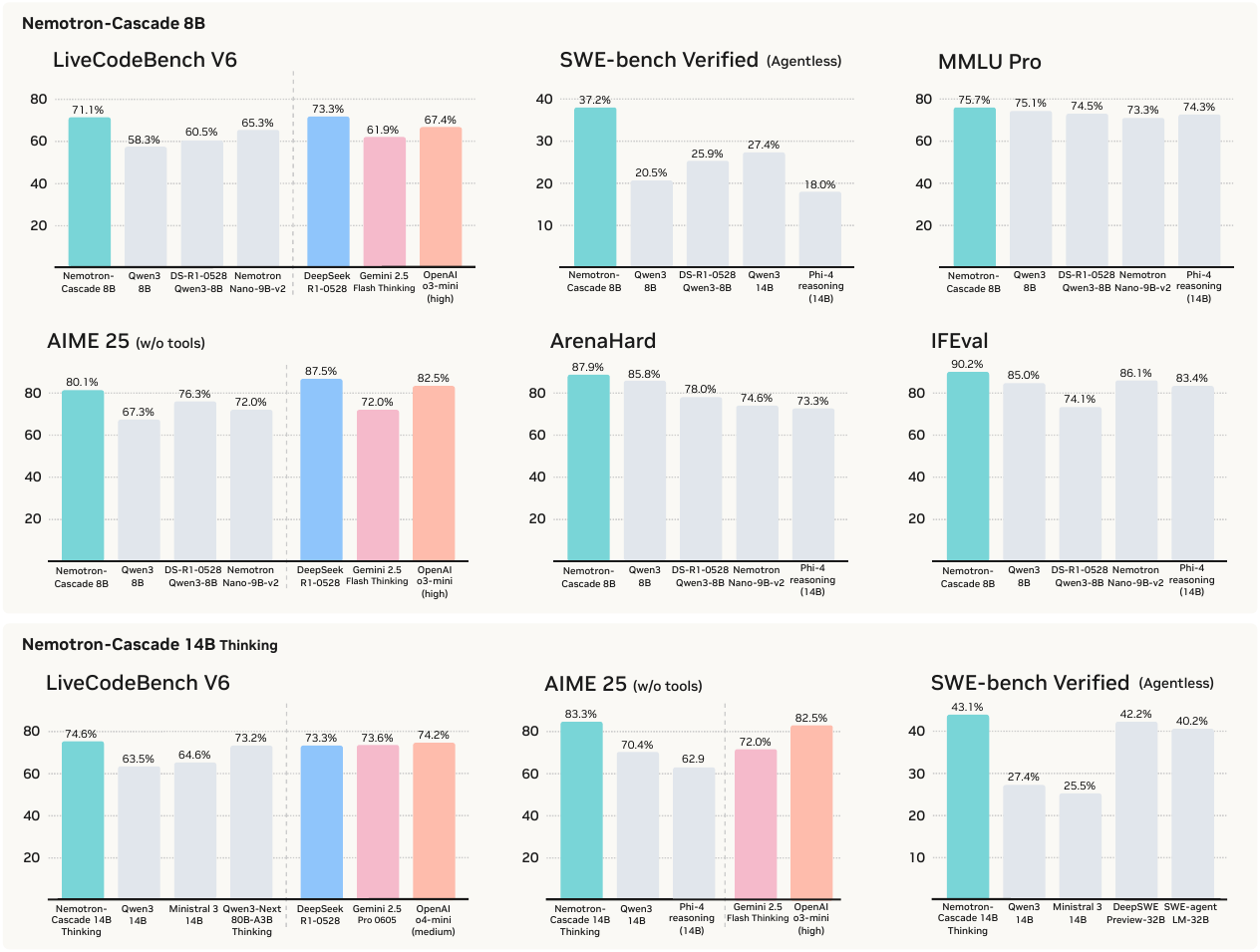}}
\label{fig:first_page}
\vspace{-.19cm}
\end{figure}

\newpage

\tableofcontents

\newpage

\renewcommand{\thefootnote}{\arabic{footnote}}

\section{Introduction}
\label{sec:introduction}

\begin{figure}[t!]
\centerline{\includegraphics[width=.85\linewidth]{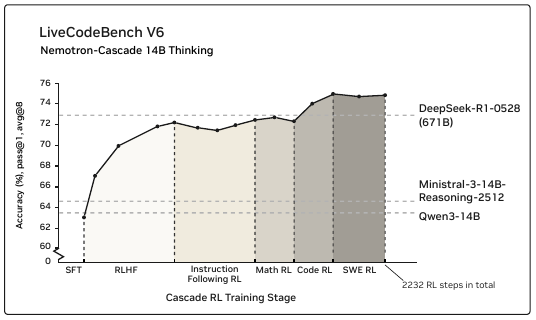}}
\caption{The LiveCodeBench v6 (08/24–05/25) performance of the Nemotron-Cascade-14B-Thinking model throughout the Cascade RL process. Note that DeepSeek-R1-0528 (671B) serves as the teacher model for SFT data curation. }
\label{fig:lcb_performance_through_cascade_rl}
\end{figure}

Reinforcement learning~(RL) serves as a cornerstone for developing general-purpose LLMs with advanced reasoning capabilities by substantially improving  alignment from human feedback~(RLHF)~\citep{ouyang2022training} and enhancing reasoning performance through verifiable rewards~(RLVR)~\citep{guo2025deepseek}.
However, training general-purpose reasoning models with RL involves substantial heterogeneity across different domains, both in response length and reward signal computation.
For example, fast symbolic rule-based verification is employed for mathematical reasoning tasks, slow execution-based verification is applied to code generation and software patching, and reward-model–based scores are computed for alignment and creative writing.
This domain-specific heterogeneity complicates the RL infrastructure, slows down training, and makes training curriculum~(e.g., maximum response length extension) and hyperparameter selection more challenging.

In our previous study~\citep{chen2025acereason}, we proposed performing  RL across the math and code domains in a cascaded manner, which first trains on math-only prompts and then on code-only prompts.
This cascaded paradigm yields several advantages:
\emph{a)} Rule-based math verification can be executed rapidly and is orders of magnitude faster than code verification, allowing the model to be updated immediately without waiting for longer verification cycles required for code prompts;
\emph{b)} math RL improves performance on both math and, surprisingly, code benchmarks; and
\emph{c)} subsequent code RL significantly enhances code benchmark performance without degrading math results.
In this work, we scale up the cascaded cross-domain RL paradigm to a much broader range of domains to build general-purpose reasoning models.

Since the introduction of OpenAI o1~\citep{openai2024o1}, model releases in the LLM community have generally fallen into two categories: \emph{thinking} models that generate substantially more reasoning tokens before providing an answer (e.g., DeepSeek-R1~\citep{guo2025deepseek}, OpenAI o3 and o4-mini~\citep{openai2025o3o4mini}, Kimi-K2-Thinking~\citep{Kimi_K2_Thinking}), and \emph{instruct} or \emph{non-thinking}  models that produce instant answers (e.g., DeepSeek-V3~\citep{liu2024deepseek}, GPT-4.5~\citep{openai2025gpt45}, Kimi-K2-Instruct~\citep{team2025kimi_k2}).
Meanwhile, it would be ideal to build one \emph{unified reasoning model} that can operate in both \emph{non-thinking} and \emph{thinking} modes while integrating all capabilities together in a single model.
This would \emph{i)}~greatly simplify model release and production pipelines, and \emph{ii)}~align more closely with the ultimate goal of artificial general intelligence.
Consequently, substantial efforts have been devoted to developing a single \emph{unified} model~\citep{yang2025qwen3, openai2025gpt5, deepseek_v3_1}.
%in which the \emph{thinking} and \emph{non-thinking} modes are controlled through the use of tags or special tokens.

%
Certain technical challenges have been recognized in efforts to “smoothly integrate everything”~\citep[e.g.,][]{altman2025}, including degradation in reasoning benchmark performance of the \emph{unified} model when operating in \emph{thinking} mode, compared to a dedicated \emph{thinking} model.
For example, although the Qwen3 series~\citep{yang2025qwen3} was initially released as a set of unified reasoning models, it was later reverted to separate \emph{thinking} and \emph{instruct} variants, with the dedicated \emph{thinking} models~\citep{Qwen3_235B_thinking} significantly outperforming the unified models in thinking mode.
The GPT-5 release has explored routing between two specialized models, where a standard \emph{instruct} model and a dedicated \emph{thinking} model are used together. However, the ultimate goal remains to integrate them into a single model~\citep{openai2025gpt5}.
%
% The DeepSeek-V3.1~\citep{deepseek_v3_1}, when operating in thinking mode, achieves performance comparable to the previous dedicated reasoning model, DeepSeek-R1-0528~\citep{deepseek_r1_0528}, on reasoning benchmarks. 
The DeepSeek-V3.1~\citep{deepseek_v3_1}, a unified model, achieves thinking-mode performance comparable to the earlier dedicated reasoning model DeepSeek-R1-0528 on reasoning benchmarks.
However, the technical details have not been disclosed, except that DeepSeek-V3.1 and DeepSeek-R1-0528 are based on different base models and were likely trained with distinct data blends.

\begin{figure}[t!]
\centerline{\includegraphics[width=.95\linewidth]{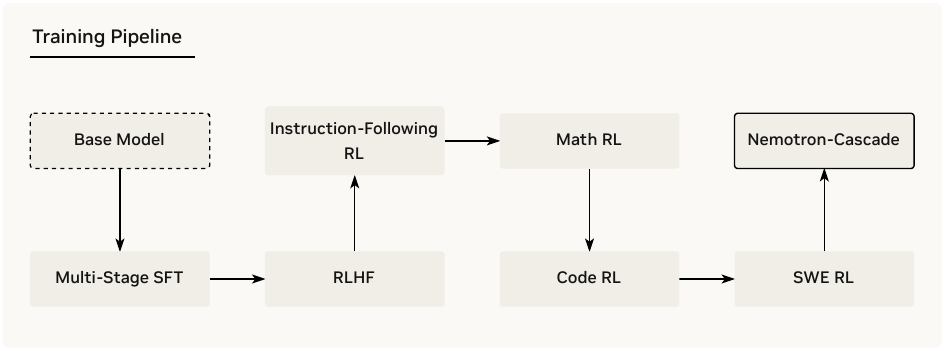}}
\caption{
Cascade RL applies sequential, domain-wise reinforcement learning after SFT, leading to substantial improvements across the corresponding domains.
}
\label{fig:training_pipeline}
\end{figure}

In this work, we focus on developing an open post-training recipe, using the pretrained Qwen3-8B-Base and Qwen3-14B-Base~\citep{yang2025qwen3} as starting points to support transparent comparison and facilitate knowledge sharing within the community.
In particular, we scale up the cascaded reinforcement learning~(\textbf{Cascade RL}) framework to develop Nemotron-Cascade models, setting new state-of-the-art results across multiple domains.
An overview of the training pipeline is shown in Figure~\ref{fig:training_pipeline}.
Cascade RL trains models sequentially across domains, in contrast to approaches such as DeepSeek-R1~\citep{guo2025deepseek} and Qwen3~\citep{yang2025qwen3}, which blend diverse prompt distributions from all (reasoning) domains for joint RL training.
Moreover, we show that a \emph{unified reasoning model} can operate effectively in both \emph{thinking} and \emph{non-thinking} modes, closing the reasoning gap with the dedicated \emph{thinking} model while ensuring transparency via open data and training recipes.

Specifically, the contributions of our work include:
\begin{enumerate}[leftmargin=3.1em]
    % instruction tuning
    \item[$\blacksquare$]
    We scale up Cascaded Reinforcement Learning~(Cascade RL) across a broad spectrum of domains, including human-feedback alignment, strict instruction following, mathematical reasoning, competitive programming, and software engineering.
    The proposed Cascade RL framework offers notable advantages:
    \emph{i)} RLHF substantially improves overall response quality (e.g., reduces verbosity), thereby enhancing reasoning performance;
    \emph{ii)} subsequent domain-specific RL stages rarely degrade the benchmark performance attained in earlier domains and may even improve it, since RL is resistant to \emph{catastrophic forgetting}~(see Figure~\ref{fig:lcb_performance_through_cascade_rl} for a demonstration and the in-depth discussion in Section~\S\ref{subsec:rl_forgetting}); and
    \emph{iii)} RL hyperparameters and training curriculum can be tailored to each specific domain for optimal performance.
    \item[$\blacksquare$]
    We develop Nemotron-Cascade-8B \emph{unified} reasoning model that enable user control over \emph{thinking} and \emph{non-thinking / instruct}modes at each conversational turn. 
    We challenge the assumption that LLMs, especially smaller ones, lack the capacity to learn effectively from both non-thinking and thinking data. We show that the reasoning performance gap between the 8B unified model in \emph{thinking} mode and a dedicated 8B-Thinking model can be closed, even when both models are trained on the same thinking/reasoning data, while the unified model is additionally trained on non-thinking data.
    The \textbf{key techniques} behind this result are: (1) SFT data curation with parallel responses in thinking and instruct modes for the same prompts, and (2) RLHF training that fuses the two modes by allocating an equal split of sampled prompts to each mode in every batch.
    % \vspace{0.1em}
    %
    \item[$\blacksquare$]
    Our 8B/14B models trained using  Cascade RL method achieve state-of-the-art, best-in-class performance across a broad range of benchmarks encompassing all these domains.
    For example, our 14B dedicated Thinking model, with a inference budget of 64K-token, outperforms Gemini-2.5-Pro-06-05, o4-mini (medium), Qwen3-235B-A22B (thinking mode), and DeepSeek-R1-0528 (its SFT teacher) on the LiveCodeBench v5/v6~\citep{jain2024livecodebench}~(see Figure~\ref{fig:lcb_performance_through_cascade_rl}).
    It also achieves silver-medal performance on the 2025 International Olympiad in Informatics (IOI).
    %
    %
    % \vspace{0.1em}
    \item[$\blacksquare$]
    We transparently share our training and data curation recipes, and release the full collection of models and training data at: \url{https://huggingface.co/collections/nvidia/nemotron-cascade}.
\end{enumerate}

We organize the remainder of this report as follows.
We first highlight the main results in Section \S~\ref{sec:main_results}, and present the technical details in later sections.
Section \S~\ref{sec:sft} describes the supervised fine-tuning stage of our post-training recipe.
In Section \S~\ref{sec:rl}, we present the proposed Cascade RL framework.
We highlight the competitive programming results~(including IOI 2025) in Section~\S~\ref{sec:deepdive_on_code}.
We offer an in-depth analysis of RLHF in Sections \S~\ref{sec:deepdive_on_rlfh} and a further exploration of software engineering (SWE) tasks in Section \S~\ref{sec:deepdive_swe}.
Related work is discussed in Section~\S~\ref{sec:relatedwork}.

\section{Main Results}
\label{sec:main_results}

We evaluate our models and baselines on a comprehensive suite of benchmarks, including
MMLU~\citep{hendrycks2020measuring}, MMLU-Pro~\citep{wang2024mmlu}, GPQA-Diamond~\citep{rein2024gpqa}, IFEval~\citep{zhou2023instructionfollowingevaluationlargelanguage}, IFBench~\citep{pyatkin2025generalizingverifiableinstructionfollowing}, ArenaHard~\citep{li2024crowdsourced}, LiveCodeBench v5 and v6~\citep{jain2024livecodebench}, LiveCodeBench Pro~\citep{zheng2025livecodebench}, SWE-bench Verified~\citep{openai2024swe_verified, jimenez2023swe}, and BFCL-V3~\citep{patil2025bfcl}.
These benchmarks collectively cover general-knowledge reasoning, alignment and instruction following, mathematical reasoning, competitive programming, software engineering, and tool-use proficiency.
For baseline models, we use officially reported results whenever available.
For Nemotron-Cascade models, we use a maximum generation length of 64K tokens and set the temperature to 0.6 and top-p to 0.95 for reasoning tasks.
The benchmarks and detailed evaluation setups are described in detail in Appendix~\ref{appendix:benchmarks}.

{
\addtolength{\tabcolsep}{-0.3em}
\begin{table}[t]
\centering
\footnotesize
\renewcommand{\arraystretch}{1.15}
\caption{\textbf{Main results}. For \emph{unified} reasoning models, we report reasoning-related benchmark results in thinking mode. For IFEval and IFBench, we report the higher score obtained from either \emph{thinking} or \emph{non-thinking} mode. \textbf{LCB} stands for LiveCodeBench.
When comparing with our inital SFT models in Table~\ref{tab:results_after_sft}, \textbf{\xa{number}} indicates the improvement achieved by applying the  Cascade RL to initial SFT models.
$^\dagger$Genini-2.5 uses its own scaffolds for SWE evaluation rather than  Agentless setup.
}
\label{tab:main_results}
\begin{adjustbox}{width={1.0\textwidth}}
% \hspace{-.8cm}
\begin{tabular}{lccc|cc|cc}
\toprule
\shortstack[l]{\textbf{Benchmark}\\\textbf{Metric: pass@1}}
 & \shortstack{\textbf{Qwen3}\\\textbf{8B}}
 & \shortstack{\textbf{Nemotron-Nano}\\\textbf{9B-v2}}
  & \shortstack{\textbf{Qwen3}\\\textbf{14B}}
 & \shortstack{\textbf{DeepSeek-R1}\\\textbf{0528 671B}}
 & \shortstack{\textbf{Gemini-2.5}\\\textbf{Flash-Thinking}}
  & \shortstack{\textbf{Nemotron}\\\textbf{Cascade-8B}} 
 & \shortstack{\textbf{Nemotron-Cascade}\\\textbf{14B-Thinking}} \\
\midrule
\multicolumn{7}{l}{\textbf{Knowledge Reasoning}} \\
MMLU                    & 83.0 & 82.6 & 84.9 & 89.9  & -- & 83.7~\xa{0.7} & 85.1 \xa{0.2} \\
MMLU-Pro                & 75.1 & 73.3 & 77.6 & 85.0 & 81.9 & 75.7~\xa{1.3}  & 77.0 \xa{1.0} \\
GPQA-Diamond            & 62.0 & 64.0 & 64.0 & 81.0 & 82.8 & 66.5~\xa{3.0}  & 69.6 \xa{1.3} \\
\midrule
\multicolumn{7}{l}{\textbf{Alignment}} \\
ArenaHard & 85.8 & 74.6 & 91.7 & 95.1 & 95.7 & 87.9~\xa{17.9} & 89.5 \xa{12.6} \\
IFEval (strict prompt)  & 85.0 &  86.1 & 85.4 & 84.1 & 89.8 & 90.2~\xa{19.4} & 81.9 \xa{12.1} \\
IFBench     & 34.4 & 37.4 & 33.7 & 38.0 & 36.1 & 40.8~\xa{19.6} & 41.7 \xa{17.4} \\
\midrule
\multicolumn{7}{l}{\textbf{Math}} \\
AIME 2024 (no tools)               & 76.0 & 81.9 & 79.3 & 91.4 & 82.3 & 89.5~\xa{5.9} & 89.7 \xa{2.8} \\
AIME 2025 (no tools)              & 67.3 & 72.0 & 70.4 & 87.5 & 72.0 & 80.1~\xa{7.3} & 83.3 \xa{2.2} \\
\midrule
\multicolumn{7}{l}{\textbf{Code}} \\
LCB v5 (08/24-02/25)        & 61.2 & 68.2 & 65.2 &74.8 & 63.4 & {74.3}~\xa{15.1} & \textbf{77.5} \xa{11.4} \\
LCB v6 (08/24-05/25)        & 58.3 & 65.3 & 63.5 & 73.3 & 61.9 & {71.1}~\xa{14.4}  & \textbf{74.6} \xa{11.5} \\
LCB Pro 25Q2 (Easy)         & 46.1 &  59.3 & 53.6 & 63.9 & 47.4 & {65.7}~\xa{20.6} & \textbf{68.9} \xa{11.2} \\
LCB Pro 25Q2 (Med)              & 2.2 &  4.8 & 2.6 & 7.0 &  1.8 & {6.4}~\xa{3.8} & \textbf{10.5} \xa{3.5} \\
SWE Verified~(Agentless)       & 20.5 & -- & 27.4 & 57.6 & $^\dagger$48.9 & 37.2~\xa{11.1} & \textbf{43.1} \xa{8.6} \\
--Test-Time Scaling              & -- & -- & -- & -- & -- & 43.6 & 53.8 \\
\midrule
\multicolumn{7}{l}{\textbf{Tool Calling}} \\
BFCL V3                 & 68.1 & 66.9 &  70.4 & 67.9 & 68.6 & 64.4  & 67.5 \\
\bottomrule
\end{tabular}
\end{adjustbox}
\end{table}
}

The main results are shown in Table~\ref{tab:main_results}. All scores are reported as pass@1, averaged over k generations~(avg@k) per prompt, where k is chosen appropriately (typically between 4 and 64, depending on the test set size). 
Our \emph{unified} reasoning model, Nemotron-Cascade-8B, and our dedicated \emph{thinking} model, Nemotron-Cascade-14B-Thinking, achieve best-in-class performance across almost all benchmarks.
We \xa{highlight} the substantial improvements achieved by applying the complete Cascade RL pipeline to our SFT models, compared with the results in Table~\ref{tab:results_after_sft}.

In this work, we also perform comprehensive ablations by evaluating the models after each stage of the Cascade RL pipeline illustrated in Figure~\ref{fig:training_pipeline}.
The results after each stage of the pipeline are summarized as follows: Table~\ref{tab:results_after_sft} for SFT, Table~\ref{tab:results_after_rlhf} for RLHF, Table~\ref{tab:results_after_ifrl} for Instruction-Following RL, Table~\ref{tab:results_after_mathrl} for Math RL, Table~\ref{tab:results_after_coderl} for Code RL, and Table~\ref{tab:results_after_swerl} for SWE RL.

Notably, we observe substantial improvements on LiveCodeBench (LCB) and LCB Pro, with Nemotron-Cascade-8B achieving 74.3 on LCB v5 and 71.1 on LCB v6. Despite being only an 8B model, its performance is highly comparable to DeepSeek-R1-0528 (671B), which reports 74.8 on LCB v5 and 73.3 on LCB v6.
Note that DeepSeek-R1-0528 (671B) serves as the teacher model during SFT, generating all responses to code prompts used in our SFT data curation~(see Section~\S~\ref{subsec:code_sft_data}).
Remarkably, our Nemotron-Cascade-14B-Thinking model surpasses DeepSeek-R1-0528 by a clear margin across all splits of LCB and LCB Pro benchmarks. These results highlight the exceptional effectiveness of the proposed Cascade RL framework in enhancing reasoning capabilities.
Additional results, including those for IOI 2025, are provided in Section~\ref{sec:deepdive_on_code}.

For SWE-bench Verified, the best general-purpose open 8B and 14B LLMs perform poorly on this challenging benchmark. The specialized model, DeepSWE-32B~\citep{deepswe}, built on Qwen3-32B and specialized for SWE tasks, achieves a pass@1 accuracy of 42.2\%. In comparison, our general-purpose 8B and 14B models, attain 37.2\% and 43.1\%, respectively. 
Additional details and test-time scaling results are provided in Section~\S\ref{sec:deepdive_swe}.

\section{Supervised Fine-Tuning}
\label{sec:sft}
In this section, we describe the training framework and data curation for supervised fine-tuning~(SFT), the first stage of our post-training pipeline.
This stage equips the model with foundational skills and capabilities, which are then substantially enhanced through cascaded reinforcement learning (Cascade RL) in the subsequent stages.

\subsection{Training Framework}

\subsubsection{Multi-Stage SFT}
Our SFT curriculum consists of two stages, spanning a broad spectrum of domains, including math, coding, science, tool use, and software engineering, as well as general domains such as multi-turn dialogue, knowledge-intensive question answering, creative writing, role-playing, safety, and instruction following.
Details of data curation for these domains are provided in \S~\ref{subsec:sft_data_curation}.
The SFT curriculum is outlined as follows:
\begin{itemize}
    \item 
    \textbf{Stage 1 (16K).}  This stage includes general-domain data as well as math, science, and code reasoning data, with a maximum sequence length of 16K tokens.  
    For the general-domain data, each prompt contains parallel responses in both \emph{thinking} and \emph{non-thinking} modes, whereas the math, science, and code data include only \emph{thinking} mode responses.  
    Training is conducted for one epoch.
    \item 
    \textbf{Stage 2 (32K).}  This stage further enhances the model’s reasoning capabilities with longer responses of up to 32K tokens and equips it with tool use and software engineering skills.
    To achieve this, we recombine the general-domain data with new Stage-2 math, science, and code reasoning data (up to 32K tokens), along with the tool use and software engineering datasets.
    Except for the general-domain data, all other domains contain only \emph{thinking}-mode responses.
    Training is conducted for one epoch.
\end{itemize}
The SFT training hyperparamters can be found in Appendix~\ref{appendix:training_hyperparams}.

\begin{figure}[t!]
\centerline{\includegraphics[width=1.0\linewidth]{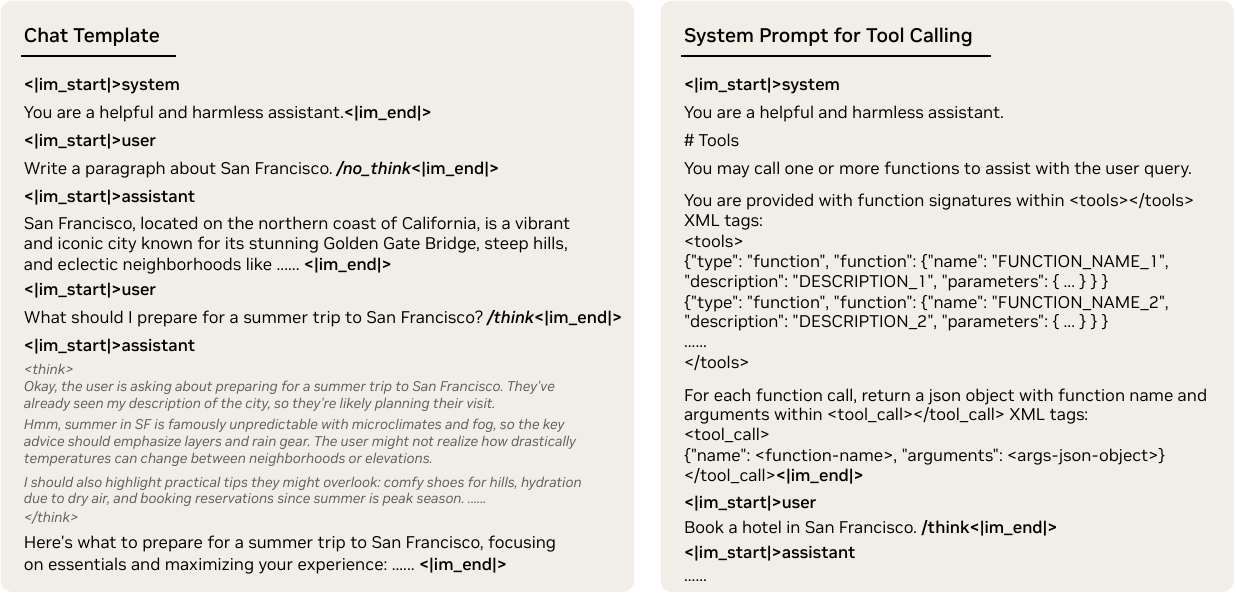}}
\caption{(Left)~The chat template employs the \emph{/think} and \emph{/no\_think} tags in the user prompt to control whether the model operates in the \emph{thinking} or \emph{non-thinking} generation mode. 
(Right)~For tool calling, the available tools are listed in the system prompt. The model is instructed to call tools within the \emph{<tool\_call>} and \emph{</tool\_call>} tags.
}
\label{fig:chat_templates}
\end{figure}

\subsubsection{Chat Template}
% @wenliang
We define the model’s interaction schema, which is particularly important for our \emph{unified} reasoning model supporting both \emph{thinking} and \emph{non-thinking} generation modes.
We adopt the standard ChatML template~\citep{chatml} and introduce two control flags in the user prompt, \emph{/think} and \emph{/no\_think}, which explicitly instruct the model to generate responses in the corresponding mode.

Although prior work~\citep{yang2025qwen3,bakouch2025smollm3} adopts similar control flag mechanisms, we introduce several simplifications and enhancements to enable more precise and flexible control over the model’s generation behavior.
In contrast to \citet{bakouch2025smollm3}, which places the \emph{/think} and \emph{/no\_think} flags in the system prompt and therefore controls the entire conversation globally, we instead append the flag to each individual user prompt.
This design supports both global and local control: appending the same flag to every user turn enforces consistent global behavior, while varying the flags across multi-turns enables dynamic switching within a single conversation.

In contrast to the Qwen3 reasoning models~\citep{yang2025qwen3}, our approach further simplifies mode control.
Qwen3 employs a redundant mechanism that enables mode switching in two ways: either through explicit flags or by modifying the template via the \emph{enable\_thinking} argument, which implicitly determines the mode (defaulting to the thinking mode, while prepending an empty \emph{<think></think>} block activates the non-thinking mode). 
Our early experiments show that explicit flags result in more reliable mode transitions than template-based cues. Moreover, the flag-only design covers all use cases without any performance degradation. Consequently, we adopt the flag-based approach exclusively. With this simplification, the empty \emph{<think></think>} block is omitted in the non-thinking mode, as it is no longer needed.

% @zihan tool calling
For tool calling task, we specify all available tools in the system prompt within the \emph{<tools>} and \emph{</tools>} tags as illustrated on the right side of Figure~\ref{fig:chat_templates}. We further instruct the model to perform tool calls enclosed within the \emph{<tool\_call>} and \emph{</tool\_call>} tags.
%
% Specifically, the system prompt for tool calling is structured in the right side of Figure~\ref{fig:chat_templates}.

% \newpage
% \begin{promptbox}
% <|im\_start|>system\\
% You are a helpful and harmless assistant. \\ \\
% \# Tools \\ \\
% You may call one or more functions to assist with the user query. \\ \\
% You are provided with function signatures within \emph{<tools>}\emph{</tools>} XML tags: \\
% \emph{<tools>} \\
% \{"type": "function", "function": \{"name": "FUNCTION\_NAME\_1", "description": "DESCRIPTION\_1", "parameters": \{ ... \} \} \} \\
% \{"type": "function", "function": \{"name": "FUNCTION\_NAME\_2", "description": "DESCRIPTION\_2", "parameters": \{ ... \} \} \} \\
% ... ... \\
% \emph{</tools>} \\ \\
% For each function call, return a json object with function name and arguments within \emph{<tool\_call>}\emph{</tool\_call>} XML tags:\\
% \emph{<tool\_call>}\\
% \{"name": <function-name>, "arguments": <args-json-object>\}\\
% \emph{</tool\_call>}\\
% <|im\_start|>user\\
% Book a hotel in San Francisco. /think
% \end{promptbox}

\subsection{SFT Data Curation}
\label{subsec:sft_data_curation}
% \subsubsection{Multi-turn Conversational Data}
\subsubsection{General-Domain Data} \label{sec:general_conv_data}
% @Wenliang
We curate a comprehensive corpus of 2.8M samples, comprising 3.2B tokens from diverse general-domain datasets, to equip models with foundational skills and robust conversational abilities.
This corpus encompasses a wide range of tasks, including daily dialogue, question answering~\citep{SlimOrca,yuan2024advancing}, creative writing~\citep{xu2024magpie,allal2025smollm2}, safety~\citep{bercovich2025llama}, instruction following~\citep{lambert2024tulu3}, role-playing~\citep{lambert2024tulu3}, and others.
In addition, for knowledge-intensive tasks~\citep{hendryckstest2021,gema2024mmlu,wang2024mmlu} spanning general domains, we collect questions from publicly available datasets~\citep[e.g.,][]{longpre2023flan,allenai:qasc} and further augment them with domain-specific questions from challenging areas such as professional law and ethics, resulting in 1.2M samples comprising 1.5B tokens.

However, directly combining these corpora poses three notable challenges. First, many responses are excessively brief (e.g., single-word or minimal-sentence outputs) and therefore lack sufficient detail and elaboration. Second, response quality varies widely, with some datasets containing inaccurate or suboptimal answers. Third, due to the diverse origins and labeling conventions of these datasets, directly training on them leads to stylistic inconsistencies in model generation.
To address these issues, for each prompt, we generate parallel responses using DeepSeek-R1-0528~\citep{deepseek_r1_0528} and DeepSeek-V3-0324~\citep{deepseek_v3_0324} to obtain \emph{thinking} and \emph{non-thinking} format data, respectively, with a maximum sequence length of 16K, thereby ensuring stylistic and qualitative consistency.

To further enhance the training data quality, we apply several post-processing steps.
For samples with high-quality annotations, we retain their original responses to preserve diversity. 
Furthermore, for prompts with verifiable ground-truth answers (e.g., multiple-choice questions), we enhance generation accuracy by discarding responses that deviate from the ground-truth.
For samples without ground-truth answers, we cross-validate the generated responses using an auxiliary model~(Qwen2.5-32B-Instruct~\citep{yang2024qwen2_5}) to filter out potentially low-quality generations.

To address data scarcity in domains such as instruction following and creative writing, we generate multiple responses for each prompt using different random seeds, thereby enriching diversity and improving generation quality. To further enhance multi-turn conversational capabilities, we manually augment multi-turn samples in two ways. First, for single-turn samples in the creative writing domain, we add a second turn that instructs the model to rewrite or edit its previous response under specific requirements. Second, we randomly concatenate single-turn samples to construct multi-turn conversations, emulating real-world chatbot interactions.

% Overall, our final SFT dataset for general domains comprises 3.2B tokens, with 2.4B in the thinking format and 0.8B in the non-thinking format.
% multi-turn
% num of tokens
% multi responses with different random seeds

% @Wenliang 3.7B tokens MMLU 
% longpre2023flan,allenai:qasc
% To enable model's general conversational abilities in the thinking mode, we generate a corpus of 3.7B tokens with prompts from the same data sources mentioned in Section~\ref{sec:general_conv_data}. Specifically, we 
% For each prompt from same data sources mentioned in Section~\ref{sec:general_conv_data}, we generate responses in the reasoning format (i.e., thinking content in the <think></think> block and a summary after that) with a total of 3.7B tokens.  

\subsubsection{Math Reasoning Data}
% @Zihan
We utilize the math reasoning SFT prompts from AceReason-Nemotron-1.1~\citep{liu2025acereason} for our Stage-1 SFT training. These prompts encompass a diverse set of data sources, including AceMath~\citep{liu2024acemath}, NuminaMath~\citep{li2024numinamath}, and OpenMathReasoning~\citep{moshkov2025aimo2}. 
The corresponding responses are generated by DeepSeek-R1~\citep{guo2025deepseek}. 
We set the maximum context length to 16,384 tokens (16K) and filter out samples exceeding this limit to prevent response truncation, following the SFT configuration in AceReason-Nemotron-1.1. In total, we gather 353K unique prompts and generate multiple responses for each, resulting in 2.77M samples with an average of 7.8 responses per prompt. We perform data decontamination by removing any samples that share a 9-gram overlap with test samples from standard math benchmarks.

To further enhance the model’s reasoning capability, we employ DeepSeek-R1-0528~\citep{deepseek_r1_0528} to generate responses and construct the Stage-2 math SFT dataset. Compared to the original DeepSeek-R1, the updated DeepSeek-R1-0528 produces longer and more detailed reasoning trajectories, leading to improved performance on challenging problems. We set the maximum context length to 32,768 tokens (32K) to provide a larger reasoning token budget for the model. The Stage-2 prompt set is derived from Stage-1 by filtering out relatively easy questions, specifically those whose DeepSeek-R1 responses contain fewer than 2K tokens. In total, we obtain 163K prompts and generate 1.88M samples, with an average of 11.5 responses per prompt.
All math reasoning data in both Stage-1 and Stage-2 SFT are formatted in the \emph{thinking} mode.

\subsubsection{Code Reasoning Data}
\label{subsec:code_sft_data}
% @Zihan
Following a similar procedure as the math reasoning data construction, we adopt the code reasoning SFT prompts from AceReason-Nemotron-1.1~\citep{liu2025acereason}, which include data from TACO~\citep{li2023taco}, APPs~\citep{hendrycks2021measuring}, OpenCoder-Stage-2~\citep{huang2024opencoder}, and  OpenCodeReasoning~\citep{ahmad2025opencodereasoning}. We perform dataset deduplication to ensure all prompts are unique, resulting in 172K distinct prompts. Using DeepSeek-R1~\citep{guo2025deepseek}, we generate 1.42M samples for Stage-1 SFT, with an average of 8.3 responses per prompt and a maximum context length of 16,384 tokens (16K). For data decontamination, we filter out any samples that have a 9-gram overlap with any test sample from coding benchmarks.

To construct the Stage-2 code SFT dataset, we leverage prompts from OpenCodeReasoning~\citep{ahmad2025opencodereasoning} and OpenCoder-Stage2~\citep{huang2024opencoder}. OpenCodeReasoning provides a diverse set of challenging coding prompts, while OpenCoder-Stage2 covers coding tasks with starter code entry points. Similar to the Stage-2 math SFT dataset, we set the maximum context length to 32,768 tokens (32K). In total, we build 79K unique prompts and use DeepSeek-R1-0528 to generate 1.39M samples, averaging 17.6 responses per prompt.
All code reasoning data in both Stage-1 and Stage-2 SFT are formatted in the \emph{thinking} mode.

\subsubsection{Science Reasoning Data}
% @Zihan: prompt collection
% @wenliang: response re-generation
We curate science-related prompts from S1K~\citep{muennighoff2025s1simpletesttimescaling} and the post training datasets used in Llama-Nemotron~\citep{bercovich2025llama,NemotronPostTrainingDatasetV1}.
Given that many prompts in these sources are multiple-choice, we exclude samples where models focus on analyzing each option rather than directly solving the problem and determining the correct answer. As a result, we retain questions that require strong scientific knowledge and involve substantial reasoning or complex calculations.

To enrich the dataset with less common and more diverse question types, we leverage DeepSeek-R1-0528 \citep{deepseek_r1_0528} to generate rarer questions from each given prompt, following the synthetic question generation strategy used in~\citet{liu2024acemath}.
Finally, we perform data decontamination and remove any samples that have a 9-gram overlap with any test sample from science benchmark. 
In total, we collect 226K science prompts, generating 289K samples (up to 16K tokens) for Stage-1 SFT with DeepSeek-R1,  and 345K samples (up to 32K tokens) for Stage-2 SFT with DeepSeek-R1-0528.
All science reasoning data are formatted in the \emph{thinking} mode, and multiple responses are generated for selected high-quality prompts.
The Stage-2 science reasoning data are upsampled by 2$\times$ before being blended into the Stage-2 SFT dataset.

\subsubsection{Tool Calling Data}
% @Zihan
% BFCL~\citep{patil2025bfcl}
We utilize the tool calling dataset from Llama-Nemotron~\citep{NemotronPostTrainingDatasetV1}, which is specifically designed to train models for scenarios involving external tool usage, such as function calls.
The dataset is comprehensive, encompassing single-turn, multi-turn, and multi-step interactions.
For instance, some prompts require the model to ask clarification questions to gather sufficient information for a tool call, 
while other prompts may involve using several tools or even performing multiple rounds of tool calls before reaching a final answer.
It also includes cases where the model cannot find a suitable tool in the provided tool list.
For each conversation, all available tools are included in the system prompt, following the setup used in Qwen3~\citep{yang2025qwen3}. 
On average, each conversation includes 4.4 available tools.
This tool calling SFT dataset is used in Stage-2 SFT training, with responses generated by Qwen3-235B-A22B~\citep{yang2025qwen3}.
Note that all tool-calling data are formatted in the \emph{thinking} mode.
Overall, we collect 310K conversations comprising 1.41M user-assistant turns.

\subsection{Software Engineering Task}
\label{sec:swe}
Software engineering has become one of the most important applications of LLMs~\citep[e.g.,][]{anthropic2025claude_code}.
For example, SWE-bench Verified~\citep{jimenez2023swe} has emerged as a widely used benchmark for software engineering, consisting of real-world GitHub issues paired with their corresponding codebases and descriptions, where the goal is to generate repair patches that successfully resolve the issued problems.

\subsubsection{Agentless Framework}
\label{subsec:agentless_framework}
To evaluate automated software engineering capabilities, we adopt Agentless~\citep{xia2024agentless}, which decomposes the overall task into three stages (i.e., \textbf{localization}, \textbf{repair} and \textbf{patch validation}) without requiring the LLM itself to plan action sequences or operate external tools.

In this work, we employ a simplified Agentless framework similar to Agentless Mini~\citep{wei2025swe}, which streamlines the localization process to focus solely on identifying relevant issue files. This approach differs from the original Agentless workflow, which adopts a three-stage hierarchical localization strategy progressing from file-level to class/function-level and finally to line-level identification before proceeding to the repair and patch validation stages. By simplifying the localization process, the LLM can dedicate more reasoning capacity to the repair task itself, while reinforcement learning is consolidated into a focused objective that directly optimizes repair patch generation.

  For the code repair stage, the primary objective is to generate effective patch candidates that resolve the identified repository-level issue. Once the relevant files have been localized through earlier stages, the LLM is prompted to generate the repair edits that modify only the necessary portions of the codebase. To maximize contextual understanding, we concatenate multiple localized files and their surrounding code snippets (e.g., imports, class definitions, and dependent functions) into a unified prompt, enabling the model to reason over broader repository-level dependencies. Instead of rewriting entire repair files, the model is guided to produce targeted, diff-style patches that preserve unrelated code structure to reduce hallucinations and syntactic errors. 

For the patch validation stage, our framework operates through three phases: regression, reproduction and majority voting. In the regression phase, each candidate patch is initially evaluated through the repository’s existing regression test to ensure the compatibility. This step filters out the patches that introduce the failures or disrupt previously correct functionality to ensure that subsequent evaluation focuses exclusively on stable and syntactically valid candidates. In the reproduction phase, the framework generates 10 reproduction tests per issue instance to replicate the original bug behavior on the unmodified repository and to verify functional correctness after patch application. Reproduction tests that fail to trigger the original bug are discarded to maintain high diagnostic precision and the survived tests are then executed on each candidate patch, allowing the system to identify which patches successfully eliminate the reported issues. Finally, in the majority voting phase, we aggregate results across multiple sampled generations to select the most reliable patch. Among the highest-scoring candidates, we first prioritize the most frequently generated patch across test-time samples, reflecting consensus in the model’s repair reasoning capabilities. In the event of ties, we favor the patch with a shorter generation sequence or minimal edit distance, promoting conciseness and interpretability. This multi-phase validation framework ensures that the final selected patch is both functionally correct and robust to repository-level dependencies, balancing precision with efficiency.

We refer readers to Appendix~\ref{appendix:swe_template} for the prompts used at each stage. 
We detail the improved techniques used in this work, including enhanced code file localization and patch validation, compared to prior studies in \S~\ref{sec:deepdive_swe}.

\subsubsection{Data Curation}
\label{subsec:swe_data_curation}
\paragraph{Data source:}

Our training data for the software engineering task consist of the following open-source datasets:
\begin{itemize}
    \item SWE-Bench-Train~\citep{jimenez2023swe}, the training split generated through the same pipeline as the SWE-Bench evaluation set but without human verification.
    \item SWE-Fixer-Train~\citep{xie2025swe}, which consists of Python repositories with more than 100 pull requests, yielding 115K instances after applying heuristic filtering rules.
    \item SWE-reBench~\citep{badertdinov2025swe}, a public dataset containing over 21K interactive Python-based SWE tasks, constructed through a novel, automated, and scalable pipeline.
    \item SWE-Smith~\citep{yang2025swe}, a synthetic dataset comprising 50K instances from 128 GitHub repositories, generated by automatically injecting bugs into codebases.
\end{itemize}
To prevent evaluation data contamination, we implement a comprehensive deduplication process against SWE-bench Verified~\citep{jimenez2023swe}.
Specifically, we exclude all instances originating from repositories present in the evaluation dataset.
Additionally, we perform deduplication across training data from different sources to eliminate duplicate instances.
This deduplication process relies on matching both repository names and base commit identifiers to ensure that identical instances are removed.

\paragraph{Response generation:}

We construct SFT datasets for three sub-tasks in Agentless framework:
\begin{enumerate}
\item Code localization: Given a problem statement and the corresponding GitHub repository structure, the model identifies and lists the code files that are likely to contain bugs.
\item Code repair: Given a problem statement and the contents of one or more buggy code files, the model generates revised code patches that address the issues described in the problem statement.
\item Test code generation: Given a problem statement, code localization and repair patches, the model generates test code that validates the generated code patches.
\end{enumerate}

To construct the SFT datasets, we employ DeepSeek-R1-0528~\citep{deepseek_r1_0528} to generate multiple responses across the four datasets listed in Data Source.
We generate 8 responses per prompt for SWE-Bench-Train, SWE-reBench, and SWE-Smith, while producing 4 responses per prompt for the larger-scale SWE-Fixer-Train dataset. 
The input prompts are structured to include task specifications, problem statements, content of code files for repair tasks, and desired output formats. See Appendix~\ref{appendix:swe_template} for details of the template.
The model is instructed to output the comprehensive reasoning chain and the solution. 
For code localization, the solution contains a prioritized list of potential buggy code file names, ranked from most to least likely to contain bugs.
For code repair, the solution contains the code blocks to be replaced, and the code patch used to replace. 
For test code generation, the solution contains a set of unit tests and reproduction tests designed to verify both bug replication and patch correctness.

% \paragraph{Quality Check:}
% To ensure high-quality training data, we validate all generated responses against ground truth patches written by humans. 
% % 
% For repair task, we examine how closely the generated patches match the ground truth patches.
% Specifically, we measure the similarity between two patches with \emph{Unidiff}, a simple Python library to parse and interact with unified \emph{diff} data, following the method of~\citet{wei2025swe}.
% Through preliminary manual inspection, we find that a similarity threshold of 0.5 strikes a good balance, effectively identifying patches of sufficient quality for fine-tuning while filtering out lower-quality generations.

\paragraph{Data filtering and splitting for SFT and RL:}
To ensure high-quality training data, we validate all generated responses against ground truth annotations. 
Our filtering strategy varies across the three tasks.
For the localization task, we retain only those samples that include all buggy code files required to address the issue (i.e., recall equals 1.0).
For repair tasks,  we measure the similarity between the generated and ground-truth patches using \emph{Unidiff}, a lightweight Python library for parsing and interacting with unified \emph{diff} data, following the approach of~\citet{wei2025swe}.
We adopt a stratified approach: instances demonstrating consistent solution quality—defined as at least 4 out of 8 sampled responses exceeding the 0.5 similarity threshold for SWE-Bench-Train, SWE-reBench, and SWE-Smith—are included in the SFT dataset. 
More challenging instances—defined as those with at least 1 out of 8 sampled responses achieving non-zero similarity, excluding SFT samples—are reserved for the SWE RL training described in \S~\ref{subsec:swe_rl}.
Given the scale and diversity of SWE-Fixer-Train dataset, we include all prompt-response pairs surpassing the 0.5 similarity threshold.
For test code generation, we retain only the trajectories that can be successfully parsed and executed as reproduction tests without any syntax error.%

\paragraph{Dataset composition summary:}
The resulting Code Repair dataset comprises 127K instances distributed as follows: 17K from SWE-Bench-Train, 17K from SWE-reBench, 18K from SWE-Smith, and 77K from SWE-Fixer-Train.
This composition ensures comprehensive coverage of diverse coding scenarios while maintaining high data quality standards.
The resulting datasets for localization and test case generation consist of 92K and 31K samples, respectively.
All SWE datasets are upsampled by 3$\times$ before being incorporated into the Stage-2 SFT data blend.

\subsection{Results after SFT}
After the multi-stage SFT process, the results for the 8B \emph{unified} model and the 8B/14B \emph{thinking} models are summarized in Table~\ref{tab:results_after_sft}.
We find that 8B unified model performs on par with  dedicated 8B Thinking model across all reasoning-related benchmarks, while surpassing it on the IFEval benchmark—a task more naturally suited to the \emph{instruct} mode.
It is worth noting that both models are trained on the same SFT \emph{thinking} data; however, the unified model further incorporates \emph{non-thinking} data.
Due to resource constraints, we only train a 14B-Thinking model and provide stronger results than 8B models.
Next, we examine the results across all RL stages, highlighting the robustness and overall effectiveness of the Cascade RL framework.

\begin{table}[!t]
\centering
\footnotesize
\renewcommand{\arraystretch}{1.15}
\caption{The evaluation results of 8B/14B-Thinking and the \emph{unified} 8B models \textbf{after SFT} are presented below. For unified 8B model, we evaluate IFEval in the \emph{non-thinking} mode and all other benchmarks in the \emph{thinking} mode.
}
\label{tab:results_after_sft}
% \begin{adjustbox}{width={0.9\textwidth}}
\begin{tabular}{lcccc}
\toprule
\shortstack[l]{\textbf{Benchmark}\\\textbf{Metric: pass@1}}
 & \shortstack{\textbf{8B-Thinking}\\\textbf{SFT}}
 & \shortstack{\textbf{8B~(unified)}\\\textbf{SFT}} 
  & \shortstack{\textbf{14B-Thinking}\\\textbf{SFT}} 
 \\
\midrule
\multicolumn{4}{l}{\textbf{Knowledge and Reasoning}}  \\
MMLU~{\footnotesize\textcolor{gray}{(EM)}}                                                            & 83.6 & 83.0 & 84.9   \\
MMLU-Pro~{\footnotesize\textcolor{gray}{(EM)}}                                                             & 74.5 & 74.4  & 76.0   \\
GPQA Diamond~{\footnotesize\textcolor{gray}{(avg@8)}} & 64.2 & 63.5 & 68.3  \\
\midrule
\multicolumn{4}{l}{\textbf{Alignment}} \\
ArenaHard (GPT4-turbo-2024-04-09)                 & 71.7 & 70.0 & 76.9  \\
IFEval (strict prompt)~{\footnotesize\textcolor{gray}{(avg@8)}}                      & 66.3 & 70.8 & 69.8   \\
IFBench~{\footnotesize\textcolor{gray}{(avg@8)}}                                   & 23.2 & 21.2 & 24.3  \\
\midrule
\multicolumn{4}{l}{\textbf{Math}} \\
AIME 2024~{\footnotesize\textcolor{gray}{(avg@64)}}   & 83.8 & 83.6 & 86.9   \\
AIME 2025~{\footnotesize\textcolor{gray}{(avg@64)}}   & 71.6 & 72.8 & 81.1   \\
\midrule
\multicolumn{4}{l}{\textbf{Code}} \\
LiveCodeBench v5 (08/24-02/25)~{\footnotesize\textcolor{gray}{(avg@8)}}    & 59.6 & 59.2 & 66.1  \\
LiveCodeBench v6 (08/24-05/25)~{\footnotesize\textcolor{gray}{(avg@8)}}    & 56.7 & 56.7 & 63.1  \\
LiveCodeBench Pro 25Q2 (Easy)~{\footnotesize\textcolor{gray}{(avg@8)}} & 48.5 & 45.1 & 57.7 \\
LiveCodeBench Pro 25Q2 (Med)~{\footnotesize\textcolor{gray}{(avg@8)}} & 3.1 & 2.6 & 7.0 \\
SWE-bench Verified~{\footnotesize\textcolor{gray}{(avg@4)}}  & 30.2 & 26.1 & 34.5 \\
% \midrule
% \multicolumn{4}{l}{\textbf{Tool calling}} \\
% BFCL V3~{\footnotesize\textcolor{gray}{(avg@4)}}     & 66.1 & 60.4 & 66.83 & -- \\
\bottomrule
\end{tabular}
% \end{adjustbox}
\end{table}

\section{Cascade RL}
\label{sec:rl}

In this section, we describe the proposed Cascaded Reinforcement Learning~(Cascade RL) method.
In curating the RL data, we ensure that the \textbf{SFT and RL datasets are strictly disjoint in terms of prompts}, so the model cannot leverage memorized answers for given prompts from SFT during RL training.

\subsection{Training Framework}
As illustrated in Figure~\ref{fig:training_pipeline}, the Cascade RL process begins with applying general-domain Reinforcement Learning from Human Feedback~(RLHF) to the SFT models described in \S~\ref{subsec:rlhf}, followed by domain-wise Reinforcement Learning with Verifiable Rewards~(RLVR). We first apply RLHF and then RLVR~(e.g., Math RL), as RLHF substantially improves the quality of generated responses by reducing \emph{verbosity} and \emph{repetition}, thereby enhancing the reasoning performance within constrained response lengths (e.g., 64K tokens).
In Cascade RL, we sequentially apply RLHF~(\S~\ref{subsec:rlhf}), Instruction-Following RL~(\S~\ref{subsec:rl_ifeval_ifbench}), Math RL~(\S~\ref{subsec:math_rl}), Code RL~(\S~\ref{subsec:code_rl}), and finally SWE RL~(\S~\ref{subsec:swe_rl}), progressively from more general domains towards more specialized ones.

\subsubsection{Why Cascade RL for LLMs Is Resistant to Catastrophic Forgetting}
\label{subsec:rl_forgetting}

\emph{Catastrophic forgetting} occurs when a model trained sequentially on multiple domains overwrites previously learned knowledge while acquiring new ones, a common issue in supervised learning, where disjoint training datasets cause updates to push the model toward the new distribution.

Cascaded Cross-Domain RL for LLMs differs in several structural ways that mitigate this issue:
\begin{itemize}
\item 
\textbf{\emph{i}}) In RL, the training data distribution is policy-dependent; the LLM generates its own experience.
When a new objective or task is introduced, the LLM still explores across states, meaning old behaviors are continuously sampled if they remain useful or high-reward.
This contrasts with supervised learning~(e.g., SFT), where the samples in the previous domain disappear unless explicitly replayed.
\item 
\textbf{\emph{ii}}) RL optimizes expected cumulative reward, not exact targets for each input.
As a result, the updates focus on improving long-term outcomes rather than explicitly fitting a new token-level distribution.
Old knowledge that remains reward-relevant naturally persists.
When new tasks share structure with old ones, updates tend to generalize rather than overwrite.
\item 
\textbf{\emph{iii}})~\emph{Catastrophic forgetting} may still occur when the reward of a new domain sharply conflicts with that of a previous one (e.g., optimizing for concise responses versus detailed, step-by-step reasoning), particularly when prompts from different domains are semantically similar.
However, the reward structures of RLHF and RLVR overlap substantially across domains, such as math, code, reasoning, and instruction following, since they all aim to make outputs better, more accurate, and more aligned with human preferences or verification signals. For instance, reducing verbosity or hallucination generally benefits all domains.
\item 
\textbf{\emph{iv}})~
In our Cascade RL framework, we further minimize prompt overlap to the greatest extent possible, given that prompts across domains are generally already distinct. For instance, we remove all math and competitive programming–related prompts from the RLHF stage to reduce cross-domain interference.
Furthermore, domain-wise RL is organized from more general domains~(e.g., RLHF, instruction-following) to more specialized ones~(e.g., math, code, SWE), preventing specialized capabilities from being overwritten by generic behaviors.
\end{itemize}

\subsubsection{RL Training Configuration}
\label{subsec:rl_config}
Throughout the entire Cascade RL process, we use Group Relative Policy Optimization (GRPO) algorithm~\citep{shao2024deepseekmath} with strict \textbf{on-policy} training following AceReason-Nemotron~\citep{chen2025acereason}.
We adopt on-policy training for improved stability and higher accuracy.
We conduct our training using the verl repository~\citep{sheng2025hybridflow}.

At each iteration, we generate a group of $G$ rollouts from the current policy $\pi_{\theta}$ and then perform a \emph{single} gradient update. This ensures that the policy used for data collection always matches the one being updated, making the importance sampling ratio exactly 1. This on-policy setup contributes to stable RL training and mitigates entropy collapse. In addition, we {remove KL divergence} term entirely, which simplifies the GRPO objective to the standard \emph{REINFORCE} objective~\citep{williams1992simple} with group-normalized rewards and {token-level} loss~\citep{yu2025dapo}:
%
% \begin{equation}
\begin{align}
\mathcal{J}_\text{GRPO}(\theta)& = \mathbb{E}_{(q,a)\sim \mathcal{D},~ \{o_i\}_{i=1}^G\sim \pi_{\theta}(\cdot\mid q)}
\Bigg[
\frac{1}{\sum_{i=1}^{G}|o_i|}\sum_{i=1}^{G}\sum_{t=1}^{|o_i|}
\hat{A}_{i,t}
\Bigg], 
~\text{where}~
    \hat{A}_{i,t} = \frac{r_i - \text{mean}(\{r_i\}_{i=1}^G)}{\text{std}(\{r_i\}_{i=1}^G)} ~\text{for all } t,
\label{equation:grpo_objective}
\end{align}
% \nonumber 
% \end{equation}
%
and $\{r_i\}_{i=1}^{G}$ denotes the group of G rewards assigned to the sampled responses $\{o\}_{i=1}^G$ for a given question $q$ drawn from the dataset $\mathcal{D}$, verified against the ground-truth answer $a$ in RLVR~(e.g., Math RL).
For RLHF, $r_i$ is the scalar-output from the reward model for response $o_i$ and question $q$.
Details of the reward functions for different domains will be provided in the corresponding subsections.

\subsection{Reward Modeling}
\label{subsec:reward_model}
In this subsection, we describe the construction of the reward model~(RM) used in the RLHF phase presented later in \S~\ref{subsec:rlhf}.
\subsubsection{Data Curation}
Our reward modeling preference dataset is a mixture of open-source and in-house data, comprising a total of 82K preference pairs. % exact 81808
Specifically, we use the following open-source data:
\begin{itemize}
    \item HelpSteer2~\citep{wang2024helpsteer2} — a 10K high-quality, human-annotated preference dataset with multi-aspect annotations covering helpfulness, correctness, coherence, complexity, and verbosity.
    \item HelpSteer3~\citep{wang2025helpsteer3preferenc}   — a 40K preference dataset spanning multiple domains, including general, STEM, code, and multilingual.
    Each sample~(pair of response) is annotated with a preference score ranging from –3 (Response 1 is much better than Response 2) to 3 (Response 2 is much better than Response 1). We filter out samples with a score of 0 (Response 1 is similar to Response 2), resulting in 36K remaining samples.
    %
    % \item OffsetBias~\citep{park2024offsetbias} — an 8K pairwise preference dataset designed to mitigate biases in judge models (e.g., a tendency to prefer longer and more fluent but incorrect generations over shorter, less fluent yet accurate ones).
    %
    % \item cha~\citep{wildguard2024} — an safety preference dataset covering various categories of harmful content, including discrimination, abuse, violence, self-harm, sexual content, and misinformation. It consists of both vanilla and adversarial prompts that span harmful and benign scenarios. We use the filtered subset from~\citet{liu2024skywork} which is 6709 samples.
\end{itemize}

Inspired by~\citet{park2024offsetbias}, we generated additional data to improve our final preference data blend. The core idea is to construct preference pairs with bad responses from the stronger LLM and good responses from the weaker one. Note that, inducing the stronger LLM to generate bad responses is crucial for making the data effective; otherwise, the preference pairs would be too easy for the reward model to distinguish.

One specific approach is to generate a slightly off-topic prompt and use it to obtain a bad response from the stronger model. DeepSeek-V3 was employed to produce these off-topic prompts by rewriting the original prompt, and the quality of the rewrites was verified to be high through both manual inspection and automatic evaluation using LLM-as-a-Judge.
For detailed prompts, please refer to Appendix~\S\ref{appendix:rm_data_gen_template}.
We explored different combinations of LLMs as the weaker and stronger models and ultimately selected DeepSeek-V3-0324 and DeepSeek-V3 as the stronger and weaker models, respectively. 
We also tried explicitly instructing strong LLMs to produce subtly erroneous answers to the given prompts; however, this approach was unsuccessful.

\subsubsection{Training Recipe}
We train a scalar-output reward model~(RM) on pairwise human preference data using the Bradley–Terry objective~\citep{bradley1952rank}:
\[
P_{\text{ BT}}(y^+ \succ y^- \mid x) = 
\frac{\exp(r_\theta(x, y^+))}
     {\exp(r_\theta(x, y^+)) + \exp(r_\theta(x, y^-))}
\]
where $y^+$ is preferred and $y^-$ is dispreferred.
Our reward model is initialized with Qwen2.5-72B-Instruct~\citep{yang2024qwen2_5} with a linear predictor on top of its last hidden layer, and is trained by maximizing the log-likelihood of human preferences:
\[
\mathcal{L}(\theta) 
= \mathbb{E}_{(x, y^+, y^-) \sim \mathcal{D}} 
\left[ 
    \log P_{\text{ BT}}(y^+ \succ y^- \mid x) 
\right]
\]
For each prompt, two responses—one preferred and one dispreferred—are compared in a contrastive manner, enabling the RM to learn to assign higher scalar scores to preferred responses and lower scores to dispreferred ones. We treat this scalar score as a proxy metric for the “quality” of the model’s response. The training hyperparameters are as follows: batch size 256, learning rate 2e-6, AdamW optimizer~\citep{loshchilov2017decoupled}, and 1 epoch. Note that we also experimented with longer training schedules, but found that a single epoch yielded the best results.

\paragraph{RM evaluation:}
In this study, we primarily used RewardBench~\citep{lambert2024rewardbench} to evaluate and select our reward model~(RM) for the RLHF process. In practice, we found that reward models with low RewardBench scores typically lead to poor policy alignment after RLHF. However, the RM with the highest RewardBench score does not necessarily yield the best aligned policy model (i.e., as measured by alignment benchmark), since RewardBench can be an imperfect proxy for identifying the optimal reward model for RLHF, and the RLHF process itself introduces additional variance.
There are ongoing research efforts to establish robust RM benchmarks that can serve as more reliable proxy metrics for identifying reward models likely to produce the best final alignment. However, there is still no general consensus on a standard benchmark to use.

\paragraph{Ablation studies on RM:}

We conducted ablation studies to determine the choice of backbone LLM, and our key findings are as follows:
\begin{itemize}
    \item 
    \textit{Model Size}: We trained reward models of various sizes using the Qwen2.5-Instruct series (7B, 14B, 32B, and 72B) and observed that performance scales positively with model size, confirming that the scaling law~\citep{kaplan2020scaling} holds in the context of reward model training as well (see \textit{(a)-(d)} in Table~\ref{table:rlhf_rm_ablation}). Larger LLMs demonstrated greater robustness to stylistic artifacts in the preference data, whereas smaller models tended to focus more on the style of a response rather than its overall quality. This observation was further validated by the Arena-Hard scores~\citep{li2024crowdsourced} when applying the reward models in RLHF, both with and without style control: smaller models exhibited a larger performance drop under style-controlled evaluation compared to their larger, as futher discussed in  \S~\ref{abl:rmsize}.
    \item 
    \textit{With vs. Without Large-scale Preference Pre-training}: \citet{WorldPM} released the WorldPM checkpoint, a Qwen2.5-72B model further pre-trained on 15M-scale diverse preference data, which can serve as a strong initialization for reward model training. In our experiments, models initialized from WorldPM perform better during the early stages of training; however, with extended training, models initialized from the vanilla Qwen2.5-72B-Instruct eventually catch up and slightly outperform them on RewardBench~\citep{lambert2024rewardbench} - see {(d)} vs {(e)} in Table~\ref{table:rlhf_rm_ablation}.
    \item 
    \textit{Reasoning vs. Instruct Models}: During the development of our post-training pipeline, the Qwen3 unified reasoning models were released~\citep{yang2025qwen3}, prompting us to explore the Qwen3 8B and 14B checkpoints as well. However, we empirically found that the Qwen3 reasoning models consistently underperformed their Qwen2.5 \emph{instruct}/\emph{non-thinking} counterparts of the same size when used as the backbone for reward models trained with BT loss (see (b) vs (f) in Table~\ref{table:rlhf_rm_ablation}). We suspected this was due to our preference data being less suitable for Qwen3 models operating in \emph{thinking} mode, so we conducted additional experiments with the \emph{non-thinking} mode enabled. Although performance improved considerably, the Qwen3 models in \emph{non-thinking} mode still failed to surpass Qwen2.5-Instruct—their dedicated \emph{non-thinking} counterparts (refer to (b) vs (g) in Table~\ref{table:rlhf_rm_ablation}). We hypothesize this is because the Qwen3 reasoning models are primarily optimized for reasoning-centric tasks (e.g., math and code) rather than general human preference alignment.
\end{itemize}

\begin{table}[t]
\centering
\footnotesize
\captionsetup{justification=centering}
\caption{RewardBench~\citep{lambert2024rewardbench} performance showing ablation of RM's LLM backbone choice.}
\begin{adjustbox}{width={0.7\textwidth}}
\begin{tabular}{clccccc}
\toprule
& \textbf{LLM Backbone} & \textbf{Overall} & \textbf{Chat} & \textbf{Chat Hard} & \textbf{Safety} & \textbf{Reasoning} \\ \midrule
(a) & Qwen2.5-Instruct-7B & 91.98 & 96.65 & 85.09 & 90.54 & 95.64 \\
(b) & Qwen2.5-Instruct-14B & 93.22 & 96.93 & 87.72 & 91.62 & 96.63 \\
{(c)} & Qwen2.5-Instruct-32B & 93.56 & 96.37 & 89.04 & 91.76 & 97.09 \\
{(d)} & Qwen2.5-Instruct-72B & \textbf{95.15} & \textbf{98.60} & \textbf{89.69} & \textbf{93.92} & \textbf{98.40} \\ \midrule
{(e)} & WorldPM-72B & 94.08 & 97.77 & 86.40 & 93.78 & 98.36 \\ \midrule
{(f)} & Qwen3-14B (thinking) & 46.38 & 29.05 & 46.27 & 60.95 & 49.24 \\ 
{(g)} & Qwen3-14B (non-thinking) & 91.72 & 94.69 & 87.94 & 89.73 & 94.52 \\
% (g) & Qwen3-Instruct-14B (half think) & 89.19 & 94.41 & 80.70 & 89.46 & 92.19 \\
\bottomrule
\end{tabular}
\end{adjustbox}
\label{table:rlhf_rm_ablation}
\end{table}

% \textbf{TODO: try running RewardBench2 -- see if the numbers are worth to include.}

\subsection{Reinforcement Learning from Human Feedback~(RLHF)}
\label{subsec:rlhf}
In this subsection, we describe our RLHF recipe, which serves as the first stage of the Cascade RL process.
We find that the generalization capability of the reward model plays a crucial role in ensuring stable RLHF training, and that larger reward models (e.g., 72B RM) are more resilient to out-of-distribution~(OOD) samples generated by the policy LLM.

\subsubsection{Data Curation}
We find that introducing OOD prompts of the reward model to the RLHF stage often leads to instability or even training collapse due to inaccurate or misleading reward signals.
Therefore, during the RLHF stage, we use a subset of prompts from the reward model’s preference dataset described in \S~\ref{subsec:reward_model}.
Additionally, we exclude prompts related to mathematics and competitive programming, since the reward model may not provide reward signals as reliable as those produced by rule-based or execution-based verifiers used in subsequent Math RL and Code RL stages.
For example, in our early experiments, we observed that failing to exclude math-related prompts during RLHF resulted in a 2\% performance drop on the AIME25 benchmark.
As a result, our RLHF dataset primarily focuses on improving helpfulness, harmlessness, and alignment with human preferences, while remaining disjoint from the domains that will be enhanced in the subsequent Cascade RL stages.

\subsubsection{Training Recipe}
\label{subsec:rlhf_recipe}

RLHF helps LLMs better follow user intentions and align with human preferences.
We also observe that RLHF improves overall generation quality and, interestingly, enhances reasoning performance on math and code benchmarks, despite our curated RLHF datasets containing no math or code-related prompts.
Moreover, RLHF tends to reduce repetition and verbosity, thereby compressing the number of thinking tokens for simpler questions. This, in turn, enhances reasoning efficiency and training stability in the subsequent Math RL and Code RL stages.
Therefore, we design our cascade RL pipeline to begin with the RLHF stage.
Our RLHF training initializes from the SFT checkpoint, employs the GRPO algorithm, and follows the unified RL training configuration in~\S~\ref{subsec:rl_config}~(e.g., on-policy, token-level loss, no KL divergence).
For dedicated \emph{thinking} models, we naturally perform RLHF in \emph{thinking} mode.
For \emph{unified} models, we perform RLHF training in both \emph{non-thinking} and \emph{thinking} modes, with an equal split of prompts allocated to each mode within every batch. We provide further studies in \S~\ref{abl:rlhf_unified}.

\paragraph{Reward function:}
RLHF training uses the reward scores produced by the reward model as the reward function.
Specifically, we extract the model’s answer, concatenate it with the corresponding question, apply the reward model’s chat template, and feed the formatted input into the reward model to obtain a point-wise reward score.
We handle answer extraction differently for LLMs operating in the \textit{non-thinking} mode and \textit{thinking} mode:
for the \textit{non-thinking} mode, we directly extract the model answer following the assisant role; 
for the \textit{thinking} mode, we exclude the reasoning traces and only extract the final summary after the thinking process (\textit{i.e.}, model generation after the \texttt{<\textbackslash think>} token). 
If the thinking process does not terminate properly (\textit{i.e.}, the \texttt{<\textbackslash think>} token is missing), we fall back to sending the entire unfinished response to the reward model.
Such incomplete generations typically receive low reward scores because the reward model was not trained on unfinished or unseen reasoning traces, effectively penalizing verbose or incomplete thinking processes. During training, we use a maximum response length of 12K for both the 8B and 14B models in RLHF, without applying overlong filtering, which encourages more succinct generations.

% \paragraph{Code-switch Penalty:}
To prevent language mixing in the generation, we apply an additional \textit{code-switch penalty} when the prompt is purely in English but the generated response, including both thinking traces and summary, contains non-English tokens.
As the reward scores from the reward model are unbounded,we adaptively assign the reward for mixed-language generations to be the lowest score in the batch minus 10, ensuring they receive the lowest relative score under the GRPO algorithm and thus the strongest penalty for code-switching behavior.
We do not apply additional reward shaping techniques, as the reward signals from our 72B reward model are already of high quality.

\paragraph{Hyperparameters:}
For both our 8B and 14B models, we use a maximum response length of 12K during RLHF without applying overlong filtering, which effectively encourages more succinct generations.
We use a batch size of 128, generating 8 rollout per prompt with a temperature of 0.6 and a top-p value of 0.95.
We adopt a learning rate of 2e-6 with AdamW~\citep{kingma2014adam}, and set both the entropy loss coefficient and KL loss coefficient to 0. 
The training takes around 800 steps.
More details of training hyperparamters can be found in Appendix~\ref{appendix:training_hyperparams}.

\subsubsection{Results after RLHF}
After RLHF, the results of our 8B and 14B models are shown in Table~\ref{tab:results_after_rlhf}.
One can observe significant improvements on nearly every benchmark, except for IFEval.
The main reason is that our RLHF process substantially improves response quality by penalizing overly long, verbose, and repetitive generations, especially for \emph{thinking} mode.
For the degraded IFEval performance, the primary cause is the unavoidable semantic overlap between the prompts used during RLHF training and the test prompts in IFEval. Additionally, the reward model used in RLHF encourages human-preferred response qualities that can conflict with the strict instruction-following constraints evaluated by the verifier.
We believe this issue can be mitigated by training a stronger reward model (e.g., a large generative RM~\citep{wang2025RLBFF}) capable of handling strict instruction-following constraints as well. We leave this for future work.
We will revisit this in the next subsection.

\begin{table}[t!]
\centering
\footnotesize
\renewcommand{\arraystretch}{1.15}
\caption{The evaluation results of our 8B/14B-Thinking and the \emph{unified} 8B models \textbf{after RLHF} are presented below.  For unified model, we evaluate IFEval in the non-thinking mode and all other benchmarks in the thinking mode. We compare the results against those obtained after SFT in Table~\ref{tab:results_after_sft}, using \ua{} to denote improvements and \da{} to mark degradations after RLHF training.
}
\label{tab:results_after_rlhf}
\begin{tabular}{lccc}
\toprule
\shortstack[l]{\textbf{Benchmark}\\\textbf{Metric: pass@1}}
 & \shortstack{\textbf{8B-Thinking}\\\textbf{RLHF}}
 & \shortstack{\textbf{8B (\emph{unified})}\\\textbf{RLHF}} 
  & \shortstack{\textbf{14B-Thinking}\\\textbf{RLHF}}
 \\
\midrule
\multicolumn{4}{l}{\textbf{Knowledge and Reasoning}} \\
MMLU~{\footnotesize\textcolor{gray}{(EM)}}                                                       &
83.9~\ua{0.3} & 83.1~\ua{0.1} & 85.2~\ua{0.3}  \\
MMLU-Pro~{\footnotesize\textcolor{gray}{(EM)}}                                                             & 76.2~\ua{1.7} & 77.8~\ua{3.4} & 77.7~\ua{1.7}  \\
GPQA Diamond~{\footnotesize\textcolor{gray}{(avg@8)}} & 67.3~\ua{3.0} & 66.8~\ua{3.3} & 71.3~\ua{3.0} \\
\midrule
\multicolumn{4}{l}{\textbf{Alignment}} \\
ArenaHard (GPT4-turbo-2024-04-09)                 & 
89.9~\ua{18.2} & 90.1~\ua{20.1} & 93.1~\ua{19.2}  \\
IFEval (strict prompt)~{\footnotesize\textcolor{gray}{(avg@8)}}             &
45.5~\da{20.8} & 50.1~\da{20.7} & 56.2~\da{12.4} \\
IFBench~{\footnotesize\textcolor{gray}{(avg@8)}}                                &
23.9~\ua{0.7} & 24.5~\ua{3.3} & 25.6~\ua{1.9}  \\
\midrule
\multicolumn{4}{l}{\textbf{Math}} \\
AIME 2024~{\footnotesize\textcolor{gray}{(avg@64)}}   &
86.4~\ua{2.6} & 86.1~\ua{2.5} & 88.4~\ua{1.5}  \\
AIME 2025~{\footnotesize\textcolor{gray}{(avg@64)}}   & 
75.1~\ua{3.5} & 75.0~\ua{2.2} & 81.8~\ua{0.7}  \\
\midrule
\multicolumn{4}{l}{\textbf{Code}} \\
LiveCodeBench v5 (08/24-02/25)~{\footnotesize\textcolor{gray}{(avg@8)}}    & 
70.3~\ua{10.7} & 70.2~\ua{11.0} & 75.2~\ua{9.1}   \\
LiveCodeBench v6 (08/24-05/25)~{\footnotesize\textcolor{gray}{(avg@8)}}    & 
67.3~\ua{10.6} & 67.2~\ua{10.5} & 72.3~\ua{9.2}  \\
SWE-bench Verified~{\footnotesize\textcolor{gray}{(avg@4)}}  & 33.3~\ua{3.1} & 28.2~\ua{2.1} & 38.8~\ua{4.3}  \\
% \midrule
% \multicolumn{4}{l}{\textbf{Tool calling}} \\
% BFCL V3~{\footnotesize\textcolor{gray}{(avg@4)}}     & 67.7 & ?? & 67.2 & -- \\
\bottomrule
\end{tabular}
\end{table}

\subsection{Instruction-Following Reinforcement Learning~(IF-RL)}
\label{subsec:rl_ifeval_ifbench}
Verifiable instruction following is a crucial aspect of ensuring that LLMs can follow human instructions precisely.
Although our SFT data blend already contains instruction-following data, applying IF-RL with verifiable rewards further enhances the accuracy of instruction adherence.
\subsubsection{Data Curation}
    % ifeval = apache
    % ifbench = ODC-BY-1.0
    % IF-RLVR = ODC-BY-1.0
    % LMSYS-Chat-1M - commericial
We use the instruction-following dataset from Llama-Nemotron~\citep{LlamaNemotronPostTrainingDataset}, which consists of synthetically generated prompts containing one to ten detailed instruction constraints derived from the IFEval taxonomy~\citep{zhou2023instructionfollowingevaluationlargelanguage}. However, we found this dataset to be noisy due to its synthetic nature. To improve overall quality, we performed extensive preprocessing and filtering, reducing 56K samples to 40K high-quality ones.
We additionally curate 60K custom data samples to enhance the diversity of our data blend, using user prompts from LMSYS-Chat-1M~\citep{zheng2023lmsyschat1m} paired with various instruction constraints from the IFEval taxonomy.
We also incorporate the IF-RLVR training data from \citet{pyatkin2025generalizingverifiableinstructionfollowing}, which is designed to enhance robustness to unseen constraint taxonomies. This dataset comprises prompts paired with instruction constraints drawn from either the IFEval taxonomy or the IF-Bench-Train taxonomy~\citep{pyatkin2025generalizingverifiableinstructionfollowing}, with base prompts sampled from Tulu-3-SFT~\citep{lambert2025tulu3pushingfrontiers}.

\subsubsection{Training Recipe}
The IF-RL training proceeds in two stages, each utilizing a distinct data blend with progressively increasing difficulty.  The first stage focuses on instruction constraints from IFEval taxonomy, while the second stage focuses on IF-Bench-Train taxonomy.  We find the dynamic filtering~\citep{yu2025dapo} largely stabilize the IF-RL training and improve the results at both stages by ensuring all prompts in the batch with effective gradients.

One of the major challenges in the IF-RL stage was mitigating the negative impact that IF-RL could have on the human alignment capabilities acquired during the RLHF stage~(e.g., measured by ArenaHard). In our early experiments, we observed that naively using a rule-based IF verifier as the reward function  degraded human alignment results. This occurs because the rule-based IF verifier focuses solely on whether the response adheres to the constraints specified by the instruction, without considering the overall response quality. For example, a poorly written answer to the prompt “write a summary within 300 words” could still receive a full reward as long as its word count remains below 300.

\paragraph{\emph{Unified} models: IF-RL in the \emph{non-thinking} mode}
An effective strategy emerges for the \emph{unified} reasoning model: we first perform RLHF in both \emph{thinking} and \emph{non-thinking} modes, and then apply IF-RL only in the \emph{non-thinking} mode. This approach minimizes negative mutual interference between RLHF and IF-RL, while still yielding substantial improvements in the model’s instruction-following capability in the \emph{thinking} mode~(i.e., our 8B unified model achieves IFEval 85.3 in thinking mode).
We hypothesize that applying IF-RL to an RLHF-trained model in the \emph{non-thinking} mode is far less likely to generate low-quality responses than applying it in the \emph{thinking} mode, and therefore is much less prone to reward-hacking the rule-based IF verifier.
We also experimented with reversing the order of RLHF and IF-RL, but observed much worse results.
At both first-stage and second-stage IF-RL training, we set the maximum response length to 8K tokens and do not apply overlong filtering for unified reasoning model.

\paragraph{\emph{Thinking model}: IF-RL with combined reward function}
Another approach is to design a reward function in IF-RL that jointly accounts for both human preference and precise instruction-following capability.
For dedicated \emph{thinking} models, this is crucial for mitigating the negative impact of IF-RL on benchmarks such as ArenaHard~\citep{li2024crowdsourced}.

We combine signals from both the rule-based instruction-following verifier and the human preference reward model, achieving the best of both worlds. 
For a given prompt $q$ and a group of generated response $\{o_i\}_{i=1}^G$, the reward for each response $o_i$ is defined as,
\[
r_i = \begin{cases}
R_{\text{ IF}}(o_i) + \text{sigmoid}(\hat{R}_{\text{ RM}}(o_i)), & \text{if } R_{\text{ IF}}(o_i) = 1 \\
0, & \text{otherwise}
\end{cases}, 
\text{therein~} \hat{R}_{\text{ RM}}(o_i)
= \frac{R_{\text{ RM}}(o_i) - \text{mean}(\{R_{\text{ RM}}(o_i) \}_{i=1}^G)}{\text{std}(\{R_{\text{ RM}}(o_i) \}_{i=1}^G)} 
\]
where $R_{\text{ IF}}(o_i) \in \{0, 1\}$ refers to the binary reward from instruction-following verifier, $\hat{R}_{\text{ RM}}(o_i)$ is the group-normalized reward~(mean = 0, standard deviation = 1) from the same reward model used in the RLHF stage.
The sigmoid function is applied to  $\hat{R}_{\text{ RM}}$ to scale its values to the range $(0, 1)$, ensuring it is on the same scale as  $R_{\text{ IF}}$ before aggregation.

We then perform IF-RL using the same GRPO objective described in \S~\ref{subsec:rl_config}, incorporating the combined reward. 
For the dedicated \emph{thinking} model, we set the maximum response length to 8K tokens with overlong filtering during the first-stage IF-RL training, and increase it to 16K tokens with overlong filtering in the second stage to accommodate the longer reasoning required for difficult prompts in \emph{thinking} mode.

\paragraph{Hyperparameters:}
For both 8B and 14B models,  we use a batch size of 128, sampling 8 responses per prompt with temperature 0.6, top-p 0.95, and top-k 20.
We adopt a learning rate of 2e-6 with AdamW~\citep{kingma2014adam}, and set both the entropy loss coefficient and KL loss coefficient to 0. 
For \textit{non-thinking} mode IF-RL, the first-stage training takes around 2000 steps, and the second-stage training takes 1000 steps.
For \textit{thinking} mode IF-RL, the first-stage training takes around 500 steps, and the second-stage training takes around 300 steps.
More details of training hyperparamters can be found in Appendix~\ref{appendix:training_hyperparams}.

\subsubsection{Results after IF-RL}
The results after IF-RL are presented in Table~\ref{tab:results_after_ifrl}. We observe significant improvements on IFEval and IFBench, with controlled small degradation on ArenaHard when applying the improved techniques to both the \emph{unified} models and the dedicated \emph{thinking} model. 
We also find that IF-RL generally reduces model entropy and shortens the average length of reasoning tokens (see Figure~\ref{fig:ablation_coderl_cascade} for an illustration). On the negative side, this introduces minor degradations on reasoning benchmarks, although most of them (except ArenaHard) are fully recoverable and are further improved after the subsequent Math RL, Code RL, and SWE RL stages.
On the positive side, it compresses the reasoning trace and improves token efficiency. Overall, the \emph{unified} reasoning model achieves a stronger balance than dedicated \emph{thinking} model, delivering robust performance on both ArenaHard and IFEval.

\begin{table}[t!]
\centering
\footnotesize
\renewcommand{\arraystretch}{1.15}
\caption{The evaluation results of 8B/14B-Thinking and the \emph{unified} 8B models \textbf{after IF-RL} are presented below. For unified model, we evaluate IFEval in the non-thinking mode and all other benchmarks in the thinking mode.
We compare the results against those obtained after RLHF in Table~\ref{tab:results_after_rlhf}, using \ua{} to denote improvements and \da{} to mark degradations after IF-RL training.
}
\label{tab:results_after_ifrl}
\begin{tabular}{lccc}
\toprule
\shortstack[l]{\textbf{Benchmark}\\\textbf{Metric: pass@1}}
 & \shortstack{\textbf{8B-Thinking}\\\textbf{IF-RL}}
 & \shortstack{\textbf{8B (unified)}\\\textbf{IF-RL}} 
  & \shortstack{\textbf{14B-Thinking}\\\textbf{IF-RL}}
 \\
\midrule
\multicolumn{4}{l}{\textbf{Knowledge and Reasoning}} \\
MMLU~{\footnotesize\textcolor{gray}{(EM)}}                                                       &
83.8~\da{0.1} & 83.4~\ua{0.3} & 85.0 \da{0.2}  \\
MMLU-Pro~{\footnotesize\textcolor{gray}{(EM)}} & 74.8~\da{1.4} & 74.5~\da{3.3} & 76.4~\da{1.3}   \\
GPQA Diamond~{\footnotesize\textcolor{gray}{(avg@8)}} & 65.2~\da{2.1} & 66.1~\da{0.7} & 70.1~\da{1.2}  \\
\midrule
\multicolumn{4}{l}{\textbf{Alignment}} \\
ArenaHard (GPT4-turbo-2024-04-09)                 & 
86.3~\da{3.6} & 88.0~\da{2.1} & 90.2 \da{2.9}  \\
IFEval (strict prompt)~{\footnotesize\textcolor{gray}{(avg@8)}}             &
83.3~\ua{37.8} & 90.4~\ua{40.3}   & 81.3 \ua{25.1}  \\
IFBench~{\footnotesize\textcolor{gray}{(avg@8)}}                               &
42.1~\ua{18.2} & 40.5~\ua{16.0} & 40.4 \ua{14.8}  \\
\midrule
\multicolumn{4}{l}{\textbf{Math}} \\
AIME 2024~{\footnotesize\textcolor{gray}{(avg@64)}}   &
85.6~\da{0.8} & 86.2~\ua{0.1} & 89.2 \ua{0.8}  \\
AIME 2025~{\footnotesize\textcolor{gray}{(avg@64)}}   & 
72.3~\da{2.8} & 75.2~\ua{0.2} & 82.3 \ua{0.5}  \\
\midrule
\multicolumn{4}{l}{\textbf{Code}} \\
LiveCodeBench v5~{\footnotesize\textcolor{gray}{(avg@8)}}    & 
69.0~\da{1.3} & 70.2~\ua{0.0} & 75.3~\ua{0.1}  \\
LiveCodeBench v6~{\footnotesize\textcolor{gray}{(avg@8)}}    & 
65.9~\da{1.4} & 66.7~\da{0.5} & 72.7~\ua{0.4}  \\
SWE-bench Verified~{\footnotesize\textcolor{gray}{(avg@4)}}  & 32.4~\da{0.9} & 28.3~\ua{0.1} & 38.4~\da{0.4}  \\
% \midrule
% \multicolumn{4}{l}{\textbf{Tool calling}} \\
% BFCL V3~{\footnotesize\textcolor{gray}{(avg@4)}}     & -- & -- & -- & -- \\
\bottomrule
\end{tabular}
\end{table}

\subsection{Math RL}\label{subsec:math_rl}
% @Yang
% \citep{chen2025acereason}
In this subsection, we describe the Math RL stage, which focuses on enhancing the model’s mathematical reasoning and problem-solving capabilities through reinforcement learning.
In our final model training, Math RL is applied after the Instruction-Following RL stage.
Applying Math RL directly to RLHF checkpoints yielded very similar results.

\subsubsection{Data Curation}
We mainly use the AceReason-Math dataset \citep{chen2025acereason} and filter out overly simple problems, retaining 18K high-quality math problems for RL training.
The dataset merges the DeepScaleR blend~\citep{deepscaler2025,gao2024omni,Slow_Thinking_with_LLMs_2} and NuminaMath~\citep{li2024numinamath}, encompassing topics such as algebra, geometry, combinatorics, and number theory.
We apply 9-gram filtering to prevent contamination with common math benchmarks, such as AIME 2024/2025 and MATH~\citep{hendrycks2021measuring}.
We further exclude questions unsuitable for RL with symbolic rule-based verification, such as multiple-choice or true/false questions (where answers can be easily guessed), proof-based problems (which are difficult to verify for correctness), those containing multiple sub-questions, non-English questions (which increase language mixing), and questions referencing figures.
Because NuminaMath contains OCR and parsing errors, each problem is verified by the DeepSeek-R1 model with up to eight attempts. A rule-based verifier retains only problems with majority-voted correct answers, while ambiguous or noisy items are discarded. Finally, we remove overly simple problems that AceReason-Nemotron-7B~\citep{chen2025acereason} can solve with a $\geq$ 75 \% success rate over 16 samples, reducing the dataset size from the original 49K to 14K problems.

\subsubsection{Training Recipe}

Our goal is to develop a general math RL recipe that can be applied across different base models and scaled efficiently for large-scale RL training. We build on the AceReason-Nemotron~\citep{chen2025acereason} training strategy, which strictly adheres to \textbf{on-policy training} under the GRPO objective, removes KL regularization entirely, and combines \textbf{length extension training} with \textbf{dynamic filtering} to stabilize optimization. 
We find that \textbf{initializing Math RL from RLHF-trained models} plays a crucial role in achieving better performance.
Throughout the development cycle, we applied this training recipe to five different 8B checkpoints and consistently achieved accuracies of around 90\% on AIME24 within 500 RL steps, demonstrating the robustness of the approach across models with different training dynamics. Below, we describe each component in detail.

\paragraph{Initialization from models that have undergone RLHF:}
In an early study, we explored an approach that first applied Math RL and Code RL, followed by RLHF and IF-RL.
Later, we found that initializing Math RL from RLHF-trained models is highly beneficial because it \emph{(i)} provides a much stronger initial math reasoning capability than SFT checkpoints—response quality is substantially improved, and reasoning becomes more token-efficient after RLHF (e.g., less verbosity and repetition); and \emph{(ii)} significantly reduces the number of steps required for math RL training.

In practice, we insert IF-RL between RLHF and Math RL, as IF-RL reduces model entropy and shortens the reasoning trace, which can temporarily hurt reasoning-related benchmarks. Applying Math RL and Code RL with high temperature after IF-RL restores the model entropy to a normal level.

\paragraph{Reward function:}
Rewards are assigned strictly based on answer correctness, which is determined by extracting the boxed answer (\texttt{\textbackslash\textbackslash boxed\{\}}) that follows the \texttt{<\textbackslash think>} token, and verifying it using the AceMath~\citep{liu2024acemath} rule-based verifier (1 for correct and 0 for incorrect). 
To prevent language mixing during the reasoning process, we apply a code-switching penalty by assigning a reward of $-1$ whenever tokens from a language~(e.g., Chinese) different from the original prompt’s language (e.g., English) are detected in the reasoning chain.

\paragraph{Response length extension training:} 
The key driver of performance improvement lies in the model’s ability to think more deeply and produce longer reasoning chains.
We adopt a staged response length extension curriculum with a custom configuration ($\text{24K} \to \text{32K} \to \text{40K}$) where each stage plays a distinct role: compressing overlong reasoning, stabilizing reasoning length, and finally extending longer reasoning chains, respectively. 
One key benefit of starting with the compression stage (i.e., 24K) is that it brings different initial models into a consistent reasoning length range (around 16K on the full training set), which enables subsequent training stages to work effectively across diverse initial models without extensive hyperparameter tuning.

\begin{itemize}
    \item \textbf{24K (Compression Stage).} 
    We begin by training with a 24K token budget to address a key issue observed in small and medium sized SFT checkpoints: these models tend to generate \emph{overlong reasoning chains}, leading to incomplete ratios of 15--20\% on the AIME benchmark under a 32K token budget. This overgeneration wastes tokens and often leaves solutions unfinished. By starting with a shorter 24K budget, we encourage the model to \emph{compress and refine} its reasoning. At this stage, models typically exhibit very high incomplete ratios (30--50\%) initially, but after around 100 steps of training, this drops to around 15\% on training set. Importantly, we deliberately apply the overlong filtering~(i.e., we skip generations exceeding 24K tokens rather than assigning a reward of 0), as doing so may excessively penalize long reasoning on difficult problems, causing a sharp performance drop during compression~\citep{liu2025acereason} and leading to unstable training due to noisy rewards in the high–incomplete-ratio regime.
    \vspace{0.05cm}
    \item \textbf{32K (Extension Stage).} 
    Once the reasoning chains are stabilized at 24K, we extend the token budget to 32K. However, checkpoints emerging from the 24K stage vary considerably in how efficiently they use tokens: some start the 32K phase with as little as 5\% incomplete ratio, while others hover around 10\%. This variability motivates treating the 32K phase as a \emph{controlled extension stage}. Here, we do not apply overlong filtering to regularize reasoning length to fit the 32K context~(i.e., assigning a reward of 0 for overlong generation). As training progresses, models not only adapt to the larger budget but also begin to surpass their starting accuracy, reflecting a balanced trade-off between length and correctness.
    \vspace{0.05cm}
    \item \textbf{40K (Long Reasoning Stage).} 
    After 32K training, model accuracy on easy and medium problems\footnote{We categorize each problem in AIME24/25 into easy (80-100\%)/medium (30-80\%)/hard (0-30\%) based on the problem accuracy of an early checkpoint.} nearly saturates (99\% and 85\% respectively) on AIME24/25, but hard problems remain challenging, with accuracy plateauing below 30\%. Since our evaluation is performed with a 64K token budget, we observed that models were not fully exploiting the available context even with YARN length extension (factor of 2). To address this gap, we push the model on a final 40K training stage. This extension explicitly incentivizes the model to leverage more tokens during reasoning. As a result, performance on hard AIME problems improves significantly from 30 to 40\%, while performance on other problems remains at high level.
\end{itemize}

\paragraph{Dynamic filtering:} 
To simplify development, we fix a seed dataset across all math RL experiments. However, since model capabilities vary, overly simple or unsolvable problems provide no useful policy gradient signal when using the group-normalized advantage function. To address this, after each epoch, we filter out problems that achieve either 100\% or 0\% accuracy based on the verification results from that epoch’s RL training. 
Hard problems that were filtered out are re-sampled into the dataset with a 10\% probability, as the policy may learn to solve these problems during subsequent updates within the same epoch.
Easy problems are re-sampled into the dataset with a 1\% probability to stabilize training, as the policy may forget how to solve them within an epoch.
This ensures that $\sim$90\% of training samples contribute meaningful learning signals and significantly stabilizes model accuracy during training, especially at later training stage when more problems are 100\% solved.
Note that this epoch-based dynamic filtering can be viewed as a more efficient alternative to batch-based dynamic sampling~\citep{yu2025dapo, xiaomi2025mimo}, as the latter requires substantially more rollouts to construct a fixed-size batch free of overly simple or unsolvable prompts.

\begin{figure}[!t]
    \centering
    \includegraphics[width=0.8\textwidth]{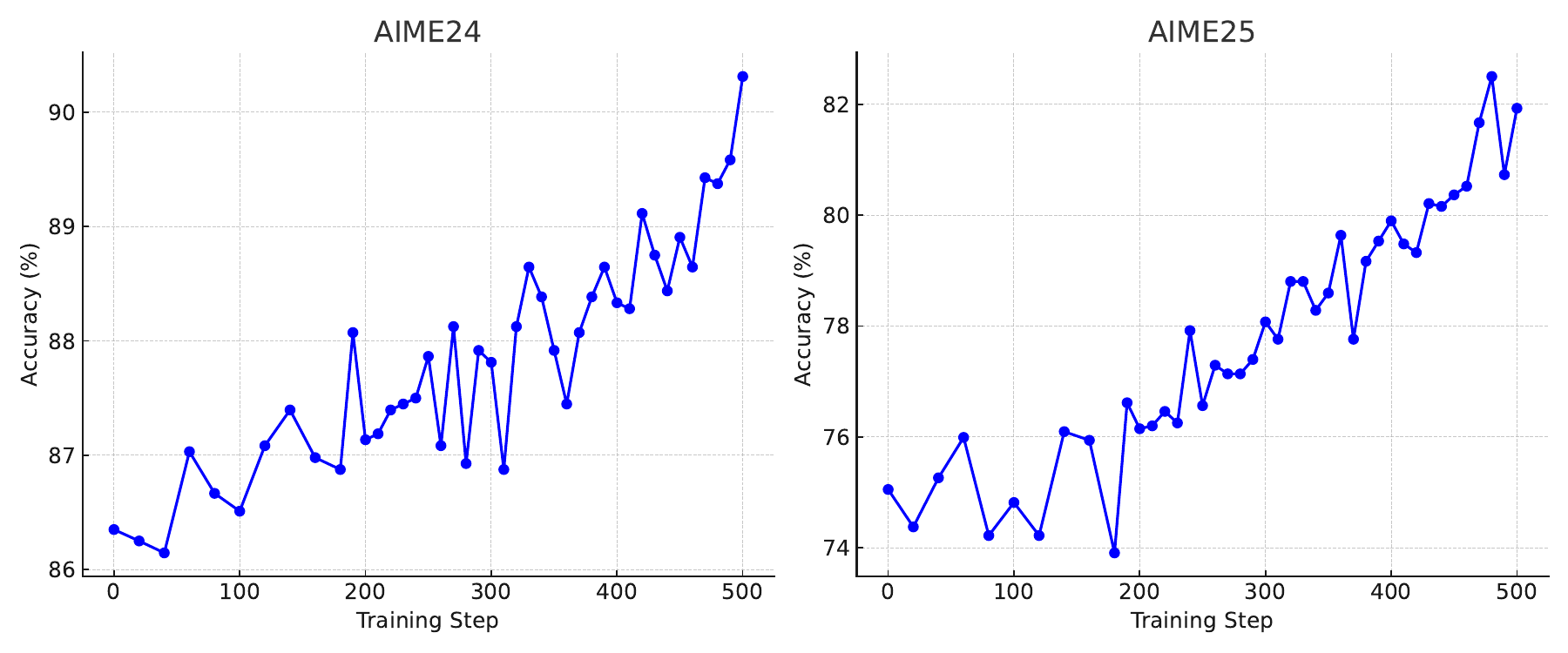}
    \caption{Training curve of math RL for 8B \emph{unified} model on AIME24 and AIME25 (max response length 64K, avg@64).}
    \label{fig:mathrl_results}
\end{figure}

\paragraph{Hyperparameters:}
We use a batch size of $128$, sampling 8 responses per prompt with temperature 1 and top-p 0.95.
We adopt a learning rate of $2 \text{ or } 2.5\times10^{-6}$ with AdamW~\citep{kingma2014adam}, and set both the entropy loss coefficient and KL loss coefficient  to $0$. Each stage of training takes around 100 to 200 steps depends on how fast the clip-ratio reaches 10\%. For 8B models, we adopt the 3 stage training with length extension from 24K $\to$ 32K $\to$ 40K. For the 14B model, as the initial policy already achieve high accuracy, we start with 28K max token length to avoid accuracy drop in the first stage and then extend to 40K directly.
More details of training hyperparamters can be found in Appendix~\ref{appendix:training_hyperparams}.

\begin{table}[t!]
\centering
\footnotesize
\renewcommand{\arraystretch}{1.15}
\caption{The evaluation results of 8B/14B-Thinking and the \emph{unified} 8B models \textbf{after Math RL} are presented below. For unified model, we evaluate IFEval in the non-thinking mode and all other benchmarks in the thinking mode.
We compare the results against those obtained after IF-RL in Table~\ref{tab:results_after_ifrl}, using \ua{} to denote improvements and \da{} to mark degradations after Math RL training.
}
\label{tab:results_after_mathrl}
\begin{tabular}{lccc}
\toprule
\shortstack[l]{\textbf{Benchmark}\\\textbf{Metric: pass@1}}
 & \shortstack{\textbf{8B-Thinking}\\\textbf{Math RL}}
 & \shortstack{\textbf{8B (unified)}\\\textbf{Math RL}} 
  & \shortstack{\textbf{14B-Thinking}\\\textbf{Math RL}}
 \\
\midrule
\multicolumn{4}{l}{\textbf{Knowledge and Reasoning}} \\
MMLU~{\footnotesize\textcolor{gray}{(EM)}}                                                       &
83.8~\ua{0} & 83.4~\ua{0} & 84.8~\da{0.2}   \\
MMLU-Pro~{\footnotesize\textcolor{gray}{(EM)}}                                                             & 75.0~\ua{0.2} & 75.0~\ua{0.5} & 76.9~\ua{0.5}   \\
GPQA Diamond~{\footnotesize\textcolor{gray}{(avg@8)}} & 63.8~\da{1.4} & 65.7~\da{0.4} & 67.6~\da{2.6}  \\
\midrule
\multicolumn{4}{l}{\textbf{Alignment}} \\
ArenaHard (GPT4-turbo-2024-04-09)                 & 
84.0~\da{2.3} & 87.0~\da{1.0}  & 89.3 \da{0.9}   \\
IFEval (strict prompt)~{\footnotesize\textcolor{gray}{(avg@8)}}             &
84.7~\ua{1.3} & 92.1~\ua{1.7} & 84.3~\ua{3.0}   \\
IFBench~{\footnotesize\textcolor{gray}{(avg@8)}}                               &
39.8~\da{2.3} & 40.4~\da{0.1} & 41.0 \ua{0.6}  \\
\midrule
\multicolumn{4}{l}{\textbf{Math}} \\
AIME 2024~{\footnotesize\textcolor{gray}{(avg@64)}}   &
 90.2~\ua{4.6} & 90.2~\ua{4.0} & 90.4~\ua{1.3}   \\
AIME 2025~{\footnotesize\textcolor{gray}{(avg@64)}}   & 
80.2~\ua{7.9} & 81.9~\ua{6.7} & 83.3~\ua{1.0}   \\
\midrule
\multicolumn{4}{l}{\textbf{Code}} \\
LiveCodeBench v5~{\footnotesize\textcolor{gray}{(avg@8)}}    & 
 71.2~\ua{2.2} & 70.6~\ua{0.4} & 75.2~\da{0.1}  \\
LiveCodeBench v6~{\footnotesize\textcolor{gray}{(avg@8)}}    & 
 68.1~\ua{2.2} & 67.4~\ua{0.7} & 72.4~\da{0.3}   \\
SWE-bench Verified~{\footnotesize\textcolor{gray}{(avg@4)}}  & 32.5~\ua{0.1} & 30.6~\ua{2.3} & 39.7~\ua{1.3}   \\
% \midrule
% \multicolumn{4}{l}{\textbf{Tool calling}} \\
% BFCL V3~{\footnotesize\textcolor{gray}{(avg@4)}}     & -- & -- & -- & -- \\
\bottomrule
\end{tabular}
\end{table}

\subsubsection{Results after Math RL}
We monitor the training dynamics of Math RL for the 8B unified model by tracking its performance on AIME24 and AIME25, as shown in Figure~\ref{fig:mathrl_results}.
The results after Math RL are presented in Table~\ref{tab:results_after_mathrl}. We observe noticeable improvements on AIME 2024 and 2025.
Overall, Math RL has minimal effect on knowledge reasoning and alignment benchmarks. Most of the observed differences can be attributed to evaluation variance and checkpoint selection.
Math RL improves the coding benchmarks, including both LiveCodeBench and SWE. Although the gains are not as pronounced as those reported in AceReason-Nemotron~\citep{chen2025acereason}, this is largely because our starting models  already exhibit strong general reasoning capabilities prior to Math RL.
% In practice, we still find that applying Math RL before Code RL helps stabilize Code RL training.

\subsection{Code RL}
\label{subsec:code_rl}
% @Zhuolin
% \citep{chen2025acereason}
In this subsection, we describe the Code RL process, which focuses on improving the model’s competitive programming performance through reinforcement learning.
We apply Code RL to the model checkpoint obtained after Math RL.

\subsubsection{Data Curation}

We construct our Code RL training dataset based on the AceReason-Nemotron coding corpus~\citep{chen2025acereason}, which is primarily curated from open-source datasets containing unit tests, including TACO~\citep{li2023taco}, APPS~\citep{hendrycksapps2021}, DeepCoder~\citep{deepcoder2025}, and others.
These problems cover a wide range of algorithmic topics commonly found in modern competitive programming.
We apply strict filtering rules to exclude problems that are incompatible with standard output comparison (e.g., interactive formats or those requiring special judges), as well as problems with insufficient unit test coverage for edge and corner cases. This filtering process substantially reduces false-positive and false-negative reward signals during training, which are known to degrade Code RL performance~\citep{chen2025acereason}.

We further perform rigorous validation of the training set to remove duplicates and prevent benchmark contamination, using 9-gram filtering and raw problem URL matching. To calibrate problem difficulty, we employ AceReason-Nemotron-7B~\citep{chen2025acereason_7b} to exclude trivial problems (solved in all 8 out of 8 rollouts) and DeepSeek-R1-0528~\citep{deepseek_r1_0528} to filter out intractable or overly difficult ones (unsolved in all 8 rollouts), resulting in a final training set of 9.8K samples.

\subsubsection{Training Recipe}

We conduct Code RL after Math RL, as the Math RL stage serves as an effective warm-up that stabilizes future RL training and enhances the model’s general reasoning capabilities~\citep{chen2025acereason}.
Following the AceReason-Nemotron recipe, we perform single-stage, on-policy Code RL (without KL regularization, using token-level loss as in \S~\ref{subsec:rl_config}) initialized from the final Math-RL model checkpoint.
During training, the maximum response length is set in the range of 44K–48K, with no overlong filtering applied.

\paragraph{Reward function:} 
Code RL adopts a strict binary rule-based reward function, where a reward of 1 is assigned only when the generated code passes all test cases for the given problem; otherwise, a reward of 0 is assigned.
For efficient and robust evaluation, we employ the parallelized code verifier from the \href{https://huggingface.co/nvidia/AceReason-Nemotron-14B/blob/main/README_EVALUATION.md}{AceReason Evaluation Toolkit} to verify the correctness of the model's generated code.
Furthermore, we apply asynchronous reward computation in VeRL~\citep{sheng2024hybridflow}, as code verification incurs significant overhead. 
This asynchronous computation substantially reduces the averaged code verification time per batch. For example, when training Code RL on 8 DGX H100 nodes with a batch size of 128 and a rollout of 8, the verification time drops from 1172.4 seconds to 416.2 seconds. % using a thread pool of 64 workers per node.

Similar to Math RL, we also apply a code-switching penalty, assigning a reward of 0 whenever tokens from a language different from the original prompt’s language are detected in the reasoning trace.
Unlike in Math RL, we find that assigning a reward of $-1$ for code-switching negatively impacts coding performance. This is likely because the additional penalty encourages the model to produce incorrect answers without code-switching in GRPO training when all rollouts in the group are either incorrect or contain language mixing.

\paragraph{Hyperparameter:}
We set the batch size to 128, the learning rate to $4\times10^{-6}$ with the AdamW optimizer, and use 8 rollouts per training prompt.
We set sampling temperature as 1.0 and \texttt{top\_p} as 0.95 as we found that Code RL is sensitive to the temperature configuration.
The detailed hyperparameters can be found in Appendix~\ref{appendix:training_hyperparams}.

\subsubsection{Results after Code RL}
The results after Code RL are presented in Table~\ref{tab:results_after_coderl}. 
We observe strong gains on LiveCodeBench (LCB). For instances, our unified 8B achieves 75.3 on LCB v5 and 71.5 on LCB v6, matching the performance of DeepSeek-R1-0528 (671B), which achieves 74.8 and 73.3, respectively. 
Our 14B-Thinking achieves 78.0 on LCB v5 and 74.8 on LCB v6, outperforming DeepSeek-R1-0528 by a clear margin.
Since DeepSeek-R1-0528 (671B) is the teacher model used during SFT, these results highlight how remarkably effective Cascade RL is at strengthening code-reasoning ability—even for small 8B and 14B models.
Code RL also has minimal impact on benchmarks from other domains, aside from normal checkpoint and evaluation variance.

The superb coding capability of our Nemotron-Cascade models is further examined in Section \S\ref{sec:deepdive_on_code}.

\begin{table}[t]
\centering
\footnotesize
\renewcommand{\arraystretch}{1.15}
\caption{The evaluation results of 8B/14B-Thinking and the \emph{unified} 8B models \textbf{after Code RL} are presented below.  For unified model, we evaluate IFEval in the non-thinking mode and all other benchmarks in the thinking mode.
We compare the results against those obtained in the previous stage (Math RL) in Table~\ref{tab:results_after_mathrl}, using \ua{} to denote improvements and \da{} to indicate degradations after Code RL training.
}
\label{tab:results_after_coderl}
% \begin{adjustbox}{width={0.9\textwidth}}
\begin{tabular}{lccc}
\toprule
\shortstack[l]{\textbf{Benchmark}\\\textbf{Metric: pass@1}}
 & \shortstack{\textbf{8B-Thinking}\\\textbf{Code RL}}
 & \shortstack{\textbf{8B (unified)}\\\textbf{Code RL}} 
  & \shortstack{\textbf{14B-Thinking}\\\textbf{Code RL}}
 \\
\midrule
\multicolumn{4}{l}{\textbf{Knowledge and Reasoning}} \\
MMLU~{\footnotesize\textcolor{gray}{(EM)}}                                                      &
84.3~\ua{0.5}  & 83.7~\ua{0.3} & 85.1~\ua{0.3}  \\
MMLU-Pro~{\footnotesize\textcolor{gray}{(EM)}}                                                             & 75.4~\ua{0.4} & 75.3~\ua{0.3} & 77.6~\ua{0.7}  \\
GPQA Diamond~{\footnotesize\textcolor{gray}{(avg@8)}} & 67.7~\ua{3.9} & 67.4~\ua{1.7} & 70.3~\ua{2.7} \\
\midrule
\multicolumn{4}{l}{\textbf{Alignment}} \\
ArenaHard (GPT4-turbo-2024-04-09)                 & 
85.4~\ua{1.4} & 87.8~\ua{0.8} & 89.8 \ua{0.5}  \\
IFEval (strict prompt)~{\footnotesize\textcolor{gray}{(avg@8)}}             &
83.1~\da{1.6} & 90.7~\da{1.4} & 81.8 \da{2.5}   \\
IFBench~{\footnotesize\textcolor{gray}{(avg@8)}}                               &
41.8~\ua{2.0} & 38.1~\da{2.3} & 41.0 \ua{0.0} \\
\midrule
\multicolumn{4}{l}{\textbf{Math}} \\
AIME 2024~{\footnotesize\textcolor{gray}{(avg@64)}}   &
88.3~\da{1.9} & 89.1~\da{1.1} & 90.4~\ua{0.0}  \\
AIME 2025~{\footnotesize\textcolor{gray}{(avg@64)}}   & 
81.8~\ua{1.6} & 80.5~\da{1.4} & 83.5~\ua{0.2}  \\
\midrule
\multicolumn{4}{l}{\textbf{Code}} \\
LiveCodeBench v5~{\footnotesize\textcolor{gray}{(avg@8)}}    & 
74.3~\ua{3.1} & 75.3~\ua{4.7} & 78.0~\ua{2.8}   \\
LiveCodeBench v6~{\footnotesize\textcolor{gray}{(avg@8)}}    & 
71.0~\ua{2.9} & 71.5~\ua{4.1} & 74.8~\ua{2.4}  \\
SWE-bench Verified~{\footnotesize\textcolor{gray}{(avg@4)}}  & 33.3~\ua{0.8} & 31.6~\ua{1.0} & 39.6~\da{0.1}  \\
% \midrule
% \multicolumn{4}{l}{\textbf{Tool calling}} \\
% BFCL V3~{\footnotesize\textcolor{gray}{(avg@4)}}     & -- & -- & -- & -- \\
\bottomrule
\end{tabular}
% \end{adjustbox}
\end{table}

\subsection{SWE RL}
\label{subsec:swe_rl}
% @Yangyi 
In Section \S~\ref{subsec:swe_data_curation}, we employ the Agentless framework for SWE-bench~\citep{jimenez2023swe}, decomposing the SWE task into three sub-tasks: localization, repair, and patch validation. We construct the SFT data for each of these sub-tasks accordingly.
Among these sub-tasks, code repair is the most critical one, requiring the highest level of reasoning and model capability to generate revised code patches that fix bugs and address underlying issues.
As a result, our RL process for SWE is primarily designed to enhance code repair accuracy.

\subsubsection{Data Curation}
\label{subsubsec:swe_rl_data_curation}

As described in \S~\ref{subsec:swe_data_curation}, the RL dataset for code repair consists of more challenging instances than those used in the SFT stage. Specifically, we retain prompts for which fewer than four of the eight sampled responses exceed the 0.5 similarity threshold, while at least one of the eight responses from DeepSeek-R1-0528~\citep{deepseek_r1_0528} attains non-zero similarity (indicating the prompt is not too hard or unsolvable).

During the SFT stage, our models are fine-tuned with a maximum total sequence length of 32K. Accordingly, we construct prompts that contain only the ground-truth localization files—i.e., all files that include bugs or require modifications to resolve the issue—as references for code repair.
However, when evaluating model performance under the agentless framework, we provide the model with file contents retrieved from the localization stage as input for code repair. 
This setup introduces a discrepancy between the SFT training and the final evaluation. To ensure that the ground-truth localization files are included in the repair prompts, we incorporate the top-$k$ ($k \geq 4$) localized files and extend the maximum prompt length to 60K with YaRN scaling factor 3.

This design naturally introduces out-of-distribution contexts for the SFT model in two ways: \emph{i}) the total input length during code repair exceeds the maximum sequence length used in SFT; and \emph{ii}) the inclusion of top-$k$ localized files may bring in irrelevant files, making the code repair task more challenging than during SFT.

To address this issue, we construct and combine two subsets of long prompts (up to $l$ tokens) for RL training:
\begin{enumerate}
\item Ground-truth only: Similar to SFT, prompts are constructed using only the ground-truth localization files.
\item Mixed localization: Augmented prompts are built using both the files localized by DeepSeek-R1-0528 and the ground-truth localization files. 
We include up to five files in total and ensure that all ground-truth files are present. 
Specifically, the initial prompt contains only the ground-truth files. 
We then add noisy files one by one until the total prompt length would exceed $l$; if the limit is exceeded before adding any noisy files, we discard the instance.
To enhance robustness, we also randomize the order of files within each prompt.
\end{enumerate}
To further enhance training efficiency, for both subsets, we discard prompts whose total length is shorter than 8K tokens.
In \S~\ref{subsec:ablation_swe_rl_max_prompt_length}, we will ablate the effectiveness of RL training with various $l$.

% @jack

\subsubsection{Training Recipe}
\label{subsubsec:swe_rl_training_recipe}
We perform reinforcement learning for software engineering (SWE RL) as the final stage of Cascade RL, since it represents a more specialized task compared to the general domains.
Starting from the checkpoint obtained after code RL, we conduct on-policy RL using the GRPO algorithm with a token-level loss, while removing KL regularization (see detailed configurations in \S~\ref{subsec:rl_config}).

% We define the reward for each solution patch based on its similarity with the ground truth patch written by humans:
% \[
% R(s_p, g_p) = \begin{cases}
% \text{sim}(s_p, g_p), & \text{if } \text{sim}(s_p, g_p) > 0.5 \\
% 0, & \text{otherwise}
% \end{cases}
% \]

% where $s_p$ denotes the solution patch and $g_p$ denotes the ground truth patch.
% Empirically, adding this reward shaping prevents the model from continue focusing on trivial progress by matching the simple and superficial format used in the human-written patch. 
% % 
% We set batch size to 128 and learning rate to $4\times 10^6$ with AdamW.

% 
\paragraph{Reward function:} 
Previous studies~\citep{jain2025r2e, deepswe} perform RL by executing model-generated code patches within Docker environments to obtain rewards. 
However, running and managing numerous Docker instances significantly limits scalability, constraining prior work to training datasets of around 10k unique instances. 
To overcome this limitation, we design a execution-free verifier as the reward model, enabling scalable RL training for code repair generation. 
That is, we define the reward \( r \) as the similarity between the generated patch \( \hat{p} \) and the human-annotated ground-truth patch \( p^* \):
\begin{align}
r(\hat{p}, p^*) =
\begin{cases}
1, & \text{if } s_{\text{lex}}(\hat{p}, p^*) = 1, \\
0, &  \hat{p} \text{ is identical to the original code snippet}  \\
-1, & \text{if } \hat{p} \text{ cannot be parsed}, \\ %[6pt]
s_{\text{sem}}(\hat{p}, p^*), & \text{otherwise},
\end{cases}
\label{equation:swe_rl_reward}
\end{align}
where \( s_{\text{lex}}(\hat{p}, p^*) \) denotes the lexical similarity computed with \emph{Unidiff} library following \citet{wei2025swe}, and \( s_{\text{sem}}(\hat{p}, p^*) \) represents the semantic similarity score produced by a LLM. 
Specifically, we prompt the Kimi-Dev-72B model\footnote{\url{https://huggingface.co/moonshotai/Kimi-Dev-72B}}~\citep{team2025kimi_k2} with a yes/no question to assess the semantic similarity between the generated and golden patches (see the reward modeling prompt in Appendix~\ref{appendix:swe_template}).
The probability assigned to the “YES” token is directly used as the reward score. 
Note that we assign a reward of $-1$ when the model’s generated patch fails to parse, and a reward of 0 when the generated patch is identical to the original code snippet. 
We refer readers to \S~\ref{subsec:ablation_swe_rl_reward_model} for ablation studies on reward functions.

\paragraph{Multi-stage RL training for input context extension:}

Our preliminary experimental investigations reveal a strong positive correlation between input context length and SWE task performance, specifically demonstrating that the inclusion of additional retrieved files for analysis yields substantial performance improvements. 
This finding motivates the design of our training strategy, which exploits this relationship through controlled context expansion.
To optimize the utilization of extended context while maintaining training stability, we implement a carefully designed two-stage curriculum that progressively expands the input context length from 16K to 24K tokens while maintaining a constant output length of 16K tokens. This approach ensures robust learning and avoids the degradation effects observed with immediate long-context training, which is particularly effective for 8B models, given that smaller models have limited long-context capability.

\begin{itemize}
    \item 
    \textbf{16K Context Initialization (Warmup Stage).} 
    The training process begins with a conservative 16K input token budget, which serves as an essential warmup stage. Our empirical analysis shows that directly initializing training with a 24K context length leads to suboptimal convergence and degraded final performance—a phenomenon we attribute to the model’s initial difficulty in attending to and synthesizing information across extended sequences. During this stage, the model learns fundamental long-context utilization skills and develops stable attention mechanisms for multi-file analysis within a manageable context window.
\vspace{0.05cm}
    \item 
    \textbf{24K Context Extension.} 
    Once the 16K setup reaches a reward plateau, with little improvement over successive iterations, we extend the context to 24K tokens.
    The timing of this transition is important: the model has already built strong multi-file analysis skills at 16K, forming a solid basis for scaling to longer context. During the extended phase, we observe steady gains in long-context understanding, including more advanced cross-file reasoning and improved synthesis of information across retrieved files.
    The model demonstrates increasing proficiency in leveraging the expanded context window, effectively using the additional retrieved files to produce more accurate solutions.
\end{itemize}

\begin{table}[t!]
\centering
\footnotesize
\renewcommand{\arraystretch}{1.15}
\caption{The evaluation results of our final 8B/14B-Thinking and the \emph{unified} 8B models \textbf{after SWE RL} are presented below (Note that they are the final models).  For unified model, we evaluate IFEval in the non-thinking mode and all other benchmarks in the thinking mode.
We compare the results against those obtained in the previous stage (Code RL) in Table~\ref{tab:results_after_coderl}, using \ua{} to denote improvements and \da{} to indicate degradations after SWE RL training.
}
\label{tab:results_after_swerl}
\begin{tabular}{lccc}
\toprule
\shortstack[l]{\textbf{Benchmark}\\\textbf{Metric: pass@1}}
 & \shortstack{\textbf{8B-Thinking}\\\textbf{SWE RL}}
 & \shortstack{\textbf{8B (unified)}\\\textbf{SWE RL}} 
  & \shortstack{\textbf{14B-Thinking}\\\textbf{SWE RL}}
 \\
\midrule
\multicolumn{4}{l}{\textbf{Knowledge and Reasoning}} \\
MMLU~{\footnotesize\textcolor{gray}{(EM)}}                                                       &
84.0~\da{0.3} & 83.7 \ua{0.0} & 85.1 \ua{0.0}  \\
MMLU-Pro~{\footnotesize\textcolor{gray}{(EM)}}                                                             & 75.5 \ua{0.1} & 75.7 \ua{0.4} & 77.0 \da{0.6}  \\
GPQA Diamond~{\footnotesize\textcolor{gray}{(avg@8)}} & 66.7 \da{1.0} & 66.5 \da{0.9} & 69.6 \da{0.7} \\
\midrule
\multicolumn{4}{l}{\textbf{Alignment}} \\
ArenaHard (GPT4-turbo-2024-04-09)                 & 
85.8~\ua{0.4} & 87.9 \ua{0.1} & 89.5 \da{0.3} \\
IFEval (strict prompt)~{\footnotesize\textcolor{gray}{(avg@8)}}             &
83.7~\ua{0.6} & 90.2 \da{0.5} & 81.9 \ua{0.1}  \\
IFBench~{\footnotesize\textcolor{gray}{(avg@8)}}                               &
41.4~\da{0.4} & 40.8 \ua{2.7} & 41.7  \ua{0.7} \\
\midrule
\multicolumn{4}{l}{\textbf{Math}} \\
AIME 2024~{\footnotesize\textcolor{gray}{(avg@64)}}   &
88.8~\ua{0.5} & 89.5 \ua{0.4} & 89.7 \da{0.7} \\
AIME 2025~{\footnotesize\textcolor{gray}{(avg@64)}}   & 
81.4~\da{0.4} & 80.1 \da{0.4} & 83.3 \da{0.2} \\
\midrule
\multicolumn{4}{l}{\textbf{Code}} \\
LiveCodeBench v5~{\footnotesize\textcolor{gray}{(avg@8)}}    & 
74.5~\ua{0.2} & 74.3 \da{1.0} & 77.5 \da{0.5} \\
LiveCodeBench v6~{\footnotesize\textcolor{gray}{(avg@8)}}    & 
71.4~\ua{0.4} & 71.1 \da{0.4} & 74.6 \da{0.2} \\
SWE-bench Verified~{\footnotesize\textcolor{gray}{(avg@4)}}  & 38.5 \ua{5.2} & 37.2 \ua{5.6} & 43.1 \ua{3.5}  \\
% \midrule
% \multicolumn{4}{l}{\textbf{Tool calling}} \\
% BFCL V3~{\footnotesize\textcolor{gray}{(avg@4)}}     & -- & -- & -- & -- \\
\bottomrule
\end{tabular}
\end{table}

\paragraph{Hyperparameters:} 
We set the batch size to 128 and the learning rate to $2.5\times10^{-6}$ using the AdamW optimizer. For each prompt, we generate 16 rollouts with a sampling temperature of 1 and set maximum response length to 16K. 
We apply the overlong filtering for the trajectories that reach maximum response length.
The detailed hyperparamters can be found in Appendix~\ref{appendix:training_hyperparams}.

\subsubsection{Results after SWE RL}
The results after applying SWE RL are shown in Table~\ref{tab:results_after_swerl}. SWE RL yields substantial gains on SWE-bench Verified, while its positive or negative impact on benchmarks from other domains remains minimal and is largely attributable to checkpoint and evaluation variance.
Our 14B-Thinking model achieves a pass@1 resolve rate of 43.1, already outperforming the recent open 32B specialized models, DeepSWE-32B (42.2)~\citep{deepswe} and SWE-agent-LM-32B (40.2)~\citep{yang2025swe}. 
It also performs significantly better than other 14B general-purpose open LLMs, such as Qwen3-14B~(27.4) and Ministral-3-14B-Reasoning-2512~\citep{mistralai2025Ministral3_14B}~(25.5).
Interestingly, we find that the performance gap on SWE-bench Verified between the dedicated 8B \emph{thinking} SFT model and the 8B \emph{unified} SFT model (30.2 vs.\ 26.1 in Table~\ref{tab:results_after_sft}) is largely mitigated after the full Cascade RL process (38.5 vs.\ 37.2).
In conclusion, the unified Nemotron-Cascade-8B performs comparably to Nemotron-Cascade-8B-Thinking on all reasoning-related tasks, while performing substantially better on instruction-following tasks.

\section{Deep Dive on Competitive Coding}
\label{sec:deepdive_on_code}

We evaluate the performance of our Nemotron-Cascade models on challenging competitive programming benchmarks, including LiveCodeBench~\citep{jain2024livecodebench}, which contains recently released AtCoder and LeetCode problems, and LiveCodeBench Pro~\citep{zheng2025livecodebench}, which includes newly released Codeforces problems.
To avoid benchmark contamination, we report accuracy only on problems released after our training data cutoff of 08/2024.
For LiveCodeBench, we evaluate on subsets v5 (08/2024–02/2025, \textbf{279} problems) and v6 (08/2024–05/2025, \textbf{454} problems).
For LiveCodeBench Pro, we use the two most recent subsets: 2025Q1 (01/2025–04/2025, \textbf{166} problems) and 2025Q2 (04/2025–07/2025, \textbf{167} problems).
%, and \textbf{2025Q3}~(2025/07-2025/09, \textbf{144} problems in total)
We perform our evaluation under avg@8 setting with thinking budgets as 64K tokens. We also evaluate model ELO score based on \textbf{51} Codeforces Rounds from LiveCodeBenchPro \{2025Q1, 2025Q2\} split. We also leave more details and analysis related to Elo Rating calculation in Appendix \ref{appendix:elo_rating_on_codeforces}.

% \usepackage{multirow}
\begin{table}[t!]
\centering
\caption{Competitive programming results on comprehensive benchmarks, evaluated against a significantly expanded set of proprietary and open-source baseline models.}
\begin{adjustbox}{width={1.0\textwidth}}
\setlength{\tabcolsep}{10pt}
\renewcommand{\arraystretch}{1.2}
\begin{tabular}{l|cc|cccc|cc}

\arrayrulecolor{black}\toprule
\multirow{3}{*}{\bf Models} & \multicolumn{2}{c|}{\bf LiveCodeBench} & \multicolumn{4}{c|}{\bf LiveCodeBench Pro}               & \multicolumn{2}{c}{\bf Codeforces}  \\
                        & \bf v5               & \bf  v6              & \multicolumn{2}{c}{\bf 25Q1} & \multicolumn{2}{c|}{\bf 25Q2} & \multicolumn{2}{c}{2501~-~2507 } \\
                        & 2408~-~2502      & 2408~-~2505     & {\bf \color[HTML]{808080} Easy}        & {\bf \color[HTML]{808080} Med}       & {\bf \color[HTML]{808080} Easy}         & {\bf \color[HTML]{808080} Med}        & \bf ELO         & \bf  Percentile        \\ \arrayrulecolor{black} \midrule
o4-mini (high)                    &    \bf  82.8                        & \textbf{80.2}                &    \bf      85.4                  &           \bf   51.7            &        \bf     84.5              &       \bf           29.8       
%&    80.0                 &           32.7     
&       \bf   2169                     &   \bf  99.1                           \\
o3 (high)                  &       78.1                       & 75.8                         &        79.8                  &         35.0                 &           82.5               &               26.3       
%&         73.8                 &            24.5     
&              2094                &   98.4                           \\
o4-mini (medium)                  &       76.3                       & 74.2                         &           -               &     -                     &            -              &       -            %       &                          &           
&               -               &               -               \\

%gpt-oss-120B-high (no tool)       &        85.7                      &            84.5                  &                          &                          &                          &                          &                          &                          &                              &                              \\
Gemini-2.5-Pro-06-05              &      73.5                        & 73.6                         & 76.4                     & 26.7                     & 77.3                     & 21.1              %       &            71.3              &    18.4        
&            2034                  &       98.1                       \\
%GLM-4.5                           &                              & ?                            &                          &                          &                          &                          &                          &                          &                              &                              \\
DeepSeek-R1-0528                  &   74.8                           & 73.3                         &         57.3                 &          6.7               &                63.9          &                  7.0       
%&          55.0               &               8.2 
&                 -            &             -                 \\
%DeepSeek-V3.2-Exp                 &                              & 74.1                &                          &                          &                          &                          &                          &                          &                              &                              \\
Qwen3-235B-A22B-Thinking-2507          &     81.6                         &       78.7             &      75.8                    &        18.8                  &       77.6                   &           17.5         
%&        73.8                  &          14.3    
&              1979                &        97.7                      \\
   Qwen3-235B-A22B (thinking mode)       &     70.7                        &       67.3             &     55.6                   &        9.2                &       52.3                 &           3.1         
&            1516              &    86.2                   \\

\arrayrulecolor{gray} \midrule
%Ring-1T-Preview                   &                              & 78.3                         &                          &                          &                          &                          &                          &                          &                              &                              \\ 
Qwen3-Next-80B-A3B-Thinking       &    76.0                          &       73.2           &      68.5                    &      \bf     16.3                &       \bf  69.1                  &           7.5         %      &             78.8             &      14.3
&                 1800             &           95.8                   \\
Magistral-Medium-1.2-2509	                      &        75.0                      & -                &     -                     &           -               &         -                 &             -             
%&     -                     &           %      -  
&           -                   &               -               \\
OpenReasoning-Nemotron-32B        &    71.8                          & 70.2                         &               66.9              &    11.7                      &            66.2              &     8.3               
%&                          &          
&       1766                       &                     95.3         \\
Llama-3.3-Nemotron-Super-49B-v1.5 &            70.3                  & 68.1                            &                63.5        &       11.3                   &            61.9              &        6.1           
%&                          &         
&      1716                        &         94.1                     \\
NVIDIA-Nemotron-Nano-9B-v2        & 68.2              &            65.3                  &   61.0                     &         7.9                &       59.3                   &            4.8       
% &                          &         
&              1642                &    91.2                          \\
Meta-CWM-32B                      &        -                      & 63.5                &     -                     &           -               &         -                 &             -             
%&     -                     &                 -  
&           -                   &               -               \\

%gpt-oss-20B-high (no tool)        &     72.5                         & 70.8                    &                          &                          &                          &                          &                          &                          &                              &                              \\
Klear-Reasoner-8B                 & 66.0                         & 63.0                         &           50.3               &       8.8                   &        49.7                  &       2.2                
%&                          &                  
&            1517                  &           86.2                   \\
AReaL-Boba-2-14B                  & 70.3                         &         67.4                     &                  56.2        &         4.2                 &           48.7               &      1.8               
%&                        &           
&                   1446               &             82.2                     \\
AceReason-Nemotron-1.0-14B        & 61.1                         & 58.7                         &    50.0                      &         5.0                 &     47.9                     &        2.2      
%&                          &             
&                  1437            &         81.3                     \\
AceReason-Nemotron-1.1-7B         & 57.2                         & 55.2                         &    40.2                      &    2.5                      &     42.3                     &           1.8          
%&                          &       
&      1340                        &               70.2              \\
 \midrule
\rowcolor{gray!20} Nemotron-Cascade-8B                     & 74.3                         & 71.1                         &               65.0           &    12.1                      &                 65.7            &        6.4       
%&                          &       
&       1789                       &          95.7                   \\
\rowcolor{gray!20} Nemotron-Cascade-8B-Thinking                     & 74.5                         & 71.4                         &               64.2           &    12.5                      &                 64.8            &        6.1       
%&                          &       
&       1740                       &          94.7                    \\
%\rowcolor{gray!20} Nemotron-Cascade-14B-Thinking-CodeRL                    & \bf 78.0                         & \bf 74.8                         &                 72.3         &    14.4                      &    70.0                      &         10.5
%&                          &      
%&             1914                 &                  97.0            \\
%\rowcolor{gray!20} Nemotron-Cascade-14B-Thinking-SWE-122                    & \bf 77.0                         & \bf 74.7                         &                 75.3         &    19.2                      &    67.9                      &         9.2
%&                          &      
%&             1890                 &                  96.7            \\
\rowcolor{gray!20} Nemotron-Cascade-14B-Thinking                    & \bf 77.5                         & \bf 74.6                         &                 \bf 71.6         &  \bf   16.3                      &    68.9                      &    \bf    10.5
%&                          &      
&        \bf     1932                 &  \bf                97.2            \\ \arrayrulecolor{black}\bottomrule
\end{tabular}
\label{tab:competitive_programming_results}
\end{adjustbox}
\end{table}

As shown in Table~\ref{tab:competitive_programming_results}, the Nemotron-Cascade models demonstrate strong performance across multiple competitive coding benchmarks, including the latest splits of LiveCodeBench and LiveCodeBench-Pro. Nemotron-Cascade-8B significantly outperforms nearly all recently released reasoning LLMs of comparable size and achieves comparable performance to the previous state-of-the-art distilled model, OpenReasoning-Nemotron-32B~\citep{ahmad2025opencodereasoning}, despite using far fewer parameters. 
Notably, the Nemotron-Cascade-14B-Thinking model even outperforms DeepSeek-R1-0528—its SFT teacher, Qwen3-235B-A22B, and Qwen3-Next-80B-A3B-Thinking across all competitive coding benchmarks, demonstrating the exceptional effectiveness of Cascade RL.

%Nemotron-Cascade-14B outperforms all open-source LLMs with fewer than 100B parameters and achieves performance comparable to much larger proprietary models such as Gemini-2.5-Pro and o3 (high). Notably, Nemotron-Cascade-14B even surpasses its SFT teacher, DeepSeek-R1-0528, demonstrating the potential of reinforcement learning to exceed the performance ceiling of SFT teachers while using substantially fewer parameters.

\subsection{Test-Time Scaling in Practice: IOI 2025}

Beyond standard competitive coding platform benchmarks, we further conduct evaluations on one of the most challenging competitive programming contest: the International Olympiad in Informatics (IOI) 2025. IOI imposes a strict limit of at most 50 submissions per problem, each with official judgment feedback, but does not explicitly constrain the number of model generations to construct those submissions. To fully exploit the reasoning capabilities of our strongest Nemotron-Cascade model, we deploy our Nemotron-Cascade-14B-Thinking with the total 128K token thinking budget and propose a feedback-driven, test-time scaling pipeline as follows.

The whole pipeline can be viewed as a multi-round \emph{generate–select–submit} process, with up to 50 rounds per problem (one for each submission). In each round, for every subtask of a problem, the model generates 20 candidate responses with different random seeds. We then filter out \textbf{(\emph{i})} incomplete responses that do not contain code, and \textbf{(\emph{ii})} generated code that fails to pass the provided sample test cases (if any). Among the remaining candidate generations for each subtask, we apply the Tail-10 selection heuristic from \citet{fu2025deepconfidence} to obtain the final high-quality response, and submit this response to the official judge to obtain verdicts and (for partial-score tasks) scores.

After each round, we update each subtask's generation prompt to incorporate the new feedback from official judge, so the later generations are conditioned on the history of failed submissions. Specifically, for each unsolved subtask in \emph{classic} problem, we append to the next-round prompt with up to 5 most recent submission codes to this subtask, and their corresponding official verdicts. We intentionally cap this history cache size at 5 to avoid overfitting to earlier failed attempts while still encouraging the model to analyze and improve upon past incorrect attempts. For \emph{partial-score} problems, we instead append up to 3 highest-scoring prior submissions and encourage the model to keep improving the scores.

Beyond submission history, we also bring cross-subtask insights: once a subtask is solved, its correct solution code is appended as \emph{insight} when prompting the model for other unsolved subtasks with different constraints of the same problem. This encourages the model to reason about relationships between constraints and to transfer good insights across subtasks. The complete prompt template is provided in Appendix \ref{appendix:cp_template}.

\begin{figure}[t]
\centering
\includegraphics[width=0.98\columnwidth]{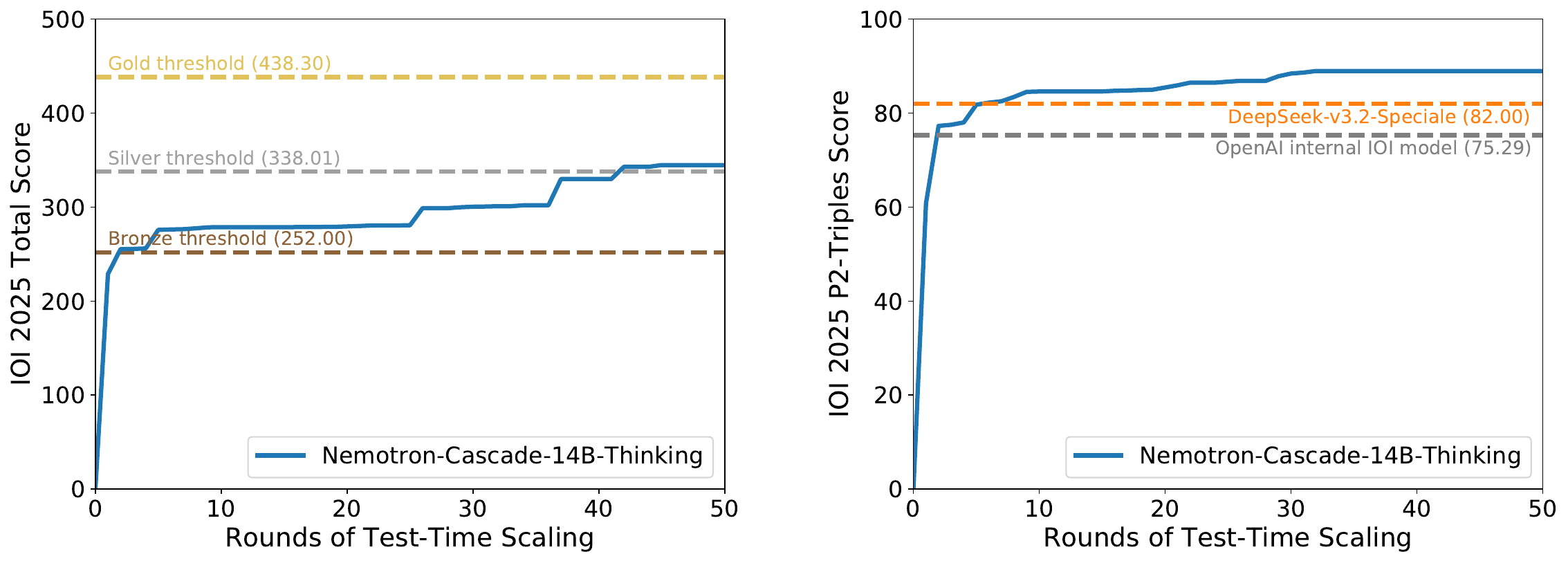}
\vspace{-.1cm}
\caption{IOI 2025: Nemotron-Cascade-14B-Thinking's score on \textbf{(Left)} Full problem set; \textbf{(Right)} Problem 2 \emph{Triples} after rounds of our proposed test-time scaling pipeline.}
\label{fig:ioi2025_tts}
\end{figure}

With this effective and explicitly self-improving test-time scaling strategy, our 14B-Thinking model achieves the overall score of \textbf{343.37} on IOI 2025, corresponding to a solid silver medal performance with at most 1000 generations (20 generations $\times$ 50 rounds), and no more than 50 official submissions to each problem. Notably, on IOI2025 Problem 2 \textit{Triples}, which contains a constructive subtask that requires proposing and iteratively refining a construction algorithm, our pipeline achieves \textbf{90.37} points, outperforming both OpenAI's internal IOI-gold model (75.29 points) and DeepSeek-V3.2-Speciale (82 points)~\citep{liu2025deepseek}. This proves the effectiveness of our feedback-driven, self-evolving test-time scaling approach, on real, high-stakes competitive programming problems. We also show our rounds of progress in Figure~\ref{fig:ioi2025_tts}.

We further analyze Code-RL training through the following ablations:

\begin{figure}[t]
\centering
\begin{subfigure}{0.48\columnwidth}
\includegraphics[width=\columnwidth]{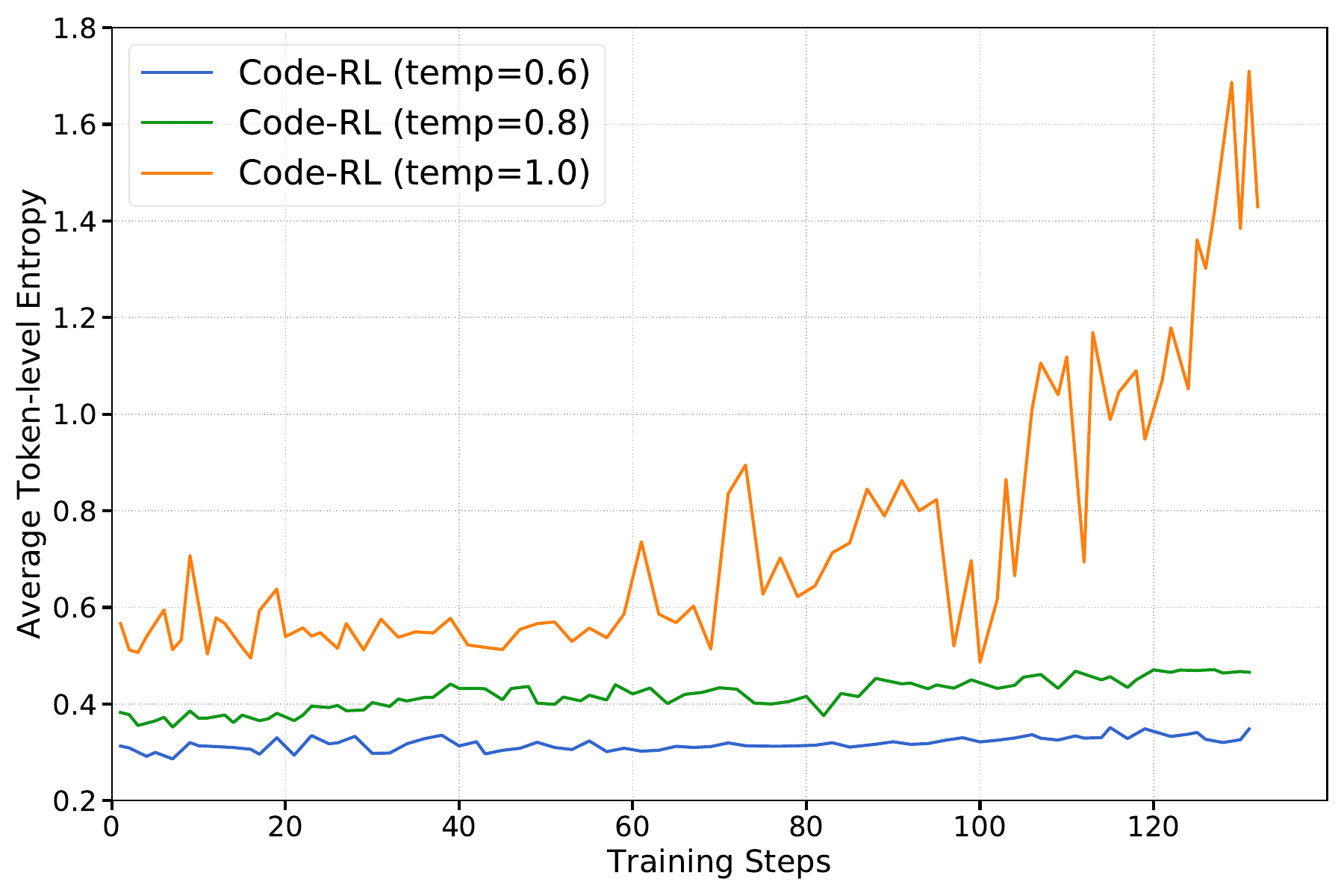}
\end{subfigure}
\hspace{0.02\columnwidth}
\begin{subfigure}{0.48\columnwidth}
\includegraphics[width=\columnwidth]{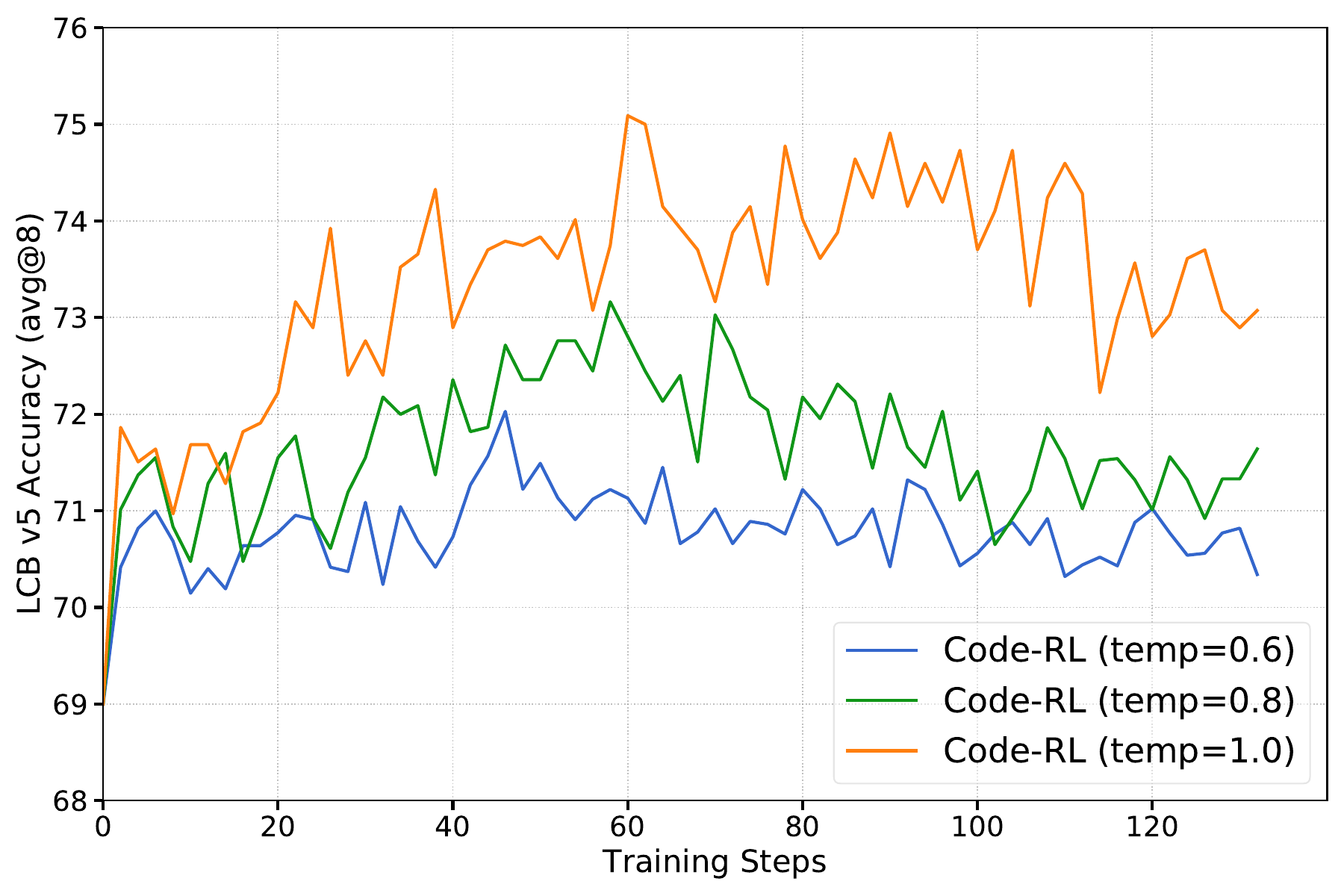}
\end{subfigure}
\vspace{-0.2cm}
\caption{\textbf{(Left)} Average Token Entropy \textbf{(Right)} Model accuracy curves of our 8B \emph{unified} model during Code RL training under different temperatures \{0.6, 0.8, 1.0\}. Training with high temperature yields better model accuracy but may suffer from training instability.}
\label{fig:ablation_coderl_temp}
% \vspace{-0.5cm}
\end{figure}

\subsection{The Role of Training Temperature in Code RL} 
To identify the most suitable temperature for Code RL training, we conduct ablation experiments using temperatures of 0.6, 0.8, and 1.0 on 8B unified model (RL curves shown in Figure~\ref{fig:ablation_coderl_temp}).
While lower temperatures produce more stable entropy curves, they lead to degraded code-reasoning performance compared with higher-temperature settings. This pattern suggests that in large, noisy sampling spaces such as code generation, higher temperatures encourage exploration and improve sample efficiency under limited rollout budgets. However, high temperatures may also induce training instability, leading to entropy explosion. Designing training frameworks that retain the benefits of high-temperature sampling while ensuring entropy stability is a promising direction for future work.

%\textbf{The Impact of Training Data Blend on Code-RL.} Figure XXX presents Code-RL accuracy curves on the LiveCodeBench v5 set across difficulty splits. As training progresses, accuracy on easy and medium splits rapidly saturates at over $90\%$, while the hard split plateaus near $52\%$, showing a large room for improvement. Notably, our initial training blend filtered out overly difficult problems based on the SFT teacher model (DeepSeek-R1-0528), potentially imposing a performance ceiling from the teacher's reasoning limitations. To address this, we re-filtered the ``overly hard'' problems using our latest Code-RL checkpoints with $8$ rollouts per problem. Surprisingly, approximately $23\%$ of problems unsolvable by DeepSeek-R1-0528 are successfully solved by our current models. Incorporating these previously excluded problems into the training blend and resuming Code-RL training yields clear performance gains on the hard split, as demonstrated in FigureXXX.

\subsection{How Cascade RL Improves Code Reasoning}
\begin{figure}[t]
\centering
\includegraphics[width=1.\columnwidth]{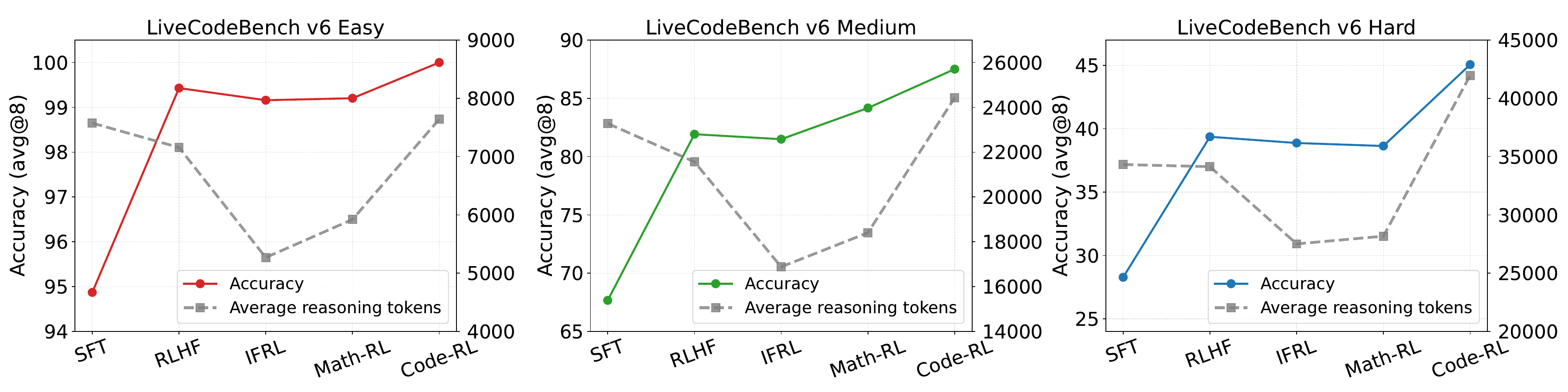}
\caption{Nemotron-Cascade-8B accuracy (avg@8) and average reasoning token counts after cascade RL stages on each difficulty splits of LiveCodeBench v6 (2408-2505). Curves indicate the improvements of reasoning capability on solving algorithmic coding problems after each stage.}
\label{fig:ablation_coderl_cascade}
% \vspace{-0.5cm}
\end{figure}

To assess the progressive effectiveness of our Cascade RL pipeline, we analyze the average reasoning-token usage and model accuracy across each difficulty split of LiveCodeBench v6 for our unified 8B model after successive cascaded RL stages—SFT, RLHF, IF-RL, Math RL, and Code RL (Figure~\ref{fig:ablation_coderl_cascade}). As shown in the figure, the initial RLHF stage provides a strong foundation: it substantially improves reasoning-token efficiency and mitigates the verbosity of the SFT model, evidenced by sharply reduced reasoning tokens alongside significant accuracy gains across all difficulty splits. The subsequent IF-RL stage further encourages conciseness, yielding an additional ~20\% reduction in token usage with only a negligible accuracy drop (0.5\%).

We further observe that performance on easy problems saturates (>99\%) after the initial stages, shifting the room for improvement to the medium and hard splits. Math RL enhances reasoning by increasing token usage, improving accuracy on medium problems, while Code-RL provides the final performance lift on both medium and hard problems through substantially expanded reasoning traces.

\begin{wrapfigure}{r}{0.41\textwidth} % r = right, width = 0.5 of text width
    \centering
    \vspace{-1.4em}
    \includegraphics[width=\linewidth]{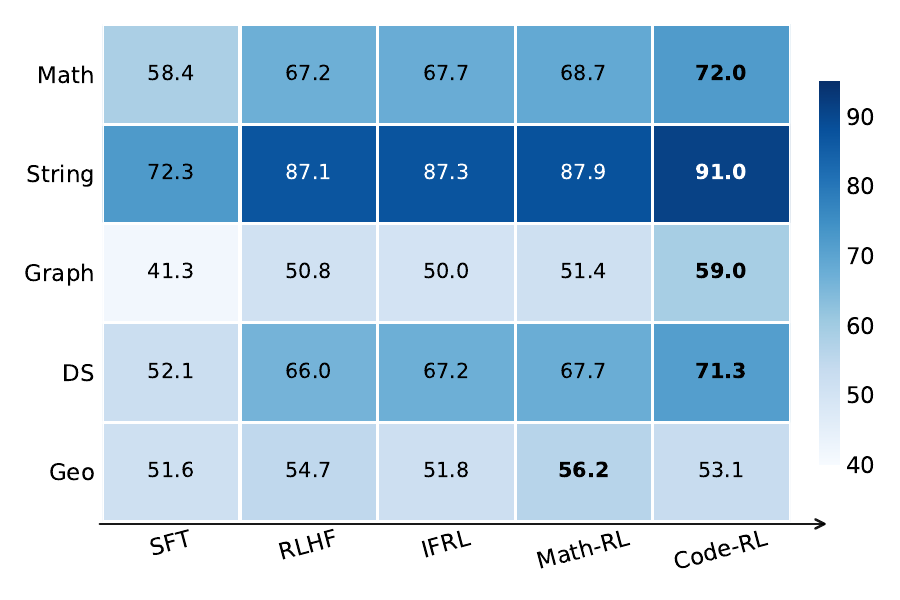}
    \vspace{-2.0em}
    \caption{\small Topic-wise accuracy of unified Nemotron-Cascade-8B after Cascade RL stages on LCB v6 set. DS refers to Data Structure and Geo refers to Geometry.}
    \vspace{-2em}
    \label{fig:ablation_coderl_topic}
\end{wrapfigure}

Additionally, we conduct ablation experiments to analyze how Cascade RL improves coding capability at the topic level. We annotate the LiveCodeBench v6 problems with five sub-categories—{Math, String, Graph, Data Structure, Geometry}—and report the topic-wise accuracy of our unified 8B model after each Cascade RL stage in Figure~\ref{fig:ablation_coderl_topic}. While RLHF provides strong initial gains across all sub-categories, Math RL primarily benefits math-related topics (Math, Graph, Geometry) and yields limited improvements on more computer-science-oriented ones (String, Data Structure). Code RL delivers the largest accuracy boost, improving performance across nearly all topics.

%present topic-wise accuracy on LiveCodeBench for each cascaded RL checkpoint to identify the sources of improvement: YYY.

\section{Deep Dive on RLHF}
\label{sec:deepdive_on_rlfh}

In this section, we present our findings on selecting effective reward models and designing a robust RLHF recipe. 
We show that RLHF trained with the largest reward model yields the strongest performance on the ArenaHard benchmark, particularly under style control \citep{chiang2024chatbot}, which helps disentangle substance from stylistic preferences in LLM responses.
We also find that smaller reward models tend to generate noisier reward signals, necessitating additional techniques such as reward shaping and KL regularization to preserve training stability.
For larger reward models, these techniques are unnecessary: their reward signals are sufficiently accurate and consistent on their own, enabling stable RLHF training and better performance on other tasks.

\subsection{RLHF Training Strategies for Unified Models}
\label{abl:rlhf_unified}

\begin{figure}[t]
\centering
\includegraphics[width=0.94\columnwidth]{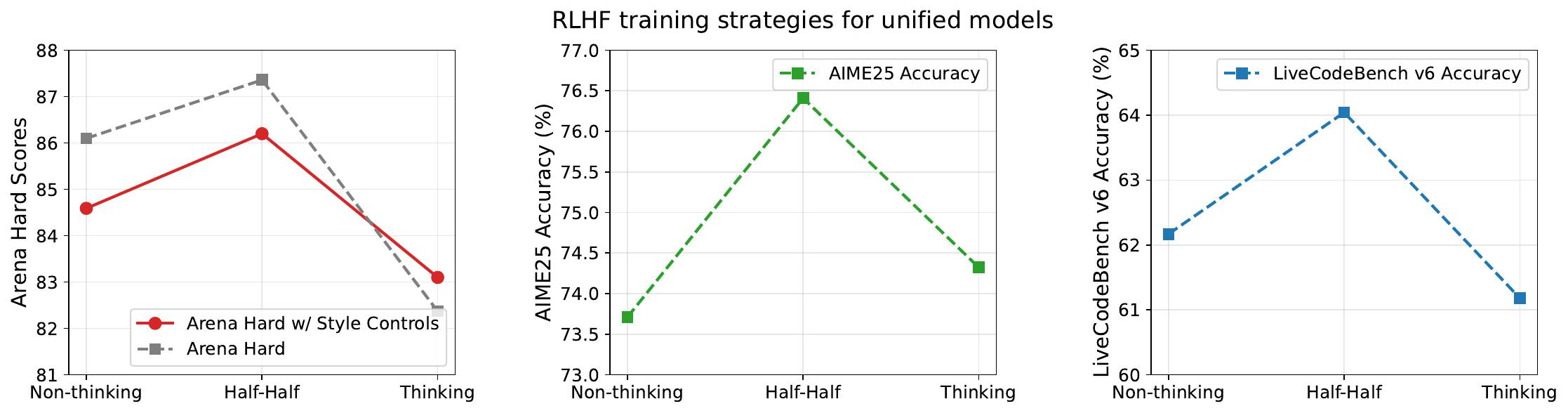}
\caption{
RLHF training results for unified 8B SFT model using the 72B reward model with different training strategies.  Training RLHF in both \textit{non-thinking} and \textit{thinking} modes, with an equal split of prompts allocated to each mode within every batch (``Half-Half''), is significantly better than training the unified model in the \textit{non-thinking} mode only (``Non-thinking'') and the \textit{thinking} mode only (``Thinking''), although the evaluation of ArenaHard, AIME, and LiveCodeBench is in the \textit{thinking} mode only. 
}
\label{fig:ablation_rlhf_unified}
% \vspace{-0.5cm}
\end{figure}

Since our unified model can respond in both \textit{thinking} and \textit{non-thinking} modes, a natural research question arises: which mode should we use for RLHF training, especially given that many benchmarks favor the \textit{thinking} mode? 
To investigate this, we apply RLHF to our 8B unified SFT model~(performance reported in Table~\ref{tab:results_after_sft}) using the same training recipe described in Section~\ref{subsec:rlhf_recipe}, but vary the training mode. Specifically, 
the ``Non-thinking'' setting uses only the \textit{non-thinking} mode during RLHF; 
the ``Thinking'' setting uses only the \textit{thinking} mode; 
and the ``Half-Half'' setting splits each batch evenly between the two modes.
The results, shown in Figure~\ref{fig:ablation_rlhf_unified}, reveal a clear trend: although ArenaHard, AIME, and LiveCodeBench are all evaluated in the \textit{thinking} mode, training in the ``Half-Half'' setting provides the strongest overall performance, yielding the highest ArenaHard scores as well as improved performance on math and code benchmarks. 
This indicates that including samples in the \textit{non-thinking} mode during RLHF can improve cross-mode transfer and alignment, leading to stronger general capabilities across both reasoning and non-reasoning settings.

% think or no think rlhf?

\subsection{Impact of Reward Model Size on RLHF Performance}
\label{abl:rmsize}

\begin{figure}[t]
\centering
\includegraphics[width=0.94\columnwidth]{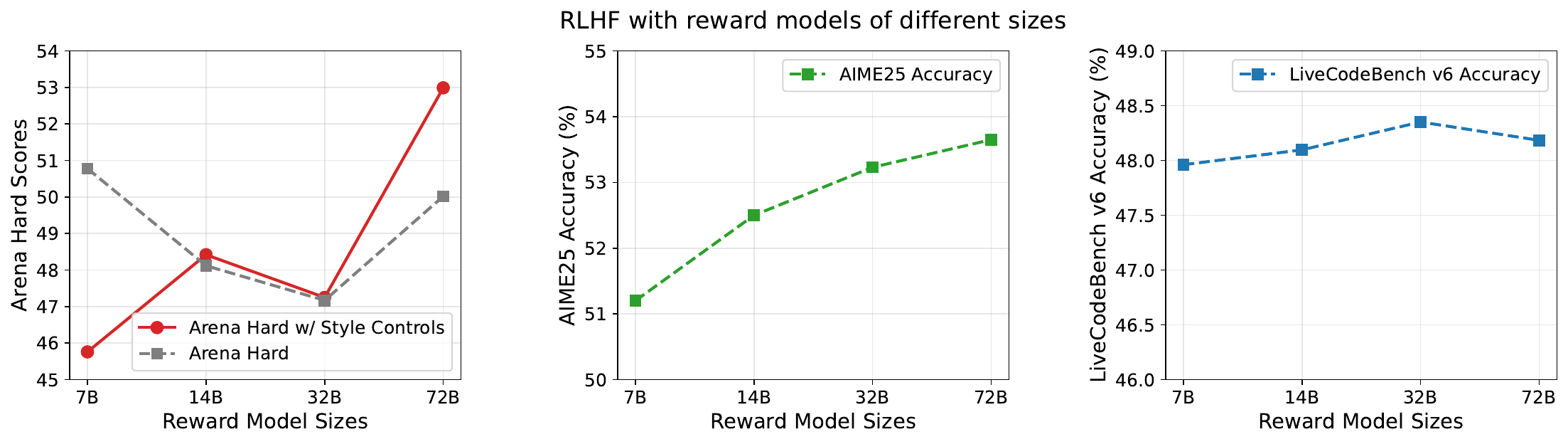}
\caption{
RLHF training results for AceReason-Nemotron-1.0-7B~\citep{chen2025acereason} using reward models ranging from 7B to 72B. Corresponding RewardBench scores for the reward models are provided in Table~\ref{table:rlhf_rm_ablation}. Using the largest reward model yields the best ArenaHard performance under style control, while smaller reward models lead to noticeable degradation in ArenaHard scores as well as math and code capabilities. 
}
\label{fig:ablation_rlhf_rm}
% \vspace{-0.5cm}
\end{figure}

A key research question in RLHF is how to select the most effective reward model. To study this systematically, we train a series of reward models ranging from 7B to 72B, and apply the same RLHF recipe described in \S~\ref{subsec:rlhf_recipe} to the AceReason-Nemotron-1.0-7B policy model~\footnote{We run early RLHF experiments with AceReason-Nemotron-1.0-7B as the policy model and perform ablations on it.}~\citep{chen2025acereason}. We report ArenaHard scores as well as performance on math and code benchmarks in Figure~\ref{fig:ablation_rlhf_rm}. Our key findings are summarized below.

\begin{enumerate}
    \item Larger reward models yield stronger ArenaHard performance. RLHF trained with the largest reward model achieves the highest ArenaHard scores, under style control \citep{chiang2024chatbot} that disentangles substance and style in the ArenaHard leaderboard. Notably, we observe a substantial gap for the 7B reward model depending on whether style control is enabled. This suggests that the 7B reward model is prone to reward hacking, e.g., improving ArenaHard scores primarily by increasing response length. We also examine the RLHF taining curves, and confirm that RLHF with the 7B reward model tends to improve reward scores by generating longer outputs, whereas training with the 72B reward model produces much more stable response lengths.
    \item RewardBench is a useful proxy but not always predictive of RLHF quality. While RewardBench scores correlate with reward model quality overall, higher RewardBench performance does not necessarily translate into better ArenaHard scores. 
    We hypothesize that RewardBench is relatively saturated (typically above 90), so marginal gains beyond this level do not meaningfully improve downstream helpfulness. In contrast, model-specific behaviors, such as vulnerability to reward hacking, play a more decisive role in determining RLHF effectiveness.
    \item Larger reward models also improve performance on other tasks such as math. For example, RLHF trained with the 72B reward model achieves around 3\% higher AIME25 accuracy than training with the 7B reward model. For code benchmarks, the choice of reward model has minimal impact, with performance differences within 1\%.
\end{enumerate}

\subsection{Bag of Tricks for Stablizing RLHF Training}
\label{abl:rewardshaping}

Although RL algorithms are crucial for enabling long-form chain-of-thought reasoning, RL training can be unstable and susceptible to early collapse. This issue is further amplified in RLHF, where training depends on model-based rewards that may be noisy or out-of-distribution.
In this subsection, we summarize the set of techniques (``bag of tricks'') we found effective for stabilizing RLHF training:

\begin{enumerate}
    \item \textbf{KL penalty loss:} 
    The KL penalty loss constrains the divergence between the online policy and the frozen reference policy, ensuring that the policy does not drift too far from the initial model. We find that when RLHF training collapses early, introducing this KL term is an effective way to maintain training stability.
    
    \item \textbf{Policy gradient loss aggregation:} Standard GRPO uses a \textit{sequence-level loss}, where token-level losses are first averaged within each sample and then aggregated across the batch. For long-CoT RL, \textit{token-level loss}, where all token losses in the batch are averaged directly, is typically recommended. However, when RLHF shows signs of early collapse, switching from token-level to sequence-level aggregation helps suppress significant increases in response length and stabilizes training.

    \item \textbf{Reward shaping:} Because our reward model is trained with a Bradley–Terry objective, its raw reward signals are unbounded. When training RLHF with unbounded rewards, noisy or outlier rewards can lead to unstable training. 
    To address this, we design a reward-shaping mechanism: for each group of rewards, we compute the mean and standard deviation, and then normalize each reward by subtracting the mean and dividing by the standard deviation, producing a centered and scaled reward. Finally, we apply a \texttt{tanh} transformation. This bounds the shaped rewards within \([-1, 1]\), effectively mitigating the impact of outliers and noisy reward signals within the group and leading to more stable RLHF updates.
\end{enumerate}

\begin{table}[t!]
\centering
\renewcommand{\arraystretch}{1.15}
\footnotesize
\caption{RLHF training results for AceReason-Nemotron-1.0-7B using the 7B reward model. ArenaHard (SC) refers to ArenaHard score with style control. We select the best checkpoint for each method from its training trajectory before collapse and report the corresponding training step.}
\vspace{-.2cm}
\label{tab:7b_reward_shaping}
\begin{adjustbox}{width={1\textwidth}}
\begin{tabular}{l|l|cc|cc|cc}
\toprule
\textbf{Techniques} &
\textbf{Step} &
\textbf{ArenaHard} &
\textbf{ArenaHard (SC)} &
\textbf{LCB v5} &
\textbf{LCB v6} &
\textbf{AIME24} &
\textbf{AIME25} \\
\midrule
KL=1e-3 + token-level loss & 350 & 
46.63 & 43.05 & 50.81 & 47.47 & 64.58 & 52.55 \\

KL=1e-3 + Seq-level loss & 550 &
48.71 & 46.60 & 50.13 & 47.38 & 65.99 & 53.23 \\

KL=1e-3 + Seq-level loss + Reward shaping & 950 &
50.78 & 45.76 & 50.76 & 47.96 & 65.05 & 51.20 \\
\bottomrule
\end{tabular}
\end{adjustbox}
\end{table}

\begin{table}[t!]
\centering
\renewcommand{\arraystretch}{1.15}
\footnotesize
\caption{RLHF training results for our unified 8B SFT model~(performance reported in Table~\ref{tab:results_after_sft}) using the 72B reward model. ArenaHard (SC) refers to ArenaHard score with style control. We select the best checkpoint for each method from its training trajectory before collapse and report the corresponding training step.}
\vspace{-.2cm}
\label{tab:72b_reward_shaping}
\begin{adjustbox}{width={1\textwidth}}
\begin{tabular}{l|l|cc|cc|cc}
\toprule
\textbf{Techniques} &
\textbf{Step} &
\textbf{ArenaHard} &
\textbf{ArenaHard (SC)} &
\textbf{LCB v5} &
\textbf{LCB v6} &
\textbf{AIME24} &
\textbf{AIME25} \\
\midrule
KL=1e-3 + Seq-level loss + Reward shaping & 500 & 90.03 & 89.11  &	68.59 &	65.66 &	86.20 &	73.80 \\
KL=0 + Token-level loss  & 600 & 91.04	& 89.37	& 68.46 & 65.86 & 86.33 & 75.03 \\
\bottomrule
\end{tabular}
\end{adjustbox}
\end{table}

In our early experiments with RLHF using the 7B reward model, we found that applying these “bag-of-tricks’’ techniques significantly improved training stability, extending the number of stable RL steps from 350 to 950, and led to better ArenaHard scores (Table~\ref{tab:7b_reward_shaping}). However, when using a stronger reward model (e.g., the 72B reward model), RLHF training is already stable, and omitting these techniques yields comparable—and in some cases slightly better—downstream performance than using them as shown in Table \ref{tab:72b_reward_shaping}. Our takeaway is that these techniques should be viewed as a toolbox to deploy only when training shows signs of instability. Otherwise, the RLHF recipe described in \S~\ref{subsec:rlhf_recipe} is sufficient.

\section{Deep Dive on SWE}
\label{sec:deepdive_swe}
In this section, we present the improved techniques for SWE tasks and provide the corresponding ablation results.

\subsection{Generation–Retrieval Approach for Code Localization}  
For the file-localization stage, we adopt a dual approach that combines generation-based and retrieval-based methods. In the generation-based approach, the model is guided to infer potentially buggy files based on the issue description and repository structure, as shown in Appendix~\ref{appendix:swe_template}.
To further enhance this method, we aggregate results from multiple rollouts and rank candidate files by their frequency of appearance, with higher-ranked files appearing more consistently across rollouts~\citep{wang2023selfconsistency}.

However, this generation-based approach only has access to the repository structure (i.e., folder and file names) rather than code contents. 
To complement this, we employ a code embedding model, NV-Embed-Code \citep{sohrabizadehnemotron}, to retrieve candidate files whose code contents are semantically similar to the problem context.\footnote{\url{https://build.nvidia.com/nvidia/nv-embedcode-7b-v1}}
The final set of relevant files is then determined by aggregating the results from both approaches using reciprocal rank fusion \citep{rrf} with the hyperparameter $k$ set to 0, which effectively integrates the complementary strengths of the two localization signals.

\begin{figure}[t!]
\begin{subfigure}{0.49\columnwidth}
\includegraphics[width=\columnwidth]{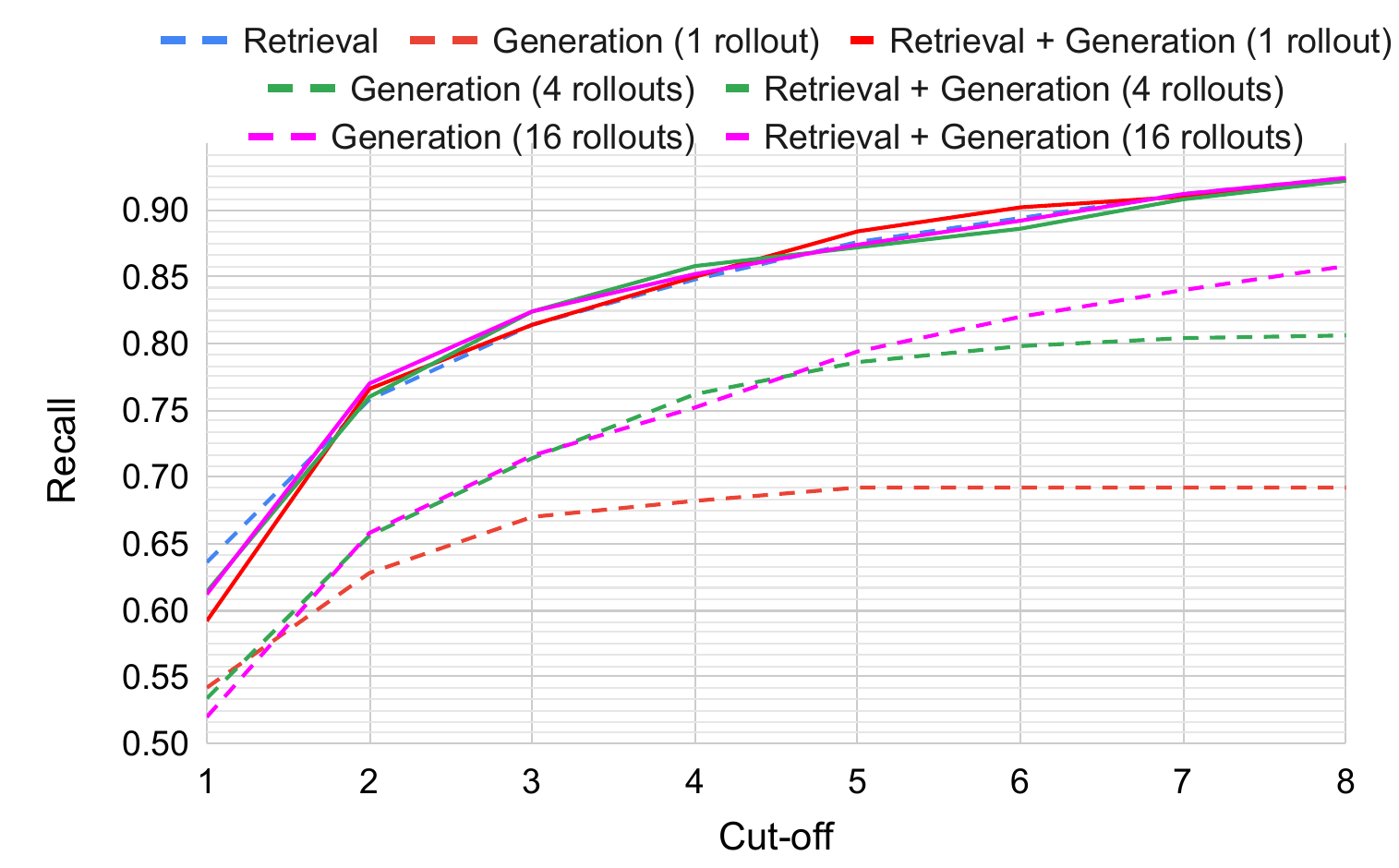}
\captionsetup{justification=centering}
\caption{Nemotron-Cascade-8B}
\end{subfigure}
\begin{subfigure}{0.49\columnwidth}
\includegraphics[width=\columnwidth]{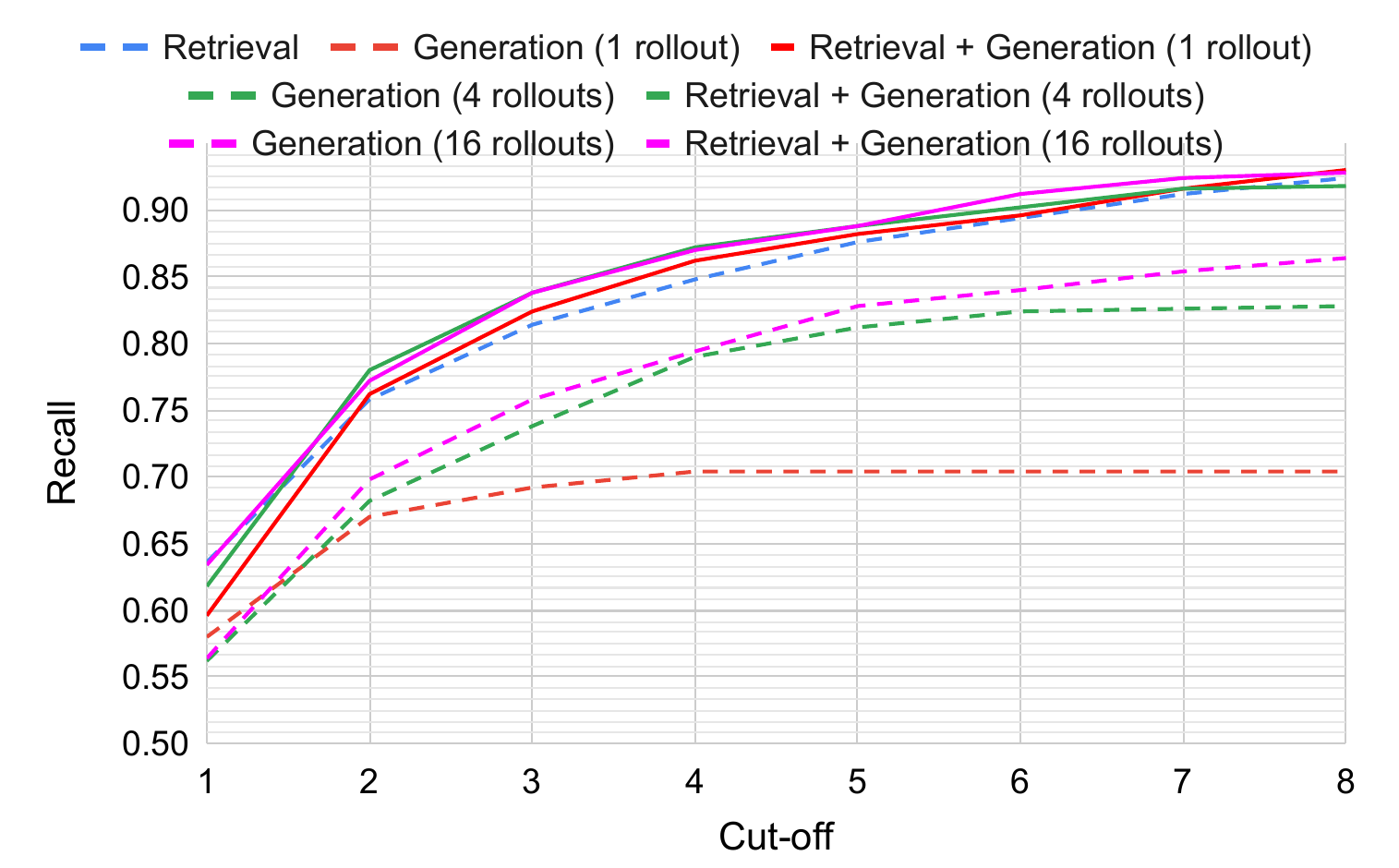}
\captionsetup{justification=centering}
\caption{Nemotron-Cascade-14B-Thinking}
\end{subfigure}
% \vspace{-0.3cm}
\caption{Ablation studies on code localization. The retrieval-based method uses NV-Embed-Code, while the generation-based method prompts the model to reason and generate the final ranking list (see Appendix \ref{appendix:swe_template}).}
\label{fig:ablation_code_loc}
% \vspace{-0.5cm}
\end{figure}

To evaluate code localization performance, we measure recall at various cutoffs~(top-$k$). Specifically, a localization is considered successful (recall = 1) for a instance if all ground-truth files requiring fixes appear within the top-$k$ retrieved candidates; otherwise, the recall is defined as 0 for this instance. 
Figure~\ref{fig:ablation_code_loc} illustrates the performance of different approaches for code localization on the SWE-bench Verified benchmark.

First, we observe that the retrieval-based method outperforms the generation-based ones.
This improvement is likely because the retrieval-based approach encodes the full source code content of each repository, whereas the generation-based approach relies only on repository structure when identifying potentially relevant files. 
Second, the generation-based approach demonstrates consistent gains at both top and higher ranks when results from multiple rollouts are aggregated. This indicates that aggregation not only improves top-ranking accuracy but also promotes ranking diversity in code localization. 
Finally, combining the generation- and retrieval-based methods with reciprocal rank fusion yields slight additional improvements, particularly at cutoffs below 5. 
In all our experiments, we directly use reciprocal fusion from generation- (with 16 rollouts) and retrieval-based methods as default.

\subsection{Execution-Free Reward Model for SWE RL}
\label{subsec:ablation_swe_rl_reward_model}
As mentioned in \S~\ref{subsubsec:swe_rl_training_recipe}, we use a execution-free reward defined in \Cref{equation:swe_rl_reward} for code repair RL training. 
That is, given a human-written golden patch, we compare its similarity to the model-generated patch using either lexical similarity—computed with the \emph{Unidiff} library following \citet{wei2025swe}—or a semantic similarity score produced by the Kimi-Dev-72B model.

In our ablation study, we compare these two approaches to compute the similarity.
We initialize from one of our intermediate 14B models without math and code RL (cond. 0 in Table~\ref{tab:swe_reward_model_ablation}) and conduct RL training for code repair using different similarity scores as reward functions. 
We follow the hyperameters in \S~\ref{subsubsec:swe_rl_training_recipe} except for setting rollouts to 8, and for the ablation of reward models, we use the training data with the maximum prompt length of 24K. 
We evaluate the trained models under two settings: \emph{i}) when ground-truth localized files are provided in the prompt, and \emph{ii}) when the top-4 localized files are obtained from generation–retrieval method.

\begin{table}[!t]
\centering
\footnotesize
\renewcommand{\arraystretch}{1.15}
\caption{Ablation results for different reward functions and reward shaping strategies on SWE-bench. Init.\ model is our intermediate 14B model without any code or math RL training.}
\label{tab:swe_reward_model_ablation}
% \begin{adjustbox}{width={0.9\textwidth}}
\begin{tabular}{c|cc|cc|cc}
\toprule
\multirow{2}{*}{\textbf{Cond.}} 
  & \multicolumn{2}{c|}{\textbf{Reward Function}} 
  & \multicolumn{2}{c|}{\textbf{Repair w/ Ground-truth Loc.}} 
  & \multicolumn{2}{c}{\textbf{Repair w/ Top-4 Loc.}} \\
\cline{2-7}
 & \textbf{Similarity} & \textbf{Reward shaping} 
 & \textbf{avg@4} & \textbf{pass@4} 
 & \textbf{avg@4} & \textbf{pass@4} \\
\midrule
0 & \multicolumn{2}{c|}{Init. 14B model} 
  & 41.7\% & 55.8\% & 38.8\% & 54.4\% \\
\midrule
1 & \multirow{2}{*}{Lexical Sim.} & no 
  & 42.6\% & 57.0\% & 41.2\% & 54.0\% \\
2 & & yes 
  & 42.8\% & 58.0\% & 41.6\% & 55.2\% \\
\midrule
3 & \multirow{2}{*}{Semantic Sim.} & no 
  & \textbf{43.0\%} & \textbf{58.6\%} & \textbf{42.3\%} & 56.4\% \\
4 & & yes 
  & 42.9\% & 57.0\% & 42.1\% & \textbf{56.8\%} \\
\bottomrule
\end{tabular}
% \end{adjustbox}
\end{table}

For the reward model based on semantic similarity, we directly apply the original reward function defined in \cref{equation:swe_rl_reward}. 
For lexical similarity, we replace \( s_{\text{sem}}(\hat{p}, p^*) \) with \( s_{\text{lex}}(\hat{p}, p^*) \) in this reward function. 
Table~\ref{tab:swe_reward_model_ablation} reports the resolve rates averaged over four runs with sampling temperature set to 0.6, along with pass@4; that is, an instance is considered resolved if it is successfully fixed in at least one of the four generations. 

First, we observe that RL training generally enhances the model’s effectiveness in code repair, and using semantic similarity as the reward model yields better effectiveness than using lexical similarity (cond. 4 vs. 2).
Second, we apply reward shaping to both reward models by setting the reward to 0 when it falls below 0.5. This adjustment improves the effectiveness of the lexical similarity reward model (cond. 2 vs. 1), suggesting that reward shaping helps filter out noisy supervision signals.
In particular, when lexical similarity is below 0.5, the reward tends to provide unreliable guidance for model training.
However, we do not observe the same effect when applying reward shaping to semantic similarity (cond. 4 vs. 3), indicating that semantic similarity continues to provide meaningful training signals even when code similarity is low. 
As a result, we apply the default reward function setup (cond. 3 in Table~\ref{tab:swe_reward_model_ablation}) for SWE RL training.

Overall, we demonstrate that using an LLM-based, execution-free reward model is a promising direction for scaling SWE RL training.
We leave the exploration of reward model training as future work.

\subsection{Improving Long-Context Analysis} 
\label{subsec:ablation_swe_rl_max_prompt_length}
To ensure prompts including all buggy code patches, we form a long prompt with code contents from multiple retrieved files. 
However, our preliminary study shows that the code resolve rate drops significantly when the input prompt length exceeds 24K, alongside a response length of 16K.
We hypothesize the suboptimal code resolve rate is due to the 32K maximum sequence length used during SFT, inherited from the 32K context window of the Qwen3-8B/14B-Base models.
Thus, at RL stage, we create training data with longer prompts by mixing models' retrieved noisy files and ground-truth files.
% \red{This is related to the ablation of training with 16K, 24K, 32K prompts}

\begin{table}[!t]
\centering
\footnotesize
\renewcommand{\arraystretch}{1.15}
\caption{Ablation studies on SWE RL for code repair using training data with various maximum prompt lengths. 
Init.\ model is our intermediate 14B model without any code or math RL training.}
\label{tab:swe_prompt_length_ablation}
% \begin{adjustbox}{width={0.85\textwidth}}
\begin{tabular}{c|c|cc|cc}
\toprule
\multirow{2}{*}{\textbf{Cond.}} &
\multirow{2}{*}{\textbf{Max Prompt Len.}} &
\multicolumn{2}{c|}{\textbf{Repair w/ Ground-truth Loc.}} &
\multicolumn{2}{c}{\textbf{Repair w/ Top-4 Loc.}} \\
\cline{3-6}
& & \textbf{avg@4} & \textbf{pass@4} & \textbf{avg@4} & \textbf{pass@4} \\
\midrule
0 & Init. 14B model & 41.7\% & 55.8\% & 38.8\% & 54.4\% \\
\midrule
1 & 16K & 44.0\% & \textbf{58.8\%} & 41.6\% & 54.6\% \\
2 & 24K & 43.0\% & 58.6\% & 42.3\% & \textbf{56.4\%} \\
3 & 32K & \textbf{44.1\%} & 57.6\% & \textbf{42.7\%} & 56.2\% \\
4 & 40K & 42.8\% & 57.6\% & 41.5\% & 55.4\% \\
\bottomrule
\end{tabular}
% \end{adjustbox}
\end{table}

In Table~\ref{tab:swe_prompt_length_ablation}, we ablate training with different data which is created with various maximum prompt length (see more detail in \S~\ref{subsubsec:swe_rl_data_curation}).  
We observe that from 16K to 32K, training with longer prompts helps improve model's repairing capability. 
We contribute the improvement to model's capability to prompts with longer context, which is especially important in the repairing task since during repairing, the models' need to identify and repair buggy code patches among all the retrieved code contents. 
However, when extending the maximum prompt length to 40K, the training shows less effective. 
We hypothesize that the model performs worse under such long prompts, causing the sampled trajectories to contain more noise for RL training, or that the pretrained Qwen3-14B-Base has limited long-context capability at 32K.
As a result, for training data, we set the maximum prompt length to 24K and 32K for the final 8B and 14B models, respectively.

\subsection{Test-Time Scaling and Patch Validation}
To further improve code-repair accuracy, we employ a test-time scaling (TTS) strategy that enhances model performance by aggregating and filtering multiple candidate patches during inference. As outlined in Section~\S\ref{subsec:agentless_framework}, the model generates a diverse set of candidate repair patches and reproduction tests using temperature-based and top-$p$ decoding. Each candidate patch is then evaluated through a patch-validation stage that applies regression and reproduction tests to identify the most reliable fixes.

For SWE-bench Verified benchmark, our TTS pipeline generates $k$ candidate repair patches along with 40 reproduction tests per instance, then filters and ranks these candidates by first assessing how many existing regression tests each patch passes, followed by executing a curated subset of generated reproduction tests to identify the most promising repairs. The patch with the highest combined pass rate is ultimately selected, with ties resolved first by majority voting and then by minimal solution length. We refer to this ranking and selection procedure as best@$k$. This approach broadens the solution search space, enhances robustness by exploring multiple reasoning trajectories, and substantially increases the likelihood of producing a correct repair.

\begin{figure}[t!]
\begin{subfigure}{0.49\columnwidth}
\includegraphics[width=\columnwidth]{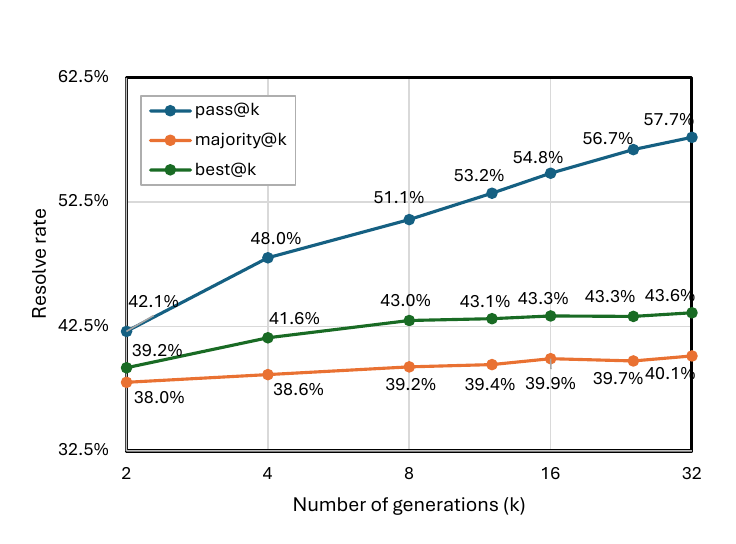}
\captionsetup{justification=centering}
\caption{Nemotron-Cascade-8B}
\end{subfigure}
\begin{subfigure}{0.49\columnwidth}
\includegraphics[width=\columnwidth]{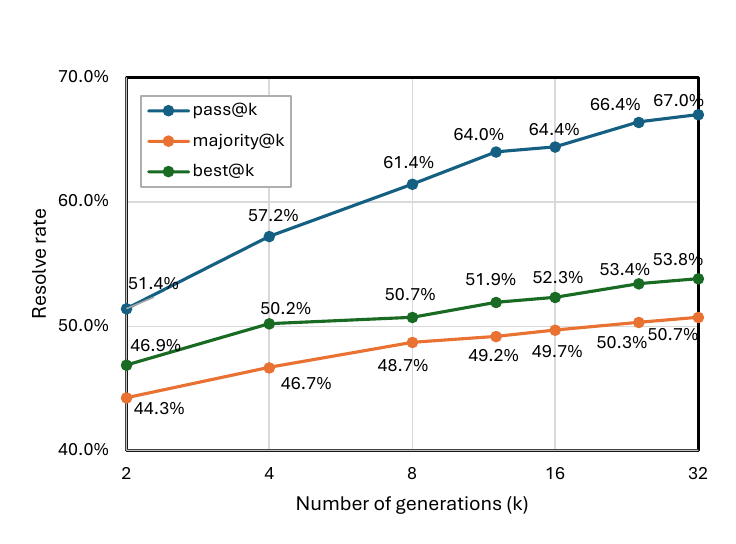}
\captionsetup{justification=centering}
\caption{Nemotron-Cascade-14B-Thinking}
\end{subfigure}
\vspace{-0.2cm}
\caption{Ablation studies on test-time-scaling (TTS) that our best@k approach and majority voting can substantially boost resolve rate on SWE-bench Verified.}
\label{fig:swe_tts}
% \vspace{-0.5cm}
\end{figure}

Figure~\ref{fig:swe_tts} presents results for (a) Nemotron-Cascade-8B and (b) Nemotron-Cascade-14B-Thinking evaluated on SWE-bench Verified using test-time scaling (TTS) combined with our patch-validation pipeline. 
As shown in Figure~\ref{fig:swe_tts}, we plot pass@$k$, majority@$k$, and best@$k$ across $k \in {2, 4, 8, 12, 16, 24, 32}$. Pass@$k$ improves monotonically with increasing $k$, while majority voting grows more slowly and saturates earlier. Best@$k$ consistently outperforms majority@$k$ by a clear margin, demonstrating the effectiveness of our patch-validation pipeline.
For Nemotron-Cascade-14B-Thinking, the improvements are more pronounced across all metrics, reflecting stronger reasoning ability and greater diversity in generated repair patches. Overall, both Nemotron-Cascade-8B and Nemotron-Cascade-14B-Thinking benefit substantially from the TTS strategy, with the 14B model achieving results competitive with larger open-weight models such as DeepSWE~\citep{deepswe}~(resolve rate: 52.4\% by performing TTS with execution-based verifier). These gains demonstrate that downstream filtering and validation remain powerful mechanisms for boosting patch-repair performance without modifying model weights.

As shown in Figure~\ref{fig:swe_tts}(a), Nemotron-Cascade-8B achieves a best@32 resolve rate of 43.6\%, improving from 39.2\% at $k=2$ and gradually increasing as more samples are considered. With TTS and patch validation, the model reaches a pass@$k$ score of 57.7\% at $k=32$, indicating a 15.6-point gap that reflects additional room for potential improvement toward best@32.
Majority voting provides a simpler alternative but plateaus around 39–40\%, showing only marginal gains as $k$ increases. These results demonstrate that, even for a smaller model, structured test-time scaling combined with validation can substantially enhance repair accuracy.
As shown in Figure~\ref{fig:swe_tts}(b), the overall metrics for Nemotron-Cascade-14B-Thinking improve more substantially. The model begins with a majority@$k$ resolve rate of 50.7\%, already surpassing the 8B variant’s best@32 score of 43.6\%. Under the TTS strategy, best@$k$ provides further gains, plateauing around 53.8\%. Moreover, the pass@$k$ curve continues to rise as $k$ increases, highlighting the considerable potential for developing more effective TTS strategies for the 14B model.

\section{Related Work}
\label{sec:relatedwork}

In this section, we briefly review related work and position our study within the existing literature.

\subsection{Reinforcement Learning for LLMs}
Reinforcement learning from human feedback~(RLHF) plays a critical role in further aligning large language models~(LLMs) following supervised fine-tuning or instruction tuning~\citep{ouyang2022training, bai2022training}.
Traditionally, a pretrained reward model is utilized to be the surrogate of human judge and provide instant reward signal in online training~\citep[e.g.,][]{wang2024helpsteer2, liu2024skywork}.
In contrast to teacher-forced training in supervised fine-tuning~(SFT), which requires high-quality and costly annotations, RLHF offers a more cost-effective and generalizable approach to capturing the subtleties of human intent and the nuances of linguistic expression.
More recently, reward-model-based reinforcement learning has also been explored for math reasoning~\citep[e.g.,][]{shao2024deepseekmath, yang2024qwen2_5math}. However, its success has been limited due to the inherent challenges of reward modeling in the mathematical domain~\citep{lightman2023let, wang2023math, zhang2025lessons}.

Large-scale reinforcement learning with verifiable rewards~(RLVR)—which employs objective and deterministic criteria~(e.g., symbolic rule-based verification for math reasoning) to provide reward signals—has achieved remarkable success in developing frontier reasoning models~\citep[e.g.,][]{guo2025deepseek, team2025kimi_k15, yang2025qwen3}.
Open RLVR recipes with publicly available datasets have been developed, such as AceReason-Nemotron~\citep{chen2025acereason, liu2025acereason}, DeepScaleR~\citep{deepscaler2025}, DeepCoder~\citep{deepcoder2025}, DAPO~\citep{yu2025dapo}, and Skywork-OR1~\citep{skywork-or1-2025}.
However, such open-recipe models focus primarily on math and code reasoning, differing from the general-purpose frontier models.

The RL training of general-purpose DeepSeek-R1 and Qwen3 follows a two-stage process: an initial reasoning-oriented RL stage, followed by a second stage covering all domains.
In each stage, diverse prompts are used for joint training. However, due to the substantial heterogeneity across tasks, this design complicates the RL infrastructure, training curriculum, and hyperparameter tuning, ultimately leading to suboptimal performance.

In this work, we present the Cascade RL framework and release open training recipes and datasets for developing general-purpose LLMs with strong reasoning capabilities across diverse domains, including math, coding, science, instruction following, software engineering, and general domain. In particular, we systematically investigate the interplay between RLHF and RLVR—a topic that has been underexplored in existing literature.

Various RL algorithms have been explored for both RLHF and RLVR, including PPO~\citep{schulman2017proximal}, DPO~\citep{rafailov2023direct}, GRPO~\citep{shao2024deepseekmath}, and their variants—such as on-policy methods (e.g., AceReason-Nemotron~\citep{chen2025acereason}) versus off-policy approaches (e.g., clipping strategies in~\citet{yu2025dapo, su2025klear})—as well as sample-level versus token-level policy gradient losses~\citep{yu2025dapo}.
Moreover, various techniques have been proposed to enhance the stability and efficiency of RL training, such as curriculum learning with gradually increasing maximum response lengths~\citep[e.g.,][]{deepscaler2025} and overlength filtering to mitigate excessive penalties from truncated generations under inference budgets~\citep{yu2025dapo}.

\subsection{Supervised Fine-Tuning and Distillation}
Supervised fine-tuning~(SFT) serves as an indispensable preparatory stage to adapt a pretrained base LLM for general conversation, instruction following, and a variety of other tasks prior to RL-based alignment~\citep{ouyang2022training, dubey2024llama, adler2024nemotron, dai2024nvlm}.
To build compact reasoning models, another approach is to distill large teacher models~\citep{guo2025deepseek, yang2025qwen3, deepseek_r1_0528}—originally trained via RL—into smaller ones.
A popular strategy is off-policy distillation~\citep{bercovich2025llama, moshkov2025aimo2, ahmad2025opencodereasoning, nano2025efficient}, in which synthetic responses or teacher outputs are first sampled from the teacher model, and the student model is then trained to predict the teacher-sampled tokens or logits.
Although generating synthetic data from large teacher models is computationally expensive, such methods enable efficient training once the SFT data have been created and allow the same data to be reused for training other models.
On-policy distillation, in which samples are generated from the student model, is also explored on top of off-policy distillation to reduce the performance gap to models trained with on-policy RL~\citep{yang2025qwen3}.

In general, RL is applied to SFT models to develop state-of-the-art reasoning models.
In our previous work~\citep{liu2025acereason}, we studied the synergy between SFT and RL. We found that the performance gap between initial SFT models narrows significantly during a well-designed RL process, provided that an appropriate balance between exploration and exploitation is achieved.

\subsection{Unified Reasoning Models}
Building general-purpose models with strong reasoning capabilities has been a central goal of recent LLM development. 
Over the past year, many dedicated \emph{thinking} models have been released, including OpenAI’s o1~\citep{openai2024o1}, o3, o4-mini~\citep{openai2025o3o4mini}, DeepSeek-R1~\citep{guo2025deepseek}, Qwen3-Thinking~\citep{Qwen3_235B_thinking}, MiniMax-M1~\citep{chen2025minimax}, gpt-oss~\citep{agarwal2025gpt}, and Kimi-K2-Thinking~\citep{Kimi_K2_Thinking}. 
These models emphasize deep reasoning through long chain-of-thought~(CoT) generation~\citep{yeo2025demystifying}, involving problem analysis, idea sketching, enumeration of alternative solution strategies, as well as verification and correction of answers.

Several recent efforts aim to unify \emph{instruct} and \emph{thinking} models into a single model.
Llama-Nemotron~\citep{bercovich2025llama} enables global control over the thinking or instruct mode through the system prompt.
Qwen3~\citep{yang2025qwen3}, GLM-4.5~\citep{zeng2025glm} and DeepSeek-V3.1~\citep{deepseek_v3_1} provide more flexible user control, allowing mode switching between \emph{thinking} and \emph{instruct} at each conversational turn.
GPT-5~\citep{openai2025gpt5} employs an automatic routing mechanism that circumvents, rather than resolves, this challenge.

% \newpage

\paragraph{\LARGE Appendix}
\vspace{0.3cm}
% \newpage

\appendix

\section{Acknowledgments}
\label{appendix:contributors}

We would like to extend our gratitude to the NVIDIA Nemo team for the valuable discussion and collaboration
on building reasoning models. We especially wish to thank Boris Ginsburg, Oleksii Kuchaiev, Igor Gitman, Olivier Delalleau, Zhilin Wang, Olivier Delalleau, Banghua Zhu, Tugrul Konuk, Wei Du, Somshubra Majumdar, Siddhartha Jain, Jiaqi Zeng, Yi Dong, Alexander Bukharin, Vahid Noroozi, Khushi Bhardwaj, Sugam Dipak Devare, Jian Zhang, and Jonathan Cohen.

We thank Ying Lin for helpful discussions and useful input in building the knowledge-intensive SFT dataset.
We also thank Atefeh Sohrabizadeh, Jialin Song, and Jonathan Raiman for valuable discussions on SWE-bench.

\section{Benchmarks and Evaluation Setups}
\label{appendix:benchmarks}

For knowledge reasoning tasks, we include:
\begin{itemize}[leftmargin=2em]
    \item
    \textbf{MMLU}~\citep{hendrycks2020measuring} is a benchmark designed to assess an LLM’s broad world knowledge and problem-solving ability. It contains 14,079 test questions across 57 subjects spanning STEM, the humanities, social sciences, and professional domains such as law and ethics. For both unified reasoning model and dedicated thinking model, we evaluate the models in \emph{thinking} mode and, due to the large test set size, report exact match~(EM) accuracy based on a single generation per question.
    \item 
    \textbf{MMLU-Pro}~\citep{wang2024mmlu} is an enhanced version of the original MMLU benchmark that mitigates model saturation by expanding to over 12,000 graduate-level questions and increasing answer choices from four to ten. We report EM accuracy in \emph{thinking} mode using one generation per question.
    \item 
    \textbf{GPQA-Diamond}~\citep{rein2024gpqa} is a benchmark for assessing an LLM’s scientific reasoning capability. It consists of the highest quality 198 GPQA questions covering graduate-level physics, biology, and chemistry. We report pass@1 accuracy in \emph{thinking} mode, averaged over 8 generations per question (avg@8) to reduce variance.
\end{itemize}
For Nemotron-Cascade models evaluated on MMLU, MMLU-Pro, and GPQA-Diamond in \emph{thinking} mode, we use a temperature of 0.6, a top-p value of 0.95, and a 64K-token thinking budget (maximum response length) with a YaRN scaling factor~\citep{peng2023yarn} of 2.

For alignment tasks, we include:
\begin{itemize}[leftmargin=2em]
    \item
    \textbf{IFEval}~\citep{zhou2023instructionfollowingevaluationlargelanguage} is a benchmark for evaluating an LLM’s instruction-following capability, focusing on verifiable instructions.
    It contains 541 prompts and 25 verifiable instructions, with each prompt including one or more instructions. We use \emph{prompt strict}, which measures the percentage of prompts where all instructions are satisfied. In contrast, prior work~\citep{nano2025efficient} adopts \emph{instruct strict}, which measures the percentage of individual instructions that are satisfied. We evaluate the unified reasoning model in \emph{non-thinking} mode, and the dedicated thinking model in \emph{thinking} mode.
    We report pass@1 accuracy, using an average over 8 generations per question (avg@8).
    \item \textbf{IFBench}~\citep{pyatkin2025generalizingverifiableinstructionfollowing} extends IFEval by introducing 58 new, diverse, and challenging verifiable out-of-domain instruction constraints. It provides a separate constraint list to ensure no overlap between training and test constraints, enabling evaluation of an LLM’s generalization ability. The test set contains 294 prompts. We report pass@1 accuracy in \emph{thinking} mode, averaged over 8 generations (avg@8).
    \item 
    \textbf{ArenaHard 1.0}~\citep{li2024crowdsourced} is a human-preference alignment benchmark consisting of 500 diverse and challenging real user prompts. It uses an automatic LLM-as-Judge approach to estimate human preferences relative to a baseline model, enabling fully automated, low-cost, and fast evaluation without human intervention.  In our experiments, we report results without style control to allow for straightforward comparison with the officially reported numbers of other models. We evaluate the models in \emph{thinking} mode, and use GPT4-Turbo-2024-0409 as the judge. 
\end{itemize}
For Nemotron-Cascade models evaluated on IFEval in \emph{non-thinking} mode, on IFBench and ArenaHard in \emph{thinking} mode, we use a temperature of 0.6, a top-p value of 0.95, and a maximum response length of 32K tokens. 
% For ArenaHard in \emph{thinking} mode, we increase the thinking budget to 32K tokens while keeping all other hyperparameters the same. 
For baseline models, we use officially reported results whenever available; if such results are absent, we evaluate them using their recommended inference configuration or the same settings as ours.

For math reasoning tasks, we include 
\begin{itemize}[leftmargin=2em]
    \item 
    \textbf{AIME 2024}~\citep{aime2024} consists of 30 problems from American Invitational Mathematics Examination at 2024.
    \item
    \textbf{AIME 2025}~\citep{aime2025} consists of 30 problems from American Invitational Mathematics Examination at 2025.
\end{itemize}
For Nemotron-Cascade models on AIME 2024 and 2025, we set the thinking budget (maximum response length) to 64K tokens, the sampling temperature to 0.6, the top-p value to 0.95, and the YaRN scaling factor to 2. For baseline models, we follow their recommended inference settings with a thinking budget of at least 64K tokens.

For code generation tasks, we include
\begin{itemize}[leftmargin=2em]
    \item 
    \textbf{LiveCodeBench}~\citep{jain2024livecodebench} contains diverse algorithm coding problems with unit tests, collected from AtCoder, LeetCode platforms. We evaluate models competitive coding capability on its two subsets: LiveCodeBench v5~(2024/08/01-2025/02/01, \textbf{279} problems in total) and v6~(2024/08/01-2025/05/01, \textbf{454} problems in total). We report pass@1 accuracy in \emph{thinking} mode, averaged over 8 generations (avg@8).
    \item 
    \textbf{LiveCodeBench Pro}~\citep{zheng2025livecodebench} contains daily-updated challenging competitive coding problems with strong unit tests, collected mainly from top-tier coding contests. We report pass@1 accuracy on Easy/Med difficulty splits in \emph{thinking} mode, averaged over 8 generations (avg@8) on two recently released subsets: 2025Q1 (2025/01-2025/04, \textbf{166} problems in total) and 2025Q2 (2025/04-2025/07, \textbf{167} problems in total).
    \item 
    \textbf{SWE-bench Verified}~\citep{openai2024swe_verified} is a subset of the original test set from SWE-bench~\citep{jimenez2023swe}, consisting of 500 samples verified to be non-problematic by human annotators. We evaluate models in \emph{thinking} mode and report pass@1 accuracy, averaged over 4 generations per prompt~(avg@4).
\end{itemize}

For Nemotron-Cascade models evaluated on LiveCodeBench (v5/v6) and LiveCodeBench Pro, we use a 64K-token thinking budget, a sampling temperature of 0.6, a top-p of 0.95, and a YaRN scaling factor of 2. 
We evaluate baseline models with their recommended inference configurations, ensuring a thinking budget of at least 64K tokens.
For SWE-bench Verified, we use 32K-token thinking budget and a sampling temperature of 0.6. 
We set maximum input prompt length to 32K and 64K tokens, and set YaRN scaling factor of 2 and 3 for 8B and 14B models, respectively.

\section{Prompt Templates}
\subsection{Unpreferrable Response Generation for RM data}
\label{appendix:rm_data_gen_template}
% \paragraph{Prompts for offtopic generation}
\vspace{.2cm}
\textbf{Step 1: Generate offtopic prompts}

\begin{promptbox}
% \#\#\#\#\# Step 1: Generate offtopic prompts \#\#\#\#\#
Given an user input (called "given input"), please generate a new user input (called "generated input") such that: \\

(1) The generated input is highly relevant to but different from the given input.

(2) The correct response to the generated input superficially resembles the correct response to the given input as much as possible.

(3) But actually, the correct response to the generated input should not be a correct response to the given input. \\

Given input:

\texttt{\{instruction\}}\\

Generated input:

\end{promptbox}

\textbf{Step 2: Judge if the offtopic prompts are really different to the original}

\begin{promptbox}
There are two instructions, Instruction A and Instruction B. Are the two instructions asking the same thing? Please answer in `YES` or `NO`.

\# Instruction A: \\
\texttt{\{instruction A\}}\\

\# Instruction B: \\
\texttt{\{instruction B\}}\\

\end{promptbox}

\subsection{Prompts and Templates for SWE Task}
\label{appendix:swe_template}
\paragraph{Code Localization}
\begin{promptbox}
Please look through a given GitHub issue and repository structure and provide a list of files that one would need to edit or look at to solve the issue. \\ \\
\#\#\# GitHub Problem Description \#\#\#

\texttt{\{problem\_statement\}}

\#\#\# \\ \\
\#\#\# Repository Structure \#\#\# \\
\texttt{\{structure\}}\\ \\
\#\#\# \\

Below are some code segments, each from a relevant file. One or more of these files may contain bugs.

Only provide the full path and return at most {n} files. The returned files should be separated by new lines ordered by most to least important and wrapped with \texttt{`}\texttt{`}\texttt{`}. For example:\\ \\
\texttt{`}\texttt{`}\texttt{`}\\
most/important/file1.xx \\
less/important/file2.yy \\
least/important/file3.zz \\
\texttt{`}\texttt{`}\texttt{`}
\end{promptbox}

\paragraph{Code Repair}

\begin{promptbox}
We are currently solving the following issue within our repository. Here is the issue text:

--- BEGIN ISSUE ---

\texttt{\{problem\_statement\}}

--- END ISSUE ---

Below are some code segments, each from a relevant file. One or more of these files may contain bugs.

--- BEGIN FILE ---
\begin{lstlisting}
{content}
\end{lstlisting}
--- END FILE ---

Please first localize the bug based on the issue statement, and then generate \textbf{SEARCH/REPLACE} edits to fix the issue.

Every \textbf{SEARCH/REPLACE} edit must use this format:
\begin{enumerate}
\item Start with \texttt{```diff\textbackslash{}n} to indicate a diff block, and end the whole block with \texttt{```}.
\item The file path
\item The start of search block: \texttt{<<<<<<< SEARCH}
\item A contiguous chunk of lines to search for in the existing source code
\item The dividing line: \texttt{=======}
\item The lines to replace into the source code
\item The end of the replace block: \texttt{>>>>>>> REPLACE}
\end{enumerate}

Here is an example:

\begin{lstlisting}[language=Python]
```diff
### mathweb/flask/app.py
<<<<<<< SEARCH
from flask import Flask
=======
import math
from flask import Flask
>>>>>>> REPLACE
```
\end{lstlisting}

Please note that the \textbf{SEARCH/REPLACE} edit REQUIRES PROPER INDENTATION. If you would like to add the line \texttt{'        print(x)'}, you must fully write that out, with all those spaces before the code!

Wrap each \textbf{SEARCH/REPLACE} edit in a code block as shown in the example above. If you have multiple \textbf{SEARCH/REPLACE} edits, use a separate code block for each one.

Output format requirement: Please put your reasoning tokens in a separate code block, starting with \texttt{<think>} and ending with \texttt{</think>}, and the solution tokens in a separate code block, starting with \texttt{<solution>} and ending with \texttt{</solution>}.
\end{promptbox}

\paragraph{Test Code Generation}
%@Chankyue

\begin{promptbox}
We are currently solving the following issue within our repository. Here is the issue text:

--- BEGIN ISSUE ---
%\begin{lstlisting}

\texttt{\{problem\_statement\}}
%\end{lstlisting}

--- END ISSUE ---

Several candidate repair patches have been generated to address this issue. You must carefully examine them and select the one that best matches the issue description when creating the test so that it specifically validates the behavior before and after applying the patch:

--- BEGIN PATCH ---
\begin{lstlisting}
{model_patch}
\end{lstlisting}
--- END PATCH ---

Below are some code segments, each from a relevant file. One or more of these files may contain bugs.

--- BEGIN FILE ---
\begin{lstlisting}
{content}
\end{lstlisting}
--- END FILE ---

Please generate a complete test that can be used to reproduce the issue.

The complete test should contain the following:
\begin{enumerate}
\item Includes all necessary imports
\item Reproduces the issue described in the issue text before the patch is applied
\item Exercises the exact functions, classes, or lines modified in the repair patch
\item Contains assertions or checks that confirm the issue is reproduced without the patch
\item Contains assertions or checks that confirm the issue is resolved after the patch is applied
\item Uses meaningful assertions tied to the patch changes (e.g., expected outputs, raised exceptions, or altered return values)
\item Print "Issue reproduced" if the outcome indicates that the issue is reproduced
\item Print "Issue resolved" if the outcome indicates that the issue has been successfully resolved
\item Print "Other issues" if the outcome indicates there are other issues with the source code
\end{enumerate}

The test should not be generic; it must directly validate the correctness of the patch.  

Here is an example:
\begin{lstlisting}[language=Python]
```python
from sqlfluff import lint

def test__rules__std_L060_raised() -> None:
    try:
        sql = "SELECT   IFNULL(NULL, 100),
            NVL(NULL,100);"
        result = lint(sql, rules=["L060"])
        assert len(result) == 2
    except:
        print("Other issues")
        return

    try:
        assert result[0]["description"] == "Use 'COALESCE' instead of 'IFNULL'."
        assert result[1]["description"] == "Use 'COALESCE' instead of 'NVL'."
        print("Issue resolved")
    except AssertionError:
        print("Issue reproduced")
        return

    return

test_rules_std_L060_raised()
\end{lstlisting}
\end{promptbox}

\paragraph{Reward Modeling}
\begin{promptbox}
\textbf{System Prompt}\\
You are an expert judge evaluating AI assistant interactions. Your task is to determine if the assistant successfully resolved the user's request given a reference golden solution.\\ \\ 
Key evaluation criteria: \\
1. Did the assistant complete the main task requested by the user? \\
2. Were there any errors or issues in the final solution? \\ \\
Respond only with "<judgement>YES</judgement>" or "<judgement>NO</judgement>". \\

\textbf{User Prompt}\\
We are currently solving the following issue within our repository. Here is the issue text:

--- BEGIN ISSUE ---

\texttt{\{problem\_statement\}}

--- END ISSUE ---

Below are some code segments, each from a relevant file. One or more of these files may contain bugs.

--- BEGIN FILE ---
\begin{lstlisting}
{content}
\end{lstlisting}
--- END FILE ---

Please first localize the bug based on the issue statement, and then generate \textbf{SEARCH/REPLACE} edits to fix the issue.

Every \textbf{SEARCH/REPLACE} edit must use this format:
\begin{enumerate}
\item Start with \texttt{```diff\textbackslash{}n} to indicate a diff block, and end the whole block with \texttt{```}.
\item The file path
\item The start of search block: \texttt{<<<<<<< SEARCH}
\item A contiguous chunk of lines to search for in the existing source code
\item The dividing line: \texttt{=======}
\item The lines to replace into the source code
\item The end of the replace block: \texttt{>>>>>>> REPLACE}
\end{enumerate}

Here is the reference golden git diff solution: \\ \\
\texttt{\{golden\_patch\}} \\ \\
Here is the solution from the assistance: \\ \\
\texttt{\{model\_patch\}} \\ \\
Please compare the assistance's solution to the reference golden git diff solution and judge whether the assistance's solution successfully resolve the issue. Note that the solution is not required to be exactly the same as reference golden solution. Use your own knowledge to judge whether the assistance's solution successfully resolve the issue. Respond with "<judgement>YES</judgement>" or "<judgement>NO</judgement>".
\end{promptbox}

% !!!!!Example:

% \begin{promptbox}
% <|im\_start|>system\\
% You are a helpful and harmless assistant. \\ \\
% \# Tools \\ \\
% You may call one or more functions to assist with the user query. \\ \\
% You are provided with function signatures within \emph{<tools>}\emph{</tools>} XML tags: \\
% \emph{<tools>} \\
% \{"type": "function", "function": \{"name": "FUNCTION\_NAME\_1", "description": "DESCRIPTION\_1", "parameters": \{ ... \} \} \} \\
% \{"type": "function", "function": \{"name": "FUNCTION\_NAME\_2", "description": "DESCRIPTION\_2", "parameters": \{ ... \} \} \} \\
% ... ... \\
% \emph{</tools>} \\ \\
% For each function call, return a json object with function name and arguments within \emph{<tool\_call>}\emph{</tool\_call>} XML tags:\\
% \emph{<tool\_call>}\\
% \{"name": <function-name>, "arguments": <args-json-object>\}\\
% \emph{</tool\_call>}\\
% <|im\_start|>user\\
% Book a hotel in San Francisco. /think
% \end{promptbox}

\subsection{Prompt Templates for Test-Time Scaling on IOI 2025}
\label{appendix:cp_template}

\begin{promptbox}
Write Python code to solve the problem. Please place the solution code in the following format:\texttt{\\```python\\ \# Your solution code here\\```}
\\
\texttt{\{problem\_statement\}}

Below you are provided the accepted correct solutions but with different input constraints. You may use them as a reference for your insights.
\\ 
=======================
\\ 
\#\# Different Constraints (for reference only):
\\ 
\texttt{\{subtask\_constraints\}}
\\ 
\#\#\# Accepted Code:
\\ 
\texttt{[CODE]}
\\
=======================
\\
\#\# Different Constraints (for reference only):
\\ 
...
\\ 
=======================
\\ 
From here, you are also given your submission history containing **incorrect** code and their corresponding official judgement verdicts as reference -- Official judgement verdicts and problem statement/conditions are 100\% reliable. You should make improvements from them if they could help:
\\
=======================
\\
\#\#\# Incorrect Code
\\ 
\texttt{[CODE]}
\\
Judgement Verdict: \texttt{[VERDICT]}, Score: \texttt{[SCORE]}
\\
=======================
\\
\#\#\# Incorrect Code
\\
... 
\\ 
=======================
\end{promptbox}

\section{Training Hyperparameters}
\label{appendix:training_hyperparams}

\subsection{Multi-Stage SFT}
We list the hyperparameters for the multi-stage SFT of the 8B and 14B models in Table~\ref{tab:hyperparams_sft}.

\begin{table}[!htbp]
\centering
\caption{Training hyperparameters of 8B and 14B models in \textbf{multi-stage SFT}. Both the \emph{unified} and the \emph{thinking} models share the same hyperparameters.}
\vspace{2pt}
\begin{adjustbox}{width={.8\textwidth}}
\setlength{\tabcolsep}{20pt} % Default value: 6pt
\renewcommand{\arraystretch}{1.5} % Default value: 1
\begin{tabular}{lcc}
\toprule
Hyperparameters                   & 8B (Stage-1 / Stage-2)                         &  14B (Stage-1 / Stage-2)           \\  \midrule
Global batch size                  & $256$                        &  $256$         \\
Max learning rate                  & $5e^{-5}$ / $2e^{-5}$        &  $5e^{-5}$          \\
Min learning rate                  & $0$                      &  $0$          \\
% Learning rate warmup steps         & 500                            &  1,000          \\
Scheduler                          & cosine                         & cosine       \\
Max Steps                          & 100K                         & 100K       \\
Optimizer                          & AdamW                          & AdamW          \\
Optimizer config                   & $\beta_1=0.9$, $\beta_2=0.95$  &    $\beta_1=0.9$, $\beta_2=0.95$       \\
Weight decay                       & $1e^{-4}$                          &     $1e^{-4}$             \\
\# of training steps               & 20K / 32K                            & 22K / 41K     \\ 
\bottomrule
\end{tabular}
\end{adjustbox}
\label{tab:hyperparams_sft}
\end{table}

\subsection{RLHF}
We present the RLHF hyperparameters for the 8B and 14B models in Table~\ref{tab:hyperparams_rlhf_8B_14B}.

\begin{table}[!hbp]
\centering
\caption{Training hyperparameters of 8B/14B  models in \textbf{RLHF}. Both the \emph{unified} and \emph{thinking} models share the same hyperparameters. \emph{Unified} models are trained in both \emph{non-thinking} and \emph{thinking} modes, with an equal split of prompts allocated to each mode.}
\vspace{2pt}
\begin{adjustbox}{width={.75\textwidth}}
\setlength{\tabcolsep}{20pt} % Default value: 6pt
\renewcommand{\arraystretch}{1.5} % Default value: 1
\begin{tabular}{lcc}
\toprule
Hyperparameters & 8B  & 14B   \\ 
\midrule
% Add rows below
Max response length & 12K &  12K \\
Batch size & 256 & 256 \\
\# Rollout size & 8 & 8  \\
Learning rate & $2e^{-6}$ & $2e^{-6}$ \\
Steps & 800 & 900  \\
Optimizer            & AdamW          & AdamW        \\
Optimizer config     & $\beta_1=0.9$, $\beta_2=0.95$  &    $\beta_1=0.9$, $\beta_2=0.95$ \\
Temperature & 0.6 & 0.6  \\
Top-p & 0.95 & 0.95  \\
Overlong filtering & False & False  \\
\bottomrule
\end{tabular}
\end{adjustbox}
\label{tab:hyperparams_rlhf_8B_14B}
\end{table}

\subsection{IF-RL}
The hyperparameters of 8B and 14B models in \textbf{IF-RL} training are in Table~\ref{tab:hyperparams_ifrl_8B_14B}.

\begin{table}[!hbp]
\centering
\caption{Training hyperparameters of 8B/14B  models in \textbf{IF-RL}. \emph{Unified} models are trained in the \emph{non-thinking} mode.}
\vspace{2pt}
\begin{adjustbox}{width={1\textwidth}}
\setlength{\tabcolsep}{20pt} % Default value: 6pt
\renewcommand{\arraystretch}{1.5} % Default value: 1
\begin{tabular}{lccc}
\toprule
Hyperparameters & 8B (unified)  & 8B-Thinking & 14B-Thinking \\ 
\midrule
% Add rows below
Max response length (stage 1) & 8K &  8K  & 8K \\
Max response length (stage 2) & 8K &  16K & 16K \\
Batch size & 256 & 256 & 256 \\
\# Rollout size & 8 & 8 & 8 \\
Learning rate & $2e^{-6}$ & $2e^{-6}$ & $2e^{-6}$ \\
Steps (stage 1) & 2300 & 550 & 800 \\
Steps (stage 2) & 800 & 300  & 120 \\
Optimizer            & AdamW          & AdamW  & AdamW      \\
Optimizer config     & $\beta_1=0.9$, $\beta_2=0.95$  &    $\beta_1=0.9$, $\beta_2=0.95$ & $\beta_1=0.9$, $\beta_2=0.95$ \\
Temperature & 0.6 & 0.6 & 0.6 \\
Top-p & 0.95 & 0.95 & 0.95 \\
Overlong filtering (stage 1) & False & False & False \\
Overlong filtering (stage 2) & False & True & True \\
\bottomrule
\end{tabular}
\end{adjustbox}
\label{tab:hyperparams_ifrl_8B_14B}
\end{table}

\subsection{Math RL}
The hyperparameters for the 8B and 14B models used in \textbf{Math RL} training are listed in Table~\ref{tab:hyperparams_mathrl_8B} and Table~\ref{tab:hyperparams_mathrl_14B}, respectively.

\begin{table}[!hbp]
\centering
\caption{Training hyperparameters of our 8B models in \textbf{Math RL}. Both the \emph{unified} and \emph{thinking} models share the same hyperparameters. Unified models are trained in the \emph{thinking} mode. }
\vspace{2pt}
\begin{adjustbox}{width={.95\textwidth}}
\setlength{\tabcolsep}{20pt} % Default value: 6pt
\renewcommand{\arraystretch}{1.5} % Default value: 1
\begin{tabular}{lccc}
\toprule
Hyper-parameters & 8B (Stage-1) & 8B (Stage-2) & 8B (Stage-3) \\ 
\midrule
% Add rows below
Max response length & 24K & 32K & 40K\\
Batch size & 128 & 128 & 128\\
\# Rollout size & 8 & 8 & 8 \\
Learning rate & $2e^{-6}$ & $2e^{-6}$ & $2e^{-6}$ \\
Steps (start-end) & 0-190 & 190-430 & 430-500 \\
Optimizer            & AdamW          & AdamW           & AdamW          \\
Optimizer config     & $\beta_1=0.9$, $\beta_2=0.95$  &    $\beta_1=0.9$, $\beta_2=0.95$  &  $\beta_1=0.9$, $\beta_2=0.95$   \\
Temperature & 1 & 1  & 0.8 \\
Top-p & 0.95 & 0.95 & 0.95 \\
Overlong filtering & True & False & False \\
\bottomrule
\end{tabular}
\end{adjustbox}
\label{tab:hyperparams_mathrl_8B}
\end{table}

\begin{table}[!hbp]
\centering
\caption{Training hyperparameters of 14B-Thinking  model in \textbf{Math RL}.}
\vspace{2pt}
\begin{adjustbox}{width={.75\textwidth}}
\setlength{\tabcolsep}{20pt} % Default value: 6pt
\renewcommand{\arraystretch}{1.5} % Default value: 1
\begin{tabular}{lcc}
\toprule
Hyper-parameters & 14B (Stage-1) & 14B (Stage-2)  \\ 
\midrule
% Add rows below
Max response length & 28K &  40K \\
Batch size & 128 & 128 \\
\# Rollout size & 8 & 8  \\
Learning rate & $2.5e^{-6}$ & $2.5e^{-6}$ \\
Steps (start-end) & 0-120 & 120-220  \\
Optimizer            & AdamW          & AdamW        \\
Optimizer config     & $\beta_1=0.9$, $\beta_2=0.95$  &    $\beta_1=0.9$, $\beta_2=0.95$ \\
Temperature & 1.2 & 1.1  \\
Top-p & 0.95 & 0.95  \\
Overlong filtering & True & False  \\
\bottomrule
\end{tabular}
\end{adjustbox}
\label{tab:hyperparams_mathrl_14B}
\end{table}

\subsection{Code RL}
The hyperparameters of 8B-Thinking,  8B unified and 14B-Thinking models in \textbf{Code RL} are in Table~\ref{tab:hyperparams_coderl_8B_14B}. 

\begin{table}[!hbp]
\centering
\caption{Training hyperparameters of 8B-Thinking, 8B unified, and 14B-Thinking models in \textbf{Code RL}. 8B unified model are trained in the \emph{thinking} mode. }
\vspace{2pt}
{\small
\setlength{\tabcolsep}{20pt} % Default value: 6pt
\renewcommand{\arraystretch}{1.5} % Default value: 1
\begin{tabular}{lccc}
\toprule
Hyper-parameters & 8B-Thinking & 8B (unified)  & 14B-Thinking  \\ 
\midrule
% Add rows below
Max response length & 44k & 44K~$\rightarrow$~48K & 56K  \\
Batch size & 128 & 128 & 128  \\
\# Rollout size & 8 & 8 & 8  \\
Learning rate & $4e^{-6}$ & $4e^{-6}$ & $4e^{-6}$ \\
Steps  & 64 & 90 & 64 \\
Optimizer            & AdamW          & AdamW      & AdamW            \\
Optimizer config     & $\beta_1=0.9$, $\beta_2=0.95$  &    $\beta_1=0.9$, $\beta_2=0.95$   &    $\beta_1=0.9$, $\beta_2=0.95$ \\
Temperature & 1.2 & 1.0~$\rightarrow$~0.8 & 1.0   \\
Top-p & 0.95 & 0.95  & 0.95 \\
Overlong filtering & False & False  & False \\
\bottomrule
\end{tabular}
}
\label{tab:hyperparams_coderl_8B_14B}
\end{table}

\subsection{SWE RL}
The hyperparameters for the 8B unified, 8B-Thinking, and 14B-Thinking models used in \textbf{SWE RL} training are listed in Table~\ref{tab:hyperparams_swerl_8B_14B}.

\begin{table}[!hbp]
\centering
\caption{Training hyperparameters of 8B unified, 8B-Thinking, and 14B-Thinking models in \textbf{SWE RL}. Both the 8B \emph{unified} and Thinking models share the same hyperparameters. Unified models are trained in the \emph{thinking} mode.}
\vspace{2pt}
\begin{adjustbox}{width={.95\textwidth}}
\setlength{\tabcolsep}{20pt} % Default value: 6pt
\renewcommand{\arraystretch}{1.5} % Default value: 1
\begin{tabular}{lccc}
\toprule
Hyper-parameters & 8B (Stage-1) & 8B (Stage-2)  & 14B (single stage)  \\ 
\midrule
% Add rows below
Input response length & 16K &  24K  & 32K \\
Max response length & 16K &  16K & 16K \\
Batch size & 128 & 128  &  128\\
\# Rollout size & 16 & 16  & 16 \\ 
Learning rate & $2.5e^{-6}$ & $2.5e^{-6}$ & $2.5e^{-6}$ \\ 
Steps (start-end) & 0-30 & 30-60 & 0-120  \\
Optimizer            & AdamW          & AdamW      & AdamW  \\ 
Optimizer config     & $\beta_1=0.9$, $\beta_2=0.95$  &    $\beta_1=0.9$, $\beta_2=0.95$ & $\beta_1=0.9$, $\beta_2=0.95$ \\
Temperature & 1 & 1 & 1  \\
Top-p & 0.95 & 0.95   & 0.95  \\
Overlong filtering & True & False  & True  \\
\bottomrule
\end{tabular}
\end{adjustbox}
\label{tab:hyperparams_swerl_8B_14B}
\end{table}

\section{ELO Rating Analysis}
\label{appendix:elo_rating_on_codeforces}

In this section, we present the details of the reported Codeforces Elo ratings for the Nemotron-Cascade-8B and Nemotron-Cascade-14B-Thinking models, based on \textbf{51} recent Codeforces contests held between 2501–2507. Problems and evaluations are provided by LiveCodeBench Pro~\citep{zheng2025livecodebench}. For each contest, we simulate participation by allowing the model up to $N$ independent submissions per problem (with $N$ set to 8 by default) and generate the model’s responses using a temperature of 0.6, \texttt{top-p} of 0.95, and a maximum token budget of 128K. Let $k$ denote the number of correct submissions among these $N$ attempts, and $N - k$ the number of incorrect submissions ($0 \leq k \leq N$). In a real contest, submissions are made sequentially and the penalty submission counts is defined by the number of incorrect submissions prior to the first correct submission. To estimate the submission penalty, we assume the ordering of the $k$ correct and $N - k$ incorrect submissions are uniformly distributed over $\binom{N}{k}$ permutations and the expected number of penalties can be derived as: $$\mathbb{E}[\text{\# of penalties}] = \frac{N - k}{k + 1}$$

We adopt the standard codeforces contest rules: for regular codeforces round, we apply score penalty as 50 for each expected penalty, and for ICPC style round (e.g. educational rounds, Div.3 rounds), we add time penalty as 10 for each incorrect instead. Penalties on problems which remained unsolved will not take into consideration. We rank our model's contest performance against $n$ real human participants as $m$ ($1 \leq m \leq n + 1$) based on the final score, and compute implied performance rating $R_{\text{model}}$ following standard Elo rating definition~\citep{glickman1999rating, quan2025codeelo} by solving: $$m = \sum_{i=1}^n \frac{1}{1 + 10^{(R_{\text{model}} - R_i)/400}}$$
where $R_i$ refers to the Elo rating of human contestant $i$ before each contest. We report the averaged performance rating over 51 codeforces rounds as our final Elo score and present the performance details of our Nemotron-Cascade-8B and Nemotron-Cascade-14B-Thinking model in Table~\ref{tab:elo_rating_8b} and Table~\ref{tab:elo_rating_14b}, respectively.

We observe large variance in the model’s estimated performance rating across contests. For instances, the Nemotron-Cascade-14B-Thinking model achieves the estimated performance rating above 2600 on Codeforces Round 1015, yet fails to solve any problems (even with 8 attempts) and receives Elo rating below 1000 on Round 1024 Div.1. We also find inconsistent behavior on coding problem solving: while the model is sometimes able to solve very difficult problems, it can also become stuck on relatively easy ones, even within the same contest. Furthermore, the model tends to perform well on problems solvable by standard techniques, heavy implementation, or straightforward intuition, but often struggles on problems that require hypothesis-driven exploration on small-scale data or ad hoc ideas, such as constructive or interactive problems. It could be an interesting direction for understanding and improving such reasoning abilities in the future.

\newcolumntype{P}[1]{>{\centering\arraybackslash}m{#1}}

% \usepackage{multirow}
\begin{table}[!htbp]
\centering
\caption{Nemotron-Cascade-8B performance details on 51 Codeforces Rounds ranging from 2501-2507. We attempt each problem with $N=8$ times in total. For regular codeforces rounds, we present the score after considering expected penalties for each problem. For ICPC style rounds, we mark passed/failed problems as \textcolor{ForestGreen}{+} and \textcolor{BrickRed}{-} correspondingly. We compute the estimated rank to human contestants and the corresponding Elo score as shown in rightmost two columns.}
\scalebox{0.47}{
\begin{tabular}{c| *{11}{P{1.1cm}} |c c c c}
\toprule
\bfseries Contest Name & \multicolumn{11}{c|}{\bfseries Contest Problems} &
\bfseries Score & \bfseries Penalty & \bfseries Est. Rank & \bfseries ELO \\ \midrule
\multirow{2}{*}{Hello 2025} & \bf A  & \bf B  & \bf C  & \bf D  & \bf E1  & \bf E2  & \bf F  & \bf G  & \bf H  &  &  &  \multirow{2}{*}{2893.75} & \multirow{2}{*}{-} & \multirow{2}{*}{1695/16703} & \multirow{2}{*}{1942} \\
&  \color{ForestGreen} 493.75 &  \color{ForestGreen} 1000.00 &  \color{ForestGreen} 1400.00 & \color{BrickRed} 0.0 & \color{BrickRed} 0.0 & \color{BrickRed} 0.0 & \color{BrickRed} 0.0 & \color{BrickRed} 0.0 & \color{BrickRed} 0.0  &  &  &  \\ \hline
\multirow{2}{*}{Codeforces Round 996 (Div. 2)} & \bf A  & \bf B  & \bf C  & \bf D  & \bf E  & \bf F  &  &  &  &  &  &  \multirow{2}{*}{2931.43} & \multirow{2}{*}{-} & \multirow{2}{*}{749/21232} & \multirow{2}{*}{1979} \\
&  \color{ForestGreen} 460.00 &  \color{ForestGreen} 985.71 &  \color{ForestGreen} 1485.71 & \color{BrickRed} 0.0 & \color{BrickRed} 0.0 & \color{BrickRed} 0.0  &  &  &  &  &  &  \\ \hline
\multirow{2}{*}{Codeforces Round 997 (Div. 2)} & \bf A  & \bf B  & \bf C  & \bf D  & \bf E  & \bf F1  & \bf F2  &  &  &  &  &  \multirow{2}{*}{2735.71} & \multirow{2}{*}{-} & \multirow{2}{*}{2025/18823} & \multirow{2}{*}{1695} \\
& \color{BrickRed} 0.0 &  \color{ForestGreen} 1250.00 &  \color{ForestGreen} 1485.71 & \color{BrickRed} 0.0 & \color{BrickRed} 0.0 & \color{BrickRed} 0.0 & \color{BrickRed} 0.0  &  &  &  &  &  \\ \hline
\multirow{2}{*}{Codeforces Round 998 (Div. 3)} & \bf A  & \bf B  & \bf C  & \bf D  & \bf E  & \bf F  & \bf G  &  &  &  &  &  \multirow{2}{*}{6} & \multirow{2}{*}{23.00} & \multirow{2}{*}{22/24247} & \multirow{2}{*}{1699} \\
& \color{ForestGreen} + & \color{ForestGreen} + & \color{ForestGreen} + & \color{ForestGreen} + & \color{ForestGreen} + & \color{ForestGreen} + & \color{BrickRed} -  &  &  &  &  &  \\ \hline
\multirow{2}{*}{IAEPC Preliminary Contest (Codeforces Round 999, Div. 1 + Div. 2)} & \bf A  & \bf B  & \bf C  & \bf D  & \bf E  & \bf F1  & \bf F2  & \bf G  & \bf H1  & \bf H2  & \bf I  &  \multirow{2}{*}{4256.25} & \multirow{2}{*}{-} & \multirow{2}{*}{1200/12647} & \multirow{2}{*}{1988} \\
&  \color{ForestGreen} 500.00 &  \color{ForestGreen} 937.50 &  \color{ForestGreen} 1493.75 &  \color{ForestGreen} 1325.00 & \color{BrickRed} 0.0 & \color{BrickRed} 0.0 & \color{BrickRed} 0.0 & \color{BrickRed} 0.0 & \color{BrickRed} 0.0 & \color{BrickRed} 0.0 & \color{BrickRed} 0.0  &  \\ \hline
\multirow{2}{*}{Codeforces Round 1000 (Div. 2)} & \bf A  & \bf B  & \bf C  & \bf D  & \bf E  & \bf F1  & \bf F2  &  &  &  &  &  \multirow{2}{*}{7804.46} & \multirow{2}{*}{-} & \multirow{2}{*}{14/17169} & \multirow{2}{*}{2200} \\
&  \color{ForestGreen} 500.00 &  \color{ForestGreen} 985.71 &  \color{ForestGreen} 1500.00 &  \color{ForestGreen} 2243.75 &  \color{ForestGreen} 2575.00 & \color{BrickRed} 0.0 & \color{BrickRed} 0.0  &  &  &  &  &  \\ \hline
\multirow{2}{*}{Ethflow Round 1 (Codeforces Round 1001, Div. 1 + Div. 2)} & \bf A  & \bf B  & \bf C  & \bf D  & \bf E1  & \bf E2  & \bf F  & \bf G  & \bf H  &  &  &  \multirow{2}{*}{2337.50} & \multirow{2}{*}{-} & \multirow{2}{*}{2336/16234} & \multirow{2}{*}{1771} \\
&  \color{ForestGreen} 500.00 &  \color{ForestGreen} 900.00 &  \color{ForestGreen} 937.50 & \color{BrickRed} 0.0 & \color{BrickRed} 0.0 & \color{BrickRed} 0.0 & \color{BrickRed} 0.0 & \color{BrickRed} 0.0 & \color{BrickRed} 0.0  &  &  &  \\ \hline
\multirow{2}{*}{Codeforces Round 1002 (Div. 2)} & \bf A  & \bf B  & \bf C  & \bf D  & \bf E1  & \bf E2  &  &  &  &  &  &  \multirow{2}{*}{1325.00} & \multirow{2}{*}{-} & \multirow{2}{*}{4240/19443} & \multirow{2}{*}{1454} \\
&  \color{ForestGreen} 500.00 &  \color{ForestGreen} 825.00 & \color{BrickRed} 0.0 & \color{BrickRed} 0.0 & \color{BrickRed} 0.0 & \color{BrickRed} 0.0  &  &  &  &  &  &  \\ \hline
\multirow{2}{*}{Codeforces Round 1003 (Div. 4)} & \bf A  & \bf B  & \bf C1  & \bf C2  & \bf D  & \bf E  & \bf F  & \bf G  & \bf H  &  &  &  \multirow{2}{*}{9} & \multirow{2}{*}{45.86} & \multirow{2}{*}{1/28033} & \multirow{2}{*}{1522} \\
& \color{ForestGreen} + & \color{ForestGreen} + & \color{ForestGreen} + & \color{ForestGreen} + & \color{ForestGreen} + & \color{ForestGreen} + & \color{ForestGreen} + & \color{ForestGreen} + & \color{ForestGreen} +  &  &  &  \\ \hline
\multirow{2}{*}{Codeforces Round 1004 (Div. 1)} & \bf A  & \bf B  & \bf C  & \bf D1  & \bf D2  & \bf E  & \bf F  &  &  &  &  &  \multirow{2}{*}{2375.00} & \multirow{2}{*}{-} & \multirow{2}{*}{252/1030} & \multirow{2}{*}{2482} \\
& \color{BrickRed} 0.0 &  \color{ForestGreen} 650.00 &  \color{ForestGreen} 1075.00 &  \color{ForestGreen} 650.00 & \color{BrickRed} 0.0 & \color{BrickRed} 0.0 & \color{BrickRed} 0.0  &  &  &  &  &  \\ \hline
\multirow{2}{*}{Codeforces Round 1004 (Div. 2)} & \bf A  & \bf B  & \bf C  & \bf D  & \bf E  & \bf F  & \bf G  &  &  &  &  &  \multirow{2}{*}{4225.00} & \multirow{2}{*}{-} & \multirow{2}{*}{174/16749} & \multirow{2}{*}{2098} \\
&  \color{ForestGreen} 500.00 & \color{BrickRed} 0.0 & \color{BrickRed} 0.0 & \color{BrickRed} 0.0 &  \color{ForestGreen} 1650.00 &  \color{ForestGreen} 2075.00 & \color{BrickRed} 0.0  &  &  &  &  &  \\ \hline
\multirow{2}{*}{Codeforces Round 1005 (Div. 2)} & \bf A  & \bf B  & \bf C  & \bf D  & \bf E  & \bf F  &  &  &  &  &  &  \multirow{2}{*}{900.00} & \multirow{2}{*}{-} & \multirow{2}{*}{9718/17621} & \multirow{2}{*}{1021} \\
& \color{BrickRed} 0.0 &  \color{ForestGreen} 900.00 & \color{BrickRed} 0.0 & \color{BrickRed} 0.0 & \color{BrickRed} 0.0 & \color{BrickRed} 0.0  &  &  &  &  &  &  \\ \hline
\multirow{2}{*}{Educational Codeforces Round 174 (Rated for Div. 2)} & \bf A  & \bf B  & \bf C  & \bf D  & \bf E  & \bf F  &  &  &  &  &  &  \multirow{2}{*}{3} & \multirow{2}{*}{4.11} & \multirow{2}{*}{1298/16701} & \multirow{2}{*}{1810} \\
& \color{ForestGreen} + & \color{ForestGreen} + & \color{ForestGreen} + & \color{BrickRed} - & \color{BrickRed} - & \color{BrickRed} -  &  &  &  &  &  &  \\ \hline
\multirow{2}{*}{Codeforces Round 1006 (Div. 3)} & \bf A  & \bf B  & \bf C  & \bf D  & \bf E  & \bf F  & \bf G  &  &  &  &  &  \multirow{2}{*}{7} & \multirow{2}{*}{21.25} & \multirow{2}{*}{1/24140} & \multirow{2}{*}{1699} \\
& \color{ForestGreen} + & \color{ForestGreen} + & \color{ForestGreen} + & \color{ForestGreen} + & \color{ForestGreen} + & \color{ForestGreen} + & \color{ForestGreen} +  &  &  &  &  &  \\ \hline
\multirow{2}{*}{Educational Codeforces Round 175 (Rated for Div. 2)} & \bf A  & \bf B  & \bf C  & \bf D  & \bf E  & \bf F  &  &  &  &  &  &  \multirow{2}{*}{4} & \multirow{2}{*}{2.86} & \multirow{2}{*}{234/16060} & \multirow{2}{*}{2195} \\
& \color{ForestGreen} + & \color{ForestGreen} + & \color{ForestGreen} + & \color{ForestGreen} + & \color{BrickRed} - & \color{BrickRed} -  &  &  &  &  &  &  \\ \hline
\multirow{2}{*}{Codeforces Round 1007 (Div. 2)} & \bf A  & \bf B  & \bf C  & \bf D1  & \bf D2  & \bf E  & \bf F  &  &  &  &  &  \multirow{2}{*}{5487.50} & \multirow{2}{*}{-} & \multirow{2}{*}{4/16254} & \multirow{2}{*}{2198} \\
&  \color{ForestGreen} 500.00 &  \color{ForestGreen} 937.50 & \color{BrickRed} 0.0 &  \color{ForestGreen} 1725.00 & \color{BrickRed} 0.0 &  \color{ForestGreen} 2325.00 & \color{BrickRed} 0.0  &  &  &  &  &  \\ \hline
\multirow{2}{*}{Codeforces Round 1008 (Div. 1)} & \bf A  & \bf B  & \bf C  & \bf D  & \bf E  & \bf F  & \bf G  &  &  &  &  &  \multirow{2}{*}{1931.25} & \multirow{2}{*}{-} & \multirow{2}{*}{371/909} & \multirow{2}{*}{2294} \\
&  \color{ForestGreen} 437.50 & \color{BrickRed} 0.0 &  \color{ForestGreen} 1493.75 & \color{BrickRed} 0.0 & \color{BrickRed} 0.0 & \color{BrickRed} 0.0 & \color{BrickRed} 0.0  &  &  &  &  &  \\ \hline
\multirow{2}{*}{Codeforces Round 1008 (Div. 2)} & \bf A  & \bf B  & \bf C  & \bf D  & \bf E  & \bf F  & \bf G  &  &  &  &  &  \multirow{2}{*}{6806.25} & \multirow{2}{*}{-} & \multirow{2}{*}{9/14641} & \multirow{2}{*}{2008} \\
&  \color{ForestGreen} 500.00 &  \color{ForestGreen} 725.00 &  \color{ForestGreen} 1187.50 &  \color{ForestGreen} 1650.00 & \color{BrickRed} 0.0 &  \color{ForestGreen} 2743.75 & \color{BrickRed} 0.0  &  &  &  &  &  \\ \hline
\multirow{2}{*}{Codeforces Round 1009 (Div. 3)} & \bf A  & \bf B  & \bf C  & \bf D  & \bf E  & \bf F  & \bf G  &  &  &  &  &  \multirow{2}{*}{5} & \multirow{2}{*}{22.86} & \multirow{2}{*}{178/23635} & \multirow{2}{*}{1708} \\
& \color{ForestGreen} + & \color{ForestGreen} + & \color{ForestGreen} + & \color{ForestGreen} + & \color{BrickRed} - & \color{BrickRed} - & \color{ForestGreen} +  &  &  &  &  &  \\ \hline
\multirow{2}{*}{Educational Codeforces Round 176 (Rated for Div. 2)} & \bf A  & \bf B  & \bf C  & \bf D  & \bf E  & \bf F  &  &  &  &  &  &  \multirow{2}{*}{4} & \multirow{2}{*}{41.25} & \multirow{2}{*}{73/18159} & \multirow{2}{*}{2198} \\
& \color{ForestGreen} + & \color{BrickRed} - & \color{ForestGreen} + & \color{ForestGreen} + & \color{ForestGreen} + & \color{BrickRed} -  &  &  &  &  &  &  \\ \hline
\multirow{2}{*}{Codeforces Round 1011 (Div. 2)} & \bf A  & \bf B  & \bf C  & \bf D  & \bf E  & \bf F1  & \bf F2  &  &  &  &  &  \multirow{2}{*}{4587.50} & \multirow{2}{*}{-} & \multirow{2}{*}{170/15906} & \multirow{2}{*}{2200} \\
&  \color{ForestGreen} 500.00 &  \color{ForestGreen} 1187.50 &  \color{ForestGreen} 1075.00 & \color{BrickRed} 0.0 & \color{BrickRed} 0.0 &  \color{ForestGreen} 1825.00 & \color{BrickRed} 0.0  &  &  &  &  &  \\ \hline
\multirow{2}{*}{Codeforces Round 1012 (Div. 1)} & \bf A  & \bf B1  & \bf B2  & \bf C1  & \bf C2  & \bf D  & \bf E  &  &  &  &  &  \multirow{2}{*}{0} & \multirow{2}{*}{-} & \multirow{2}{*}{653/653} & \multirow{2}{*}{977} \\
& \color{BrickRed} 0.0 & \color{BrickRed} 0.0 & \color{BrickRed} 0.0 & \color{BrickRed} 0.0 & \color{BrickRed} 0.0 & \color{BrickRed} 0.0 & \color{BrickRed} 0.0  &  &  &  &  &  \\ \hline
\multirow{2}{*}{Codeforces Round 1012 (Div. 2)} & \bf A  & \bf B  & \bf C  & \bf D  & \bf E1  & \bf E2  & \bf F1  & \bf F2  &  &  &  &  \multirow{2}{*}{1460.00} & \multirow{2}{*}{-} & \multirow{2}{*}{2197/8536} & \multirow{2}{*}{1466} \\
&  \color{ForestGreen} 500.00 &  \color{ForestGreen} 960.00 & \color{BrickRed} 0.0 & \color{BrickRed} 0.0 & \color{BrickRed} 0.0 & \color{BrickRed} 0.0 & \color{BrickRed} 0.0 & \color{BrickRed} 0.0  &  &  &  &  \\ \hline
\multirow{2}{*}{Codeforces Round 1013 (Div. 3)} & \bf A  & \bf B  & \bf C  & \bf D  & \bf E  & \bf F  & \bf G  &  &  &  &  &  \multirow{2}{*}{5} & \multirow{2}{*}{0.00} & \multirow{2}{*}{852/24379} & \multirow{2}{*}{1715} \\
& \color{ForestGreen} + & \color{ForestGreen} + & \color{ForestGreen} + & \color{ForestGreen} + & \color{ForestGreen} + & \color{BrickRed} - & \color{BrickRed} -  &  &  &  &  &  \\ \hline
\multirow{2}{*}{Codeforces Round 1014 (Div. 2)} & \bf A  & \bf B  & \bf C  & \bf D  & \bf E  & \bf F  &  &  &  &  &  &  \multirow{2}{*}{6400.00} & \multirow{2}{*}{-} & \multirow{2}{*}{2/15842} & \multirow{2}{*}{2213} \\
&  \color{ForestGreen} 500.00 &  \color{ForestGreen} 750.00 &  \color{ForestGreen} 1250.00 &  \color{ForestGreen} 1650.00 &  \color{ForestGreen} 2250.00 & \color{BrickRed} 0.0  &  &  &  &  &  &  \\ \hline
\multirow{2}{*}{Teza Round 1 (Codeforces Round 1015, Div. 1 + Div. 2)} & \bf A  & \bf B  & \bf C  & \bf D  & \bf E  & \bf F  & \bf G1  & \bf G2  & \bf H  &  &  &  \multirow{2}{*}{4885.71} & \multirow{2}{*}{-} & \multirow{2}{*}{489/11206} & \multirow{2}{*}{2320} \\
&  \color{ForestGreen} 750.00 &  \color{ForestGreen} 1000.00 &  \color{ForestGreen} 1485.71 &  \color{ForestGreen} 1650.00 & \color{BrickRed} 0.0 & \color{BrickRed} 0.0 & \color{BrickRed} 0.0 & \color{BrickRed} 0.0 & \color{BrickRed} 0.0  &  &  &  \\ \hline
\multirow{2}{*}{Codeforces Round 1016 (Div. 3)} & \bf A  & \bf B  & \bf C  & \bf D  & \bf E  & \bf F  & \bf G  &  &  &  &  &  \multirow{2}{*}{6} & \multirow{2}{*}{36.25} & \multirow{2}{*}{67/21249} & \multirow{2}{*}{1699} \\
& \color{ForestGreen} + & \color{ForestGreen} + & \color{ForestGreen} + & \color{ForestGreen} + & \color{ForestGreen} + & \color{ForestGreen} + & \color{BrickRed} -  &  &  &  &  &  \\ \hline
\multirow{2}{*}{Codeforces Round 1017 (Div. 4)} & \bf A  & \bf B  & \bf C  & \bf D  & \bf E  & \bf F  & \bf G  & \bf H  &  &  &  &  \multirow{2}{*}{8} & \multirow{2}{*}{52.50} & \multirow{2}{*}{1/22234} & \multirow{2}{*}{1503} \\
& \color{ForestGreen} + & \color{ForestGreen} + & \color{ForestGreen} + & \color{ForestGreen} + & \color{ForestGreen} + & \color{ForestGreen} + & \color{ForestGreen} + & \color{ForestGreen} +  &  &  &  &  \\ \hline
\multirow{2}{*}{Neowise Labs Contest 1 (Codeforces Round 1018, Div. 1 + Div. 2)} & \bf A  & \bf B  & \bf C  & \bf D  & \bf E  & \bf F  & \bf G  & \bf H  &  &  &  &  \multirow{2}{*}{2723.21} & \multirow{2}{*}{-} & \multirow{2}{*}{1382/12771} & \multirow{2}{*}{1929} \\
&  \color{ForestGreen} 485.71 &  \color{ForestGreen} 743.75 &  \color{ForestGreen} 1493.75 & \color{BrickRed} 0.0 & \color{BrickRed} 0.0 & \color{BrickRed} 0.0 & \color{BrickRed} 0.0 & \color{BrickRed} 0.0  &  &  &  &  \\ \hline
\multirow{2}{*}{Codeforces Round 1019 (Div. 2)} & \bf A  & \bf B  & \bf C  & \bf D  & \bf E  & \bf F  &  &  &  &  &  &  \multirow{2}{*}{2953.75} & \multirow{2}{*}{-} & \multirow{2}{*}{849/14465} & \multirow{2}{*}{1903} \\
&  \color{ForestGreen} 500.00 &  \color{ForestGreen} 993.75 &  \color{ForestGreen} 1460.00 & \color{BrickRed} 0.0 & \color{BrickRed} 0.0 & \color{BrickRed} 0.0  &  &  &  &  &  &  \\ \hline
\multirow{2}{*}{Codeforces Round 1020 (Div. 3)} & \bf A  & \bf B  & \bf C  & \bf D  & \bf E  & \bf F  & \bf G1  & \bf G2  &  &  &  &  \multirow{2}{*}{5} & \multirow{2}{*}{43.00} & \multirow{2}{*}{311/17451} & \multirow{2}{*}{1708} \\
& \color{ForestGreen} + & \color{ForestGreen} + & \color{ForestGreen} + & \color{ForestGreen} + & \color{ForestGreen} + & \color{BrickRed} - & \color{BrickRed} - & \color{BrickRed} -  &  &  &  &  \\ \hline
\multirow{2}{*}{Codeforces Round 1021 (Div. 1)} & \bf A  & \bf B  & \bf C  & \bf D  & \bf E  & \bf F  &  &  &  &  &  &  \multirow{2}{*}{0} & \multirow{2}{*}{-} & \multirow{2}{*}{651/651} & \multirow{2}{*}{982} \\
& \color{BrickRed} 0.0 & \color{BrickRed} 0.0 & \color{BrickRed} 0.0 & \color{BrickRed} 0.0 & \color{BrickRed} 0.0 & \color{BrickRed} 0.0  &  &  &  &  &  &  \\ \hline
\multirow{2}{*}{Codeforces Round 1021 (Div. 2)} & \bf A  & \bf B  & \bf C  & \bf D  & \bf E  & \bf F  &  &  &  &  &  &  \multirow{2}{*}{1735.71} & \multirow{2}{*}{-} & \multirow{2}{*}{1799/5824} & \multirow{2}{*}{1431} \\
&  \color{ForestGreen} 500.00 &  \color{ForestGreen} 1235.71 & \color{BrickRed} 0.0 & \color{BrickRed} 0.0 & \color{BrickRed} 0.0 & \color{BrickRed} 0.0  &  &  &  &  &  &  \\ \hline
\multirow{2}{*}{Educational Codeforces Round 178 (Rated for Div. 2)} & \bf A  & \bf B  & \bf C  & \bf D  & \bf E  & \bf F  & \bf G  &  &  &  &  &  \multirow{2}{*}{5} & \multirow{2}{*}{50.36} & \multirow{2}{*}{76/11706} & \multirow{2}{*}{2215} \\
& \color{ForestGreen} + & \color{ForestGreen} + & \color{ForestGreen} + & \color{ForestGreen} + & \color{ForestGreen} + & \color{BrickRed} - & \color{BrickRed} -  &  &  &  &  &  \\ \hline
\multirow{2}{*}{Codeforces Round 1022 (Div. 2)} & \bf A  & \bf B  & \bf C  & \bf D  & \bf E  & \bf F  &  &  &  &  &  &  \multirow{2}{*}{1825.00} & \multirow{2}{*}{-} & \multirow{2}{*}{3110/11127} & \multirow{2}{*}{1454} \\
&  \color{ForestGreen} 500.00 & \color{BrickRed} 0.0 &  \color{ForestGreen} 1325.00 & \color{BrickRed} 0.0 & \color{BrickRed} 0.0 & \color{BrickRed} 0.0  &  &  &  &  &  &  \\ \hline
\multirow{2}{*}{Codeforces Round 1023 (Div. 2)} & \bf A  & \bf B  & \bf C  & \bf D  & \bf E  & \bf F1  & \bf F2  &  &  &  &  &  \multirow{2}{*}{993.75} & \multirow{2}{*}{-} & \multirow{2}{*}{2848/11636} & \multirow{2}{*}{1485} \\
&  \color{ForestGreen} 250.00 &  \color{ForestGreen} 743.75 & \color{BrickRed} 0.0 & \color{BrickRed} 0.0 & \color{BrickRed} 0.0 & \color{BrickRed} 0.0 & \color{BrickRed} 0.0  &  &  &  &  &  \\ \hline
\multirow{2}{*}{Codeforces Round 1024 (Div. 1)} & \bf A  & \bf B  & \bf C  & \bf D  & \bf E  & \bf F  &  &  &  &  &  &  \multirow{2}{*}{0} & \multirow{2}{*}{-} & \multirow{2}{*}{857/857} & \multirow{2}{*}{938} \\
& \color{BrickRed} 0.0 & \color{BrickRed} 0.0 & \color{BrickRed} 0.0 & \color{BrickRed} 0.0 & \color{BrickRed} 0.0 & \color{BrickRed} 0.0  &  &  &  &  &  &  \\ \hline
\multirow{2}{*}{Codeforces Round 1024 (Div. 2)} & \bf A  & \bf B  & \bf C  & \bf D  & \bf E  & \bf F  &  &  &  &  &  &  \multirow{2}{*}{750.00} & \multirow{2}{*}{-} & \multirow{2}{*}{4640/11201} & \multirow{2}{*}{1246} \\
&  \color{ForestGreen} 250.00 &  \color{ForestGreen} 500.00 & \color{BrickRed} 0.0 & \color{BrickRed} 0.0 & \color{BrickRed} 0.0 & \color{BrickRed} 0.0  &  &  &  &  &  &  \\ \hline
\multirow{2}{*}{Codeforces Round 1025 (Div. 2)} & \bf A  & \bf B  & \bf C1  & \bf C2  & \bf C3  & \bf D  & \bf E  & \bf F  &  &  &  &  \multirow{2}{*}{3945.71} & \multirow{2}{*}{-} & \multirow{2}{*}{270/15945} & \multirow{2}{*}{2156} \\
&  \color{ForestGreen} 500.00 &  \color{ForestGreen} 985.71 & \color{BrickRed} 0.0 & \color{BrickRed} 0.0 & \color{BrickRed} 0.0 & \color{BrickRed} 0.0 &  \color{ForestGreen} 2460.00 & \color{BrickRed} 0.0  &  &  &  &  \\ \hline
\multirow{2}{*}{Codeforces Round 1026 (Div. 2)} & \bf A  & \bf B  & \bf C  & \bf D  & \bf E  & \bf F  &  &  &  &  &  &  \multirow{2}{*}{7581.25} & \multirow{2}{*}{-} & \multirow{2}{*}{15/17668} & \multirow{2}{*}{2198} \\
&  \color{ForestGreen} 500.00 &  \color{ForestGreen} 743.75 &  \color{ForestGreen} 1437.50 & \color{BrickRed} 0.0 &  \color{ForestGreen} 2075.00 &  \color{ForestGreen} 2825.00  &  &  &  &  &  &  \\ \hline
\multirow{2}{*}{Codeforces Round 1027 (Div. 3)} & \bf A  & \bf B  & \bf C  & \bf D  & \bf E  & \bf F  & \bf G  &  &  &  &  &  \multirow{2}{*}{6} & \multirow{2}{*}{9.11} & \multirow{2}{*}{12/22295} & \multirow{2}{*}{1709} \\
& \color{ForestGreen} + & \color{ForestGreen} + & \color{ForestGreen} + & \color{ForestGreen} + & \color{ForestGreen} + & \color{ForestGreen} + & \color{BrickRed} -  &  &  &  &  &  \\ \hline
\multirow{2}{*}{Codeforces Round 1028 (Div. 1)} & \bf A  & \bf B  & \bf C  & \bf D  & \bf E  & \bf F1  & \bf F2  &  &  &  &  &  \multirow{2}{*}{400.00} & \multirow{2}{*}{-} & \multirow{2}{*}{803/956} & \multirow{2}{*}{1840} \\
&  \color{ForestGreen} 400.00 & \color{BrickRed} 0.0 & \color{BrickRed} 0.0 & \color{BrickRed} 0.0 & \color{BrickRed} 0.0 & \color{BrickRed} 0.0 & \color{BrickRed} 0.0  &  &  &  &  &  \\ \hline
\multirow{2}{*}{Codeforces Round 1028 (Div. 2)} & \bf A  & \bf B  & \bf C  & \bf D  & \bf E  & \bf F  &  &  &  &  &  &  \multirow{2}{*}{2300.00} & \multirow{2}{*}{-} & \multirow{2}{*}{339/18314} & \multirow{2}{*}{2018} \\
&  \color{ForestGreen} 400.00 &  \color{ForestGreen} 750.00 &  \color{ForestGreen} 1150.00 & \color{BrickRed} 0.0 & \color{BrickRed} 0.0 & \color{BrickRed} 0.0  &  &  &  &  &  &  \\ \hline
\multirow{2}{*}{Educational Codeforces Round 179 (Rated for Div. 2)} & \bf A  & \bf B  & \bf C  & \bf D  & \bf E  & \bf F  & \bf G  &  &  &  &  &  \multirow{2}{*}{3} & \multirow{2}{*}{37.86} & \multirow{2}{*}{4100/12301} & \multirow{2}{*}{1371} \\
& \color{ForestGreen} + & \color{ForestGreen} + & \color{ForestGreen} + & \color{BrickRed} - & \color{BrickRed} - & \color{BrickRed} - & \color{BrickRed} -  &  &  &  &  &  \\ \hline
\multirow{2}{*}{Codeforces Round 1029 (Div. 3)} & \bf A  & \bf B  & \bf C  & \bf D  & \bf E  & \bf F  & \bf G  & \bf H  &  &  &  &  \multirow{2}{*}{5} & \multirow{2}{*}{15.36} & \multirow{2}{*}{460/20324} & \multirow{2}{*}{1707} \\
& \color{ForestGreen} + & \color{ForestGreen} + & \color{ForestGreen} + & \color{ForestGreen} + & \color{BrickRed} - & \color{BrickRed} - & \color{ForestGreen} + & \color{BrickRed} -  &  &  &  &  \\ \hline
\multirow{2}{*}{Codeforces Round 1030 (Div. 2)} & \bf A  & \bf B  & \bf C  & \bf D1  & \bf D2  & \bf E  & \bf F  &  &  &  &  &  \multirow{2}{*}{2568.75} & \multirow{2}{*}{-} & \multirow{2}{*}{1960/18335} & \multirow{2}{*}{1715} \\
&  \color{ForestGreen} 500.00 & \color{BrickRed} 0.0 &  \color{ForestGreen} 825.00 &  \color{ForestGreen} 1243.75 & \color{BrickRed} 0.0 & \color{BrickRed} 0.0 & \color{BrickRed} 0.0  &  &  &  &  &  \\ \hline
\multirow{2}{*}{Codeforces Round 1031 (Div. 2)} & \bf A  & \bf B  & \bf C  & \bf D  & \bf E  & \bf F  &  &  &  &  &  &  \multirow{2}{*}{493.75} & \multirow{2}{*}{-} & \multirow{2}{*}{5469/11032} & \multirow{2}{*}{1134} \\
&  \color{ForestGreen} 493.75 & \color{BrickRed} 0.0 & \color{BrickRed} 0.0 & \color{BrickRed} 0.0 & \color{BrickRed} 0.0 & \color{BrickRed} 0.0  &  &  &  &  &  &  \\ \hline
\multirow{2}{*}{Codeforces Round 1032 (Div. 3)} & \bf A  & \bf B  & \bf C  & \bf D  & \bf E  & \bf F  & \bf G  & \bf H  &  &  &  &  \multirow{2}{*}{7} & \multirow{2}{*}{20.00} & \multirow{2}{*}{17/22170} & \multirow{2}{*}{1733} \\
& \color{ForestGreen} + & \color{ForestGreen} + & \color{ForestGreen} + & \color{ForestGreen} + & \color{ForestGreen} + & \color{ForestGreen} + & \color{ForestGreen} + & \color{BrickRed} -  &  &  &  &  \\ \hline
\multirow{2}{*}{Codeforces Round 1033 (Div. 2) and CodeNite 2025} & \bf A  & \bf B  & \bf C  & \bf D  & \bf E  & \bf F  & \bf G  &  &  &  &  &  \multirow{2}{*}{4762.50} & \multirow{2}{*}{-} & \multirow{2}{*}{183/12948} & \multirow{2}{*}{2216} \\
&  \color{ForestGreen} 500.00 &  \color{ForestGreen} 750.00 &  \color{ForestGreen} 1187.50 & \color{BrickRed} 0.0 &  \color{ForestGreen} 2325.00 & \color{BrickRed} 0.0 & \color{BrickRed} 0.0  &  &  &  &  &  \\ \hline
\multirow{2}{*}{Educational Codeforces Round 180 (Rated for Div. 2)} & \bf A  & \bf B  & \bf C  & \bf D  & \bf E  & \bf F  &  &  &  &  &  &  \multirow{2}{*}{4} & \multirow{2}{*}{37.50} & \multirow{2}{*}{345/17128} & \multirow{2}{*}{2114} \\
& \color{ForestGreen} + & \color{ForestGreen} + & \color{ForestGreen} + & \color{BrickRed} - & \color{ForestGreen} + & \color{BrickRed} -  &  &  &  &  &  &  \\ \hline
\multirow{2}{*}{Codeforces Round 1035 (Div. 2)} & \bf A  & \bf B  & \bf C  & \bf D  & \bf E  & \bf F  &  &  &  &  &  &  \multirow{2}{*}{2985.71} & \multirow{2}{*}{-} & \multirow{2}{*}{587/15624} & \multirow{2}{*}{2008} \\
&  \color{ForestGreen} 500.00 &  \color{ForestGreen} 1000.00 &  \color{ForestGreen} 1485.71 & \color{BrickRed} 0.0 & \color{BrickRed} 0.0 & \color{BrickRed} 0.0  &  &  &  &  &  &  \\ \bottomrule
\end{tabular}}
\label{tab:elo_rating_8b}
\end{table}

%\newcolumntype{P}[1]{>{\centering\arraybackslash}m{#1}}

% \usepackage{multirow}
\begin{table}[!htbp]
\centering
\caption{Nemotron-Cascade-14B-Thinking performance details on 51 Codeforces Rounds ranging from 2501-2507. We attempt each problem with $N=8$ times in total. For regular codeforces rounds, we present the score after considering expected penalties for each problem. For ICPC style rounds, we mark passed/failed problems as \textcolor{ForestGreen}{+} and \textcolor{BrickRed}{-} correspondingly. We compute the estimated rank to human contestants and the corresponding Elo score as shown in rightmost two columns.}
\scalebox{0.47}{
\begin{tabular}{c| *{11}{P{1.1cm}} |c c c c}
\toprule
\bfseries Contest Name & \multicolumn{11}{c|}{\bfseries Contest Problems} &
\bfseries Score & \bfseries Penalty & \bfseries Est. Rank & \bfseries ELO \\ \midrule
\multirow{2}{*}{Hello 2025} & \bf A  & \bf B  & \bf C  & \bf D  & \bf E1  & \bf E2  & \bf F  & \bf G  & \bf H  &  &  &  \multirow{2}{*}{4975.00} & \multirow{2}{*}{-} & \multirow{2}{*}{679/16703} & \multirow{2}{*}{2290} \\
&  \color{ForestGreen} 500.00 &  \color{ForestGreen} 1000.00 &  \color{ForestGreen} 1400.00 &  \color{ForestGreen} 2075.00 & \color{BrickRed} 0.0 & \color{BrickRed} 0.0 & \color{BrickRed} 0.0 & \color{BrickRed} 0.0 & \color{BrickRed} 0.0  &  &  &  \\ \hline
\multirow{2}{*}{Codeforces Round 996 (Div. 2)} & \bf A  & \bf B  & \bf C  & \bf D  & \bf E  & \bf F  &  &  &  &  &  &  \multirow{2}{*}{2887.50} & \multirow{2}{*}{-} & \multirow{2}{*}{755/21232} & \multirow{2}{*}{1977} \\
&  \color{ForestGreen} 475.00 &  \color{ForestGreen} 937.50 &  \color{ForestGreen} 1475.00 & \color{BrickRed} 0.0 & \color{BrickRed} 0.0 & \color{BrickRed} 0.0  &  &  &  &  &  &  \\ \hline
\multirow{2}{*}{Codeforces Round 997 (Div. 2)} & \bf A  & \bf B  & \bf C  & \bf D  & \bf E  & \bf F1  & \bf F2  &  &  &  &  &  \multirow{2}{*}{5110.71} & \multirow{2}{*}{-} & \multirow{2}{*}{41/18823} & \multirow{2}{*}{2198} \\
&  \color{ForestGreen} 437.50 &  \color{ForestGreen} 1250.00 &  \color{ForestGreen} 1485.71 &  \color{ForestGreen} 1937.50 & \color{BrickRed} 0.0 & \color{BrickRed} 0.0 & \color{BrickRed} 0.0  &  &  &  &  &  \\ \hline
\multirow{2}{*}{Codeforces Round 998 (Div. 3)} & \bf A  & \bf B  & \bf C  & \bf D  & \bf E  & \bf F  & \bf G  &  &  &  &  &  \multirow{2}{*}{6} & \multirow{2}{*}{30.86} & \multirow{2}{*}{22/24247} & \multirow{2}{*}{1699} \\
& \color{ForestGreen} + & \color{ForestGreen} + & \color{ForestGreen} + & \color{ForestGreen} + & \color{ForestGreen} + & \color{ForestGreen} + & \color{BrickRed} -  &  &  &  &  &  \\ \hline
\multirow{2}{*}{IAEPC Preliminary Contest (Codeforces Round 999, Div. 1 + Div. 2)} & \bf A  & \bf B  & \bf C  & \bf D  & \bf E  & \bf F1  & \bf F2  & \bf G  & \bf H1  & \bf H2  & \bf I  &  \multirow{2}{*}{4397.50} & \multirow{2}{*}{-} & \multirow{2}{*}{1170/12647} & \multirow{2}{*}{1998} \\
&  \color{ForestGreen} 500.00 &  \color{ForestGreen} 937.50 &  \color{ForestGreen} 1500.00 &  \color{ForestGreen} 1460.00 & \color{BrickRed} 0.0 & \color{BrickRed} 0.0 & \color{BrickRed} 0.0 & \color{BrickRed} 0.0 & \color{BrickRed} 0.0 & \color{BrickRed} 0.0 & \color{BrickRed} 0.0  &  \\ \hline
\multirow{2}{*}{Codeforces Round 1000 (Div. 2)} & \bf A  & \bf B  & \bf C  & \bf D  & \bf E  & \bf F1  & \bf F2  &  &  &  &  &  \multirow{2}{*}{6825.00} & \multirow{2}{*}{-} & \multirow{2}{*}{39/17169} & \multirow{2}{*}{2200} \\
&  \color{ForestGreen} 500.00 & \color{BrickRed} 0.0 &  \color{ForestGreen} 1500.00 &  \color{ForestGreen} 2250.00 &  \color{ForestGreen} 2575.00 & \color{BrickRed} 0.0 & \color{BrickRed} 0.0  &  &  &  &  &  \\ \hline
\multirow{2}{*}{Ethflow Round 1 (Codeforces Round 1001, Div. 1 + Div. 2)} & \bf A  & \bf B  & \bf C  & \bf D  & \bf E1  & \bf E2  & \bf F  & \bf G  & \bf H  &  &  &  \multirow{2}{*}{2445.71} & \multirow{2}{*}{-} & \multirow{2}{*}{1737/16234} & \multirow{2}{*}{1895} \\
&  \color{ForestGreen} 500.00 &  \color{ForestGreen} 985.71 &  \color{ForestGreen} 960.00 & \color{BrickRed} 0.0 & \color{BrickRed} 0.0 & \color{BrickRed} 0.0 & \color{BrickRed} 0.0 & \color{BrickRed} 0.0 & \color{BrickRed} 0.0  &  &  &  \\ \hline
\multirow{2}{*}{Codeforces Round 1002 (Div. 2)} & \bf A  & \bf B  & \bf C  & \bf D  & \bf E1  & \bf E2  &  &  &  &  &  &  \multirow{2}{*}{1325.00} & \multirow{2}{*}{-} & \multirow{2}{*}{4240/19443} & \multirow{2}{*}{1454} \\
&  \color{ForestGreen} 500.00 &  \color{ForestGreen} 825.00 & \color{BrickRed} 0.0 & \color{BrickRed} 0.0 & \color{BrickRed} 0.0 & \color{BrickRed} 0.0  &  &  &  &  &  &  \\ \hline
\multirow{2}{*}{Codeforces Round 1003 (Div. 4)} & \bf A  & \bf B  & \bf C1  & \bf C2  & \bf D  & \bf E  & \bf F  & \bf G  & \bf H  &  &  &  \multirow{2}{*}{9} & \multirow{2}{*}{6.61} & \multirow{2}{*}{1/28033} & \multirow{2}{*}{1522} \\
& \color{ForestGreen} + & \color{ForestGreen} + & \color{ForestGreen} + & \color{ForestGreen} + & \color{ForestGreen} + & \color{ForestGreen} + & \color{ForestGreen} + & \color{ForestGreen} + & \color{ForestGreen} +  &  &  &  \\ \hline
\multirow{2}{*}{Codeforces Round 1004 (Div. 1)} & \bf A  & \bf B  & \bf C  & \bf D1  & \bf D2  & \bf E  & \bf F  &  &  &  &  &  \multirow{2}{*}{1950.00} & \multirow{2}{*}{-} & \multirow{2}{*}{373/1030} & \multirow{2}{*}{2339} \\
& \color{BrickRed} 0.0 & \color{BrickRed} 0.0 &  \color{ForestGreen} 1225.00 &  \color{ForestGreen} 725.00 & \color{BrickRed} 0.0 & \color{BrickRed} 0.0 & \color{BrickRed} 0.0  &  &  &  &  &  \\ \hline
\multirow{2}{*}{Codeforces Round 1004 (Div. 2)} & \bf A  & \bf B  & \bf C  & \bf D  & \bf E  & \bf F  & \bf G  &  &  &  &  &  \multirow{2}{*}{3625.00} & \multirow{2}{*}{-} & \multirow{2}{*}{282/16749} & \multirow{2}{*}{2087} \\
&  \color{ForestGreen} 500.00 &  \color{ForestGreen} 900.00 & \color{BrickRed} 0.0 & \color{BrickRed} 0.0 & \color{BrickRed} 0.0 &  \color{ForestGreen} 2225.00 & \color{BrickRed} 0.0  &  &  &  &  &  \\ \hline
\multirow{2}{*}{Codeforces Round 1005 (Div. 2)} & \bf A  & \bf B  & \bf C  & \bf D  & \bf E  & \bf F  &  &  &  &  &  &  \multirow{2}{*}{2543.75} & \multirow{2}{*}{-} & \multirow{2}{*}{1433/17621} & \multirow{2}{*}{1806} \\
&  \color{ForestGreen} 493.75 &  \color{ForestGreen} 975.00 &  \color{ForestGreen} 1075.00 & \color{BrickRed} 0.0 & \color{BrickRed} 0.0 & \color{BrickRed} 0.0  &  &  &  &  &  &  \\ \hline
\multirow{2}{*}{Educational Codeforces Round 174 (Rated for Div. 2)} & \bf A  & \bf B  & \bf C  & \bf D  & \bf E  & \bf F  &  &  &  &  &  &  \multirow{2}{*}{3} & \multirow{2}{*}{0.00} & \multirow{2}{*}{1298/16701} & \multirow{2}{*}{1810} \\
& \color{ForestGreen} + & \color{ForestGreen} + & \color{ForestGreen} + & \color{BrickRed} - & \color{BrickRed} - & \color{BrickRed} -  &  &  &  &  &  &  \\ \hline
\multirow{2}{*}{Codeforces Round 1006 (Div. 3)} & \bf A  & \bf B  & \bf C  & \bf D  & \bf E  & \bf F  & \bf G  &  &  &  &  &  \multirow{2}{*}{7} & \multirow{2}{*}{12.11} & \multirow{2}{*}{1/24140} & \multirow{2}{*}{1699} \\
& \color{ForestGreen} + & \color{ForestGreen} + & \color{ForestGreen} + & \color{ForestGreen} + & \color{ForestGreen} + & \color{ForestGreen} + & \color{ForestGreen} +  &  &  &  &  &  \\ \hline
\multirow{2}{*}{Educational Codeforces Round 175 (Rated for Div. 2)} & \bf A  & \bf B  & \bf C  & \bf D  & \bf E  & \bf F  &  &  &  &  &  &  \multirow{2}{*}{4} & \multirow{2}{*}{0.00} & \multirow{2}{*}{234/16060} & \multirow{2}{*}{2195} \\
& \color{ForestGreen} + & \color{ForestGreen} + & \color{ForestGreen} + & \color{ForestGreen} + & \color{BrickRed} - & \color{BrickRed} -  &  &  &  &  &  &  \\ \hline
\multirow{2}{*}{Codeforces Round 1007 (Div. 2)} & \bf A  & \bf B  & \bf C  & \bf D1  & \bf D2  & \bf E  & \bf F  &  &  &  &  &  \multirow{2}{*}{6910.71} & \multirow{2}{*}{-} & \multirow{2}{*}{2/16254} & \multirow{2}{*}{2198} \\
&  \color{ForestGreen} 485.71 &  \color{ForestGreen} 975.00 &  \color{ForestGreen} 1325.00 &  \color{ForestGreen} 1725.00 & \color{BrickRed} 0.0 &  \color{ForestGreen} 2400.00 & \color{BrickRed} 0.0  &  &  &  &  &  \\ \hline
\multirow{2}{*}{Codeforces Round 1008 (Div. 1)} & \bf A  & \bf B  & \bf C  & \bf D  & \bf E  & \bf F  & \bf G  &  &  &  &  &  \multirow{2}{*}{1985.71} & \multirow{2}{*}{-} & \multirow{2}{*}{359/909} & \multirow{2}{*}{2307} \\
&  \color{ForestGreen} 485.71 & \color{BrickRed} 0.0 &  \color{ForestGreen} 1500.00 & \color{BrickRed} 0.0 & \color{BrickRed} 0.0 & \color{BrickRed} 0.0 & \color{BrickRed} 0.0  &  &  &  &  &  \\ \hline
\multirow{2}{*}{Codeforces Round 1008 (Div. 2)} & \bf A  & \bf B  & \bf C  & \bf D  & \bf E  & \bf F  & \bf G  &  &  &  &  &  \multirow{2}{*}{6885.71} & \multirow{2}{*}{-} & \multirow{2}{*}{8/14641} & \multirow{2}{*}{2008} \\
&  \color{ForestGreen} 500.00 &  \color{ForestGreen} 750.00 &  \color{ForestGreen} 1235.71 &  \color{ForestGreen} 1650.00 & \color{BrickRed} 0.0 &  \color{ForestGreen} 2750.00 & \color{BrickRed} 0.0  &  &  &  &  &  \\ \hline
\multirow{2}{*}{Codeforces Round 1009 (Div. 3)} & \bf A  & \bf B  & \bf C  & \bf D  & \bf E  & \bf F  & \bf G  &  &  &  &  &  \multirow{2}{*}{6} & \multirow{2}{*}{44.25} & \multirow{2}{*}{15/23635} & \multirow{2}{*}{1708} \\
& \color{ForestGreen} + & \color{ForestGreen} + & \color{ForestGreen} + & \color{ForestGreen} + & \color{BrickRed} - & \color{ForestGreen} + & \color{ForestGreen} +  &  &  &  &  &  \\ \hline
\multirow{2}{*}{Educational Codeforces Round 176 (Rated for Div. 2)} & \bf A  & \bf B  & \bf C  & \bf D  & \bf E  & \bf F  &  &  &  &  &  &  \multirow{2}{*}{4} & \multirow{2}{*}{29.25} & \multirow{2}{*}{73/18159} & \multirow{2}{*}{2198} \\
& \color{ForestGreen} + & \color{BrickRed} - & \color{ForestGreen} + & \color{ForestGreen} + & \color{ForestGreen} + & \color{BrickRed} -  &  &  &  &  &  &  \\ \hline
\multirow{2}{*}{Codeforces Round 1011 (Div. 2)} & \bf A  & \bf B  & \bf C  & \bf D  & \bf E  & \bf F1  & \bf F2  &  &  &  &  &  \multirow{2}{*}{8670.71} & \multirow{2}{*}{-} & \multirow{2}{*}{1/15906} & \multirow{2}{*}{2200} \\
&  \color{ForestGreen} 500.00 &  \color{ForestGreen} 1075.00 &  \color{ForestGreen} 1210.00 &  \color{ForestGreen} 1735.71 &  \color{ForestGreen} 2325.00 &  \color{ForestGreen} 1825.00 & \color{BrickRed} 0.0  &  &  &  &  &  \\ \hline
\multirow{2}{*}{Codeforces Round 1012 (Div. 1)} & \bf A  & \bf B1  & \bf B2  & \bf C1  & \bf C2  & \bf D  & \bf E  &  &  &  &  &  \multirow{2}{*}{1650.00} & \multirow{2}{*}{-} & \multirow{2}{*}{365/653} & \multirow{2}{*}{2135} \\
& \color{BrickRed} 0.0 &  \color{ForestGreen} 825.00 & \color{BrickRed} 0.0 &  \color{ForestGreen} 825.00 & \color{BrickRed} 0.0 & \color{BrickRed} 0.0 & \color{BrickRed} 0.0  &  &  &  &  &  \\ \hline
\multirow{2}{*}{Codeforces Round 1012 (Div. 2)} & \bf A  & \bf B  & \bf C  & \bf D  & \bf E1  & \bf E2  & \bf F1  & \bf F2  &  &  &  &  \multirow{2}{*}{6737.50} & \multirow{2}{*}{-} & \multirow{2}{*}{12/8536} & \multirow{2}{*}{2007} \\
&  \color{ForestGreen} 500.00 &  \color{ForestGreen} 937.50 &  \color{ForestGreen} 1650.00 & \color{BrickRed} 0.0 &  \color{ForestGreen} 1825.00 & \color{BrickRed} 0.0 &  \color{ForestGreen} 1825.00 & \color{BrickRed} 0.0  &  &  &  &  \\ \hline
\multirow{2}{*}{Codeforces Round 1013 (Div. 3)} & \bf A  & \bf B  & \bf C  & \bf D  & \bf E  & \bf F  & \bf G  &  &  &  &  &  \multirow{2}{*}{6} & \multirow{2}{*}{6.96} & \multirow{2}{*}{22/24379} & \multirow{2}{*}{1715} \\
& \color{ForestGreen} + & \color{ForestGreen} + & \color{ForestGreen} + & \color{ForestGreen} + & \color{ForestGreen} + & \color{ForestGreen} + & \color{BrickRed} -  &  &  &  &  &  \\ \hline
\multirow{2}{*}{Codeforces Round 1014 (Div. 2)} & \bf A  & \bf B  & \bf C  & \bf D  & \bf E  & \bf F  &  &  &  &  &  &  \multirow{2}{*}{6441.25} & \multirow{2}{*}{-} & \multirow{2}{*}{2/15842} & \multirow{2}{*}{2213} \\
&  \color{ForestGreen} 500.00 &  \color{ForestGreen} 743.75 &  \color{ForestGreen} 1243.75 &  \color{ForestGreen} 1710.00 &  \color{ForestGreen} 2243.75 & \color{BrickRed} 0.0  &  &  &  &  &  &  \\ \hline
\multirow{2}{*}{Teza Round 1 (Codeforces Round 1015, Div. 1 + Div. 2)} & \bf A  & \bf B  & \bf C  & \bf D  & \bf E  & \bf F  & \bf G1  & \bf G2  & \bf H  &  &  &  \multirow{2}{*}{7037.50} & \multirow{2}{*}{-} & \multirow{2}{*}{186/11206} & \multirow{2}{*}{2631} \\
&  \color{ForestGreen} 743.75 &  \color{ForestGreen} 1000.00 &  \color{ForestGreen} 1493.75 &  \color{ForestGreen} 1725.00 &  \color{ForestGreen} 2075.00 & \color{BrickRed} 0.0 & \color{BrickRed} 0.0 & \color{BrickRed} 0.0 & \color{BrickRed} 0.0  &  &  &  \\ \hline
\multirow{2}{*}{Codeforces Round 1016 (Div. 3)} & \bf A  & \bf B  & \bf C  & \bf D  & \bf E  & \bf F  & \bf G  &  &  &  &  &  \multirow{2}{*}{7} & \multirow{2}{*}{52.50} & \multirow{2}{*}{1/21249} & \multirow{2}{*}{1699} \\
& \color{ForestGreen} + & \color{ForestGreen} + & \color{ForestGreen} + & \color{ForestGreen} + & \color{ForestGreen} + & \color{ForestGreen} + & \color{ForestGreen} +  &  &  &  &  &  \\ \hline
\multirow{2}{*}{Codeforces Round 1017 (Div. 4)} & \bf A  & \bf B  & \bf C  & \bf D  & \bf E  & \bf F  & \bf G  & \bf H  &  &  &  &  \multirow{2}{*}{8} & \multirow{2}{*}{24.00} & \multirow{2}{*}{1/22234} & \multirow{2}{*}{1503} \\
& \color{ForestGreen} + & \color{ForestGreen} + & \color{ForestGreen} + & \color{ForestGreen} + & \color{ForestGreen} + & \color{ForestGreen} + & \color{ForestGreen} + & \color{ForestGreen} +  &  &  &  &  \\ \hline
\multirow{2}{*}{Neowise Labs Contest 1 (Codeforces Round 1018, Div. 1 + Div. 2)} & \bf A  & \bf B  & \bf C  & \bf D  & \bf E  & \bf F  & \bf G  & \bf H  &  &  &  &  \multirow{2}{*}{4398.21} & \multirow{2}{*}{-} & \multirow{2}{*}{493/12771} & \multirow{2}{*}{2312} \\
&  \color{ForestGreen} 485.71 &  \color{ForestGreen} 750.00 &  \color{ForestGreen} 1475.00 &  \color{ForestGreen} 1687.50 & \color{BrickRed} 0.0 & \color{BrickRed} 0.0 & \color{BrickRed} 0.0 & \color{BrickRed} 0.0  &  &  &  &  \\ \hline
\multirow{2}{*}{Codeforces Round 1019 (Div. 2)} & \bf A  & \bf B  & \bf C  & \bf D  & \bf E  & \bf F  &  &  &  &  &  &  \multirow{2}{*}{2931.25} & \multirow{2}{*}{-} & \multirow{2}{*}{852/14465} & \multirow{2}{*}{1902} \\
&  \color{ForestGreen} 500.00 &  \color{ForestGreen} 993.75 &  \color{ForestGreen} 1437.50 & \color{BrickRed} 0.0 & \color{BrickRed} 0.0 & \color{BrickRed} 0.0  &  &  &  &  &  &  \\ \hline
\multirow{2}{*}{Codeforces Round 1020 (Div. 3)} & \bf A  & \bf B  & \bf C  & \bf D  & \bf E  & \bf F  & \bf G1  & \bf G2  &  &  &  &  \multirow{2}{*}{6} & \multirow{2}{*}{25.00} & \multirow{2}{*}{54/17451} & \multirow{2}{*}{1708} \\
& \color{ForestGreen} + & \color{ForestGreen} + & \color{ForestGreen} + & \color{ForestGreen} + & \color{ForestGreen} + & \color{ForestGreen} + & \color{BrickRed} - & \color{BrickRed} -  &  &  &  &  \\ \hline
\multirow{2}{*}{Codeforces Round 1021 (Div. 2)} & \bf A  & \bf B  & \bf C  & \bf D  & \bf E  & \bf F  &  &  &  &  &  &  \multirow{2}{*}{3068.75} & \multirow{2}{*}{-} & \multirow{2}{*}{162/5824} & \multirow{2}{*}{2019} \\
&  \color{ForestGreen} 500.00 &  \color{ForestGreen} 1243.75 &  \color{ForestGreen} 1325.00 & \color{BrickRed} 0.0 & \color{BrickRed} 0.0 & \color{BrickRed} 0.0  &  &  &  &  &  &  \\ \hline
\multirow{2}{*}{Codeforces Round 1021 (Div. 1)} & \bf A  & \bf B  & \bf C  & \bf D  & \bf E  & \bf F  &  &  &  &  &  &  \multirow{2}{*}{325.00} & \multirow{2}{*}{-} & \multirow{2}{*}{625/651} & \multirow{2}{*}{1568} \\
&  \color{ForestGreen} 325.00 & \color{BrickRed} 0.0 & \color{BrickRed} 0.0 & \color{BrickRed} 0.0 & \color{BrickRed} 0.0 & \color{BrickRed} 0.0  &  &  &  &  &  &  \\ \hline
\multirow{2}{*}{Educational Codeforces Round 178 (Rated for Div. 2)} & \bf A  & \bf B  & \bf C  & \bf D  & \bf E  & \bf F  & \bf G  &  &  &  &  &  \multirow{2}{*}{4} & \multirow{2}{*}{2.86} & \multirow{2}{*}{1810/11706} & \multirow{2}{*}{1661} \\
& \color{ForestGreen} + & \color{ForestGreen} + & \color{BrickRed} - & \color{ForestGreen} + & \color{ForestGreen} + & \color{BrickRed} - & \color{BrickRed} -  &  &  &  &  &  \\ \hline
\multirow{2}{*}{Codeforces Round 1022 (Div. 2)} & \bf A  & \bf B  & \bf C  & \bf D  & \bf E  & \bf F  &  &  &  &  &  &  \multirow{2}{*}{1900.00} & \multirow{2}{*}{-} & \multirow{2}{*}{3073/11127} & \multirow{2}{*}{1459} \\
&  \color{ForestGreen} 500.00 & \color{BrickRed} 0.0 &  \color{ForestGreen} 1400.00 & \color{BrickRed} 0.0 & \color{BrickRed} 0.0 & \color{BrickRed} 0.0  &  &  &  &  &  &  \\ \hline
\multirow{2}{*}{Codeforces Round 1023 (Div. 2)} & \bf A  & \bf B  & \bf C  & \bf D  & \bf E  & \bf F1  & \bf F2  &  &  &  &  &  \multirow{2}{*}{4337.50} & \multirow{2}{*}{-} & \multirow{2}{*}{79/11636} & \multirow{2}{*}{2209} \\
&  \color{ForestGreen} 250.00 &  \color{ForestGreen} 750.00 &  \color{ForestGreen} 1437.50 &  \color{ForestGreen} 1900.00 & \color{BrickRed} 0.0 & \color{BrickRed} 0.0 & \color{BrickRed} 0.0  &  &  &  &  &  \\ \hline
\multirow{2}{*}{Codeforces Round 1024 (Div. 1)} & \bf A  & \bf B  & \bf C  & \bf D  & \bf E  & \bf F  &  &  &  &  &  &  \multirow{2}{*}{0} & \multirow{2}{*}{-} & \multirow{2}{*}{857/857} & \multirow{2}{*}{938} \\
& \color{BrickRed} 0.0 & \color{BrickRed} 0.0 & \color{BrickRed} 0.0 & \color{BrickRed} 0.0 & \color{BrickRed} 0.0 & \color{BrickRed} 0.0  &  &  &  &  &  &  \\ \hline
\multirow{2}{*}{Codeforces Round 1024 (Div. 2)} & \bf A  & \bf B  & \bf C  & \bf D  & \bf E  & \bf F  &  &  &  &  &  &  \multirow{2}{*}{750.00} & \multirow{2}{*}{-} & \multirow{2}{*}{4640/11201} & \multirow{2}{*}{1246} \\
&  \color{ForestGreen} 250.00 &  \color{ForestGreen} 500.00 & \color{BrickRed} 0.0 & \color{BrickRed} 0.0 & \color{BrickRed} 0.0 & \color{BrickRed} 0.0  &  &  &  &  &  &  \\ \hline
\multirow{2}{*}{Codeforces Round 1025 (Div. 2)} & \bf A  & \bf B  & \bf C1  & \bf C2  & \bf C3  & \bf D  & \bf E  & \bf F  &  &  &  &  \multirow{2}{*}{3885.71} & \multirow{2}{*}{-} & \multirow{2}{*}{304/15945} & \multirow{2}{*}{2131} \\
&  \color{ForestGreen} 500.00 &  \color{ForestGreen} 985.71 & \color{BrickRed} 0.0 & \color{BrickRed} 0.0 & \color{BrickRed} 0.0 & \color{BrickRed} 0.0 &  \color{ForestGreen} 2400.00 & \color{BrickRed} 0.0  &  &  &  &  \\ \hline
\multirow{2}{*}{Codeforces Round 1026 (Div. 2)} & \bf A  & \bf B  & \bf C  & \bf D  & \bf E  & \bf F  &  &  &  &  &  &  \multirow{2}{*}{7725.00} & \multirow{2}{*}{-} & \multirow{2}{*}{10/17668} & \multirow{2}{*}{2198} \\
&  \color{ForestGreen} 500.00 &  \color{ForestGreen} 750.00 &  \color{ForestGreen} 1500.00 & \color{BrickRed} 0.0 &  \color{ForestGreen} 2150.00 &  \color{ForestGreen} 2825.00  &  &  &  &  &  &  \\ \hline
\multirow{2}{*}{Codeforces Round 1027 (Div. 3)} & \bf A  & \bf B  & \bf C  & \bf D  & \bf E  & \bf F  & \bf G  &  &  &  &  &  \multirow{2}{*}{6} & \multirow{2}{*}{28.00} & \multirow{2}{*}{12/22295} & \multirow{2}{*}{1709} \\
& \color{ForestGreen} + & \color{ForestGreen} + & \color{ForestGreen} + & \color{ForestGreen} + & \color{ForestGreen} + & \color{ForestGreen} + & \color{BrickRed} -  &  &  &  &  &  \\ \hline
\multirow{2}{*}{Codeforces Round 1028 (Div. 2)} & \bf A  & \bf B  & \bf C  & \bf D  & \bf E  & \bf F  &  &  &  &  &  &  \multirow{2}{*}{4610.00} & \multirow{2}{*}{-} & \multirow{2}{*}{5/18314} & \multirow{2}{*}{2018} \\
&  \color{ForestGreen} 325.00 &  \color{ForestGreen} 750.00 &  \color{ForestGreen} 1210.00 & \color{BrickRed} 0.0 &  \color{ForestGreen} 2325.00 & \color{BrickRed} 0.0  &  &  &  &  &  &  \\ \hline
\multirow{2}{*}{Codeforces Round 1028 (Div. 1)} & \bf A  & \bf B  & \bf C  & \bf D  & \bf E  & \bf F1  & \bf F2  &  &  &  &  &  \multirow{2}{*}{2035.00} & \multirow{2}{*}{-} & \multirow{2}{*}{135/956} & \multirow{2}{*}{2673} \\
&  \color{ForestGreen} 460.00 & \color{BrickRed} 0.0 &  \color{ForestGreen} 1575.00 & \color{BrickRed} 0.0 & \color{BrickRed} 0.0 & \color{BrickRed} 0.0 & \color{BrickRed} 0.0  &  &  &  &  &  \\ \hline
\multirow{2}{*}{Educational Codeforces Round 179 (Rated for Div. 2)} & \bf A  & \bf B  & \bf C  & \bf D  & \bf E  & \bf F  & \bf G  &  &  &  &  &  \multirow{2}{*}{4} & \multirow{2}{*}{76.25} & \multirow{2}{*}{998/12301} & \multirow{2}{*}{1848} \\
& \color{ForestGreen} + & \color{ForestGreen} + & \color{ForestGreen} + & \color{ForestGreen} + & \color{BrickRed} - & \color{BrickRed} - & \color{BrickRed} -  &  &  &  &  &  \\ \hline
\multirow{2}{*}{Codeforces Round 1029 (Div. 3)} & \bf A  & \bf B  & \bf C  & \bf D  & \bf E  & \bf F  & \bf G  & \bf H  &  &  &  &  \multirow{2}{*}{5} & \multirow{2}{*}{32.50} & \multirow{2}{*}{460/20324} & \multirow{2}{*}{1707} \\
& \color{ForestGreen} + & \color{ForestGreen} + & \color{ForestGreen} + & \color{ForestGreen} + & \color{BrickRed} - & \color{BrickRed} - & \color{ForestGreen} + & \color{BrickRed} -  &  &  &  &  \\ \hline
\multirow{2}{*}{Codeforces Round 1030 (Div. 2)} & \bf A  & \bf B  & \bf C  & \bf D1  & \bf D2  & \bf E  & \bf F  &  &  &  &  &  \multirow{2}{*}{5025.00} & \multirow{2}{*}{-} & \multirow{2}{*}{35/18335} & \multirow{2}{*}{2205} \\
&  \color{ForestGreen} 500.00 & \color{BrickRed} 0.0 &  \color{ForestGreen} 900.00 &  \color{ForestGreen} 1187.50 & \color{BrickRed} 0.0 &  \color{ForestGreen} 2437.50 & \color{BrickRed} 0.0  &  &  &  &  &  \\ \hline
\multirow{2}{*}{Codeforces Round 1031 (Div. 2)} & \bf A  & \bf B  & \bf C  & \bf D  & \bf E  & \bf F  &  &  &  &  &  &  \multirow{2}{*}{500.00} & \multirow{2}{*}{-} & \multirow{2}{*}{5433/11032} & \multirow{2}{*}{1138} \\
&  \color{ForestGreen} 500.00 & \color{BrickRed} 0.0 & \color{BrickRed} 0.0 & \color{BrickRed} 0.0 & \color{BrickRed} 0.0 & \color{BrickRed} 0.0  &  &  &  &  &  &  \\ \hline
\multirow{2}{*}{Codeforces Round 1032 (Div. 3)} & \bf A  & \bf B  & \bf C  & \bf D  & \bf E  & \bf F  & \bf G  & \bf H  &  &  &  &  \multirow{2}{*}{7} & \multirow{2}{*}{3.75} & \multirow{2}{*}{17/22170} & \multirow{2}{*}{1733} \\
& \color{ForestGreen} + & \color{ForestGreen} + & \color{ForestGreen} + & \color{ForestGreen} + & \color{ForestGreen} + & \color{ForestGreen} + & \color{ForestGreen} + & \color{BrickRed} -  &  &  &  &  \\ \hline
\multirow{2}{*}{Codeforces Round 1033 (Div. 2) and CodeNite 2025} & \bf A  & \bf B  & \bf C  & \bf D  & \bf E  & \bf F  & \bf G  &  &  &  &  &  \multirow{2}{*}{6481.25} & \multirow{2}{*}{-} & \multirow{2}{*}{31/12948} & \multirow{2}{*}{2216} \\
&  \color{ForestGreen} 493.75 &  \color{ForestGreen} 750.00 &  \color{ForestGreen} 1225.00 &  \color{ForestGreen} 1575.00 &  \color{ForestGreen} 2437.50 & \color{BrickRed} 0.0 & \color{BrickRed} 0.0  &  &  &  &  &  \\ \hline
\multirow{2}{*}{Educational Codeforces Round 180 (Rated for Div. 2)} & \bf A  & \bf B  & \bf C  & \bf D  & \bf E  & \bf F  &  &  &  &  &  &  \multirow{2}{*}{5} & \multirow{2}{*}{43.00} & \multirow{2}{*}{8/17128} & \multirow{2}{*}{2253} \\
& \color{ForestGreen} + & \color{ForestGreen} + & \color{ForestGreen} + & \color{ForestGreen} + & \color{ForestGreen} + & \color{BrickRed} -  &  &  &  &  &  &  \\ \hline
\multirow{2}{*}{Codeforces Round 1035 (Div. 2)} & \bf A  & \bf B  & \bf C  & \bf D  & \bf E  & \bf F  &  &  &  &  &  &  \multirow{2}{*}{2985.71} & \multirow{2}{*}{-} & \multirow{2}{*}{587/15624} & \multirow{2}{*}{2008} \\
&  \color{ForestGreen} 500.00 &  \color{ForestGreen} 1000.00 &  \color{ForestGreen} 1485.71 & \color{BrickRed} 0.0 & \color{BrickRed} 0.0 & \color{BrickRed} 0.0  &  &  &  &  &  &  \\ \bottomrule
\end{tabular}}
\label{tab:elo_rating_14b}
\end{table}

\clearpage
\setcitestyle{numbers}
\bibliographystyle{plainnat}
\bibliography{paper}

\end{document}